\documentclass[runningheads]{llncs}
\usepackage{graphicx}
\usepackage{comment}
\usepackage{amsmath,amssymb} 
\usepackage{color}
\usepackage{multirow}

\usepackage[width=123mm,left=12mm,paperwidth=146mm,height=193mm,top=12mm,paperheight=217mm]{geometry}

\begin{document}
\pagestyle{headings}
\mainmatter
\def\ECCVSubNumber{7131}  

\title{PanDA: Panoptic Data Augmentation} 

\titlerunning{PanDA: Panoptic Data Augmentation}
%
\author{Yang Liu\inst{1} \and Pietro Perona\inst{1} \and Markus Meister\inst{1}}

%
\authorrunning{Y. Liu et al.}
%
\institute{California Institute of Technology, Pasadena CA 91125, USA 
\email{\{yliu7,perona,meister\}@caltech.edu}}
\maketitle

\begin{abstract}
The recently proposed panoptic segmentation task presents a significant challenge of image understanding with computer vision by unifying semantic segmentation and instance segmentation tasks. In this paper we present an efficient and novel panoptic data augmentation (PanDA) method which operates exclusively in pixel space, requires no additional data or training, and is computationally cheap to implement. By retraining original state-of-the-art models on PanDA augmented datasets generated with a single frozen set of parameters, we show robust performance gains in panoptic segmentation, instance segmentation, as well as detection across models, backbones, dataset domains, and scales. Finally, the effectiveness of unrealistic-looking training images synthesized by PanDA suggest that one should rethink the need for image realism for efficient data augmentation.

\end{abstract}

\section{Introduction}

With the rapid development of convolutional neural networks (CNN) and the availability of large scale annotated datasets like ImageNet \cite{russakovsky2015imagenet}, modern computer vision models have reached or surpassed human performance in many domains of visual recognition \cite{he2015delving,cirecsan2012multi}. Much of the interest of the community has since shifted from visual recognition to formulating and solving more challenging tasks such as object detection, semantic segmentation, and instance segmentation. 

Recently, \cite{kirillov2019panoptic} proposed the panoptic segmentation task that unifies instance segmentation and semantic segmentation. The task requires a model to assign each pixel of an image a semantic label and an instance ID. Several panoptic datasets, such as the Cityscapes \cite{cordts2016cityscapes}, Microsoft COCO \cite{lin2014microsoft}, ADE20K \cite{zhou2017scene}, and Mapillary Vistas \cite{neuhold2017mapillary}, have been released. Much of research attention has focused on developing new models \cite{xiong2019upsnet,porzi2019seamless,gao2019ssap}. So far, the best performing models on the leader boards of various major panoptic challenges are exclusively CNN based. 

In this paper, insead of developing new models, we focus on the data augmentation aspect of the panoptic segmentation task, and develop a novel panoptic data augmentation method - PanDA that further improves the performance of original top performing models from 2019.

We first identify the data deficiency and class imbalance problems and propose a pixel space data augmentation method for panoptic segmentation datasets that efficiently generates synthetic datasets from the original dataset. The method is computationally cheap and fast, and it requires no training or additional data. To demonstrate the robustness of PanDA, we use \textbf{a single frozen set of parameters} throughout the paper except for the Ablation Study section.

Next, we experimentally demonstrate that training with PanDA augmented Cityscapes further improves all performance metrics of original top-performing UPSNet \cite{xiong2019upsnet} and Seamless Scene Segmentation \cite{porzi2019seamless} models. With PanDA, robust boosts in panoptic segmentation, instance segmentation, and detection with UPSNet with ResNet-50/101 backbones and Seamless Scene Segmentation model with ResNet-50 backbone are reported.

To further demonstrate the generalizability of PanDA across scales and domains, we apply it to Cityscapes subsets with 10 to 3,000 images, as well as a 10 times larger COCO subset with 30,000 images. We report performance gains across scales and datasets. By quantifying the log-linear relationship between number of training images and final performance metrics, we show that on average a PanDA generated image is 20-40\% as effective as an original image.

Finally, results from the ablation study show that, contrary to common beliefs, less realistic looking images improve model performance more. It suggests that the level of image realism is not positively correlated with data augmentation efficiency.

\section{Related Work}

 \subsection{Panoptic Segmentation}
 
 The panoptic segmentation task was first proposed in 2019 by \textit{Kirillov et al.} \cite{kirillov2019panoptic} as an attempt to formulate a unified task that bridges the gap between semantic segmentation and instance segmentation. The task divides objects in to two super-categories: \textit{things} and \textit{stuff}. For classes in things, each pixel in the image is labeled with a class ID and an instance ID. For classes in stuff, each pixel is labeled with a class ID only. In addtion to traditional metrics such as mean intersection over union (mIoU) and average precision of instance segmentation (AP), the panoptic quality (PQ) is defined as, 

\[ PQ =  \underbrace{\frac{\sum_{(p,g) \in {TP}} IoU(p,g)}{|TP|}}_\text{segmentation quality (SQ)} 
\times \underbrace{\frac{|TP|}{|TP| + \frac{1}{2}|FP| + \frac{1}{2}|FN|}}_\text{recognition quality (RQ)} \]
 
 where true positives (TP) are predicted segments that have strictly greater than 0.5 overlap with ground truth segments. PQ is the product of segmentation quality (SQ) and recognition quality (RQ). 
 
 \subsection{Data Augmentation}

Despite the efforts towards low-shot learning, modern CNN based models are still data-hungry in that they have very large model capacity to even memorize randomly labeled large datasets \cite{zhang2016understanding}. One efficient way to regularize models and promote generalization is data augmentation.

Many methods still in use for detection and segmentation models \cite{xiong2019upsnet,porzi2019seamless,gao2019ssap} are largely inherited from the ImageNet visual recognition era \cite{krizhevsky2012imagenet}. These methods take advantage of object invariances in pixel space: manipulations such as crop, horizontal flip, resize, color distortion, and noise injection do not usually change the identity of the object. These methods are simple in concept and computationally very cheap, but they do not take advantage of the more informative ground truth labelling that accompanies the harder detection and segmentation tasks. Some recent methods start to explore the use of bounding box and segmentation information \cite{liu2019pixel,beery2019synthetic,fang2019instaboost,shetty2019not,dvornik2019importance}. InstaBoost \cite{fang2019instaboost} is proposed as a method for instance segmentation where instances are cropped out and moved to a different location of the same image, then the hole in background is filled with a classical in-paint network \cite{bertalmio2001navier}. Shetty et al. \cite{shetty2019not} augment images by removing certain instances that provide context and use a CNN based in-paint model \cite{shetty2018adversarial} to fill the holes. Dvornik et al. \cite{dvornik2019importance} use a CNN based context model to guide the addition of instances from an ``instance-database'' to original images, avoiding making holes in the original images and the need for in-painting.  The Discussion Section offers a detailed comparison relating \cite{fang2019instaboost}, \cite{shetty2019not} and \cite{dvornik2019importance} to the present work.

Data simulation is another flavor of data augmentation. Thanks to the development of graphical simulation engines, one can generate photo-realistic images together with pixel perfect ground truth. The method is proven to work for many tasks such as human pose estimate \cite{varol2017learning}, wildlife classification \cite{beery2019synthetic}, and object detection \cite{hinterstoisser2019annotation}. However, this method often requires handcrafted 3D models and there usually is a significant domain gap between real images and simulated images.

Recently, with the popularization of generative models such as generative adversarial networks (GAN), researchers also add images generated by GANs to the training set \cite{frid2018gan,bowles2018gan,liu2019pixel}. However, as a type of neural network based model, the GAN itself requires training which in turn is both computationally expensive and data hungry. 

\section{Panoptic Data Augmentation} 

\begin{figure*}[h]
\begin{center}
\centering
\includegraphics[width=0.85\textwidth]{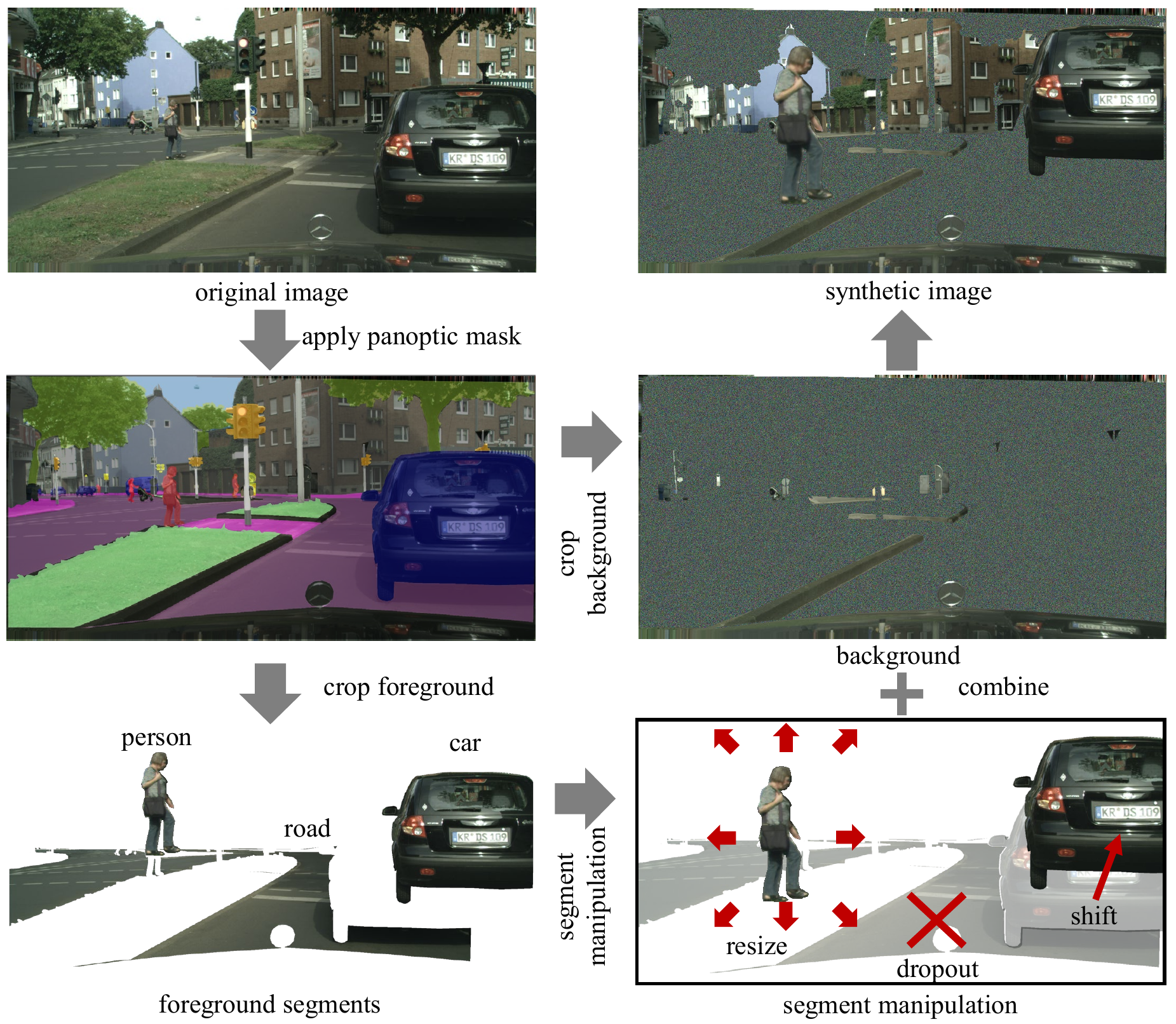}
\end{center}
   \caption{\textbf{Schematics of PanDA.} Foreground and background segments are greatly simplified for clear demonstration of the image decomposition and synthesis process. Note that background in the panoptic segmentation task usually takes a small percentage of pixels in an image}
\label{fig:schematics}
\end{figure*}

\begin{figure*}[h]
\begin{center}
\centering
\begin{tabular}{c c c}
     \includegraphics[width=0.33\textwidth]{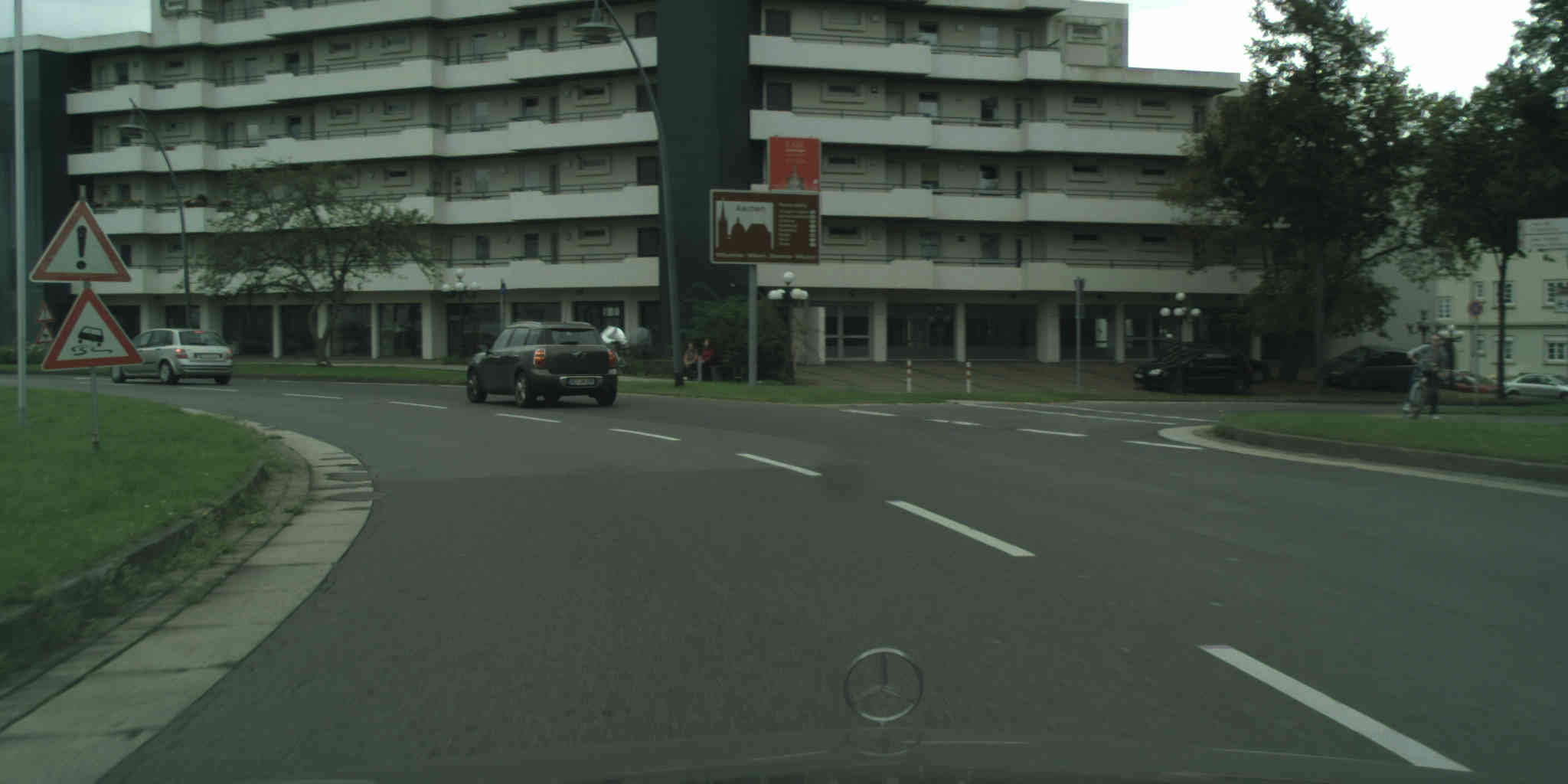} & \includegraphics[width=0.33\textwidth]{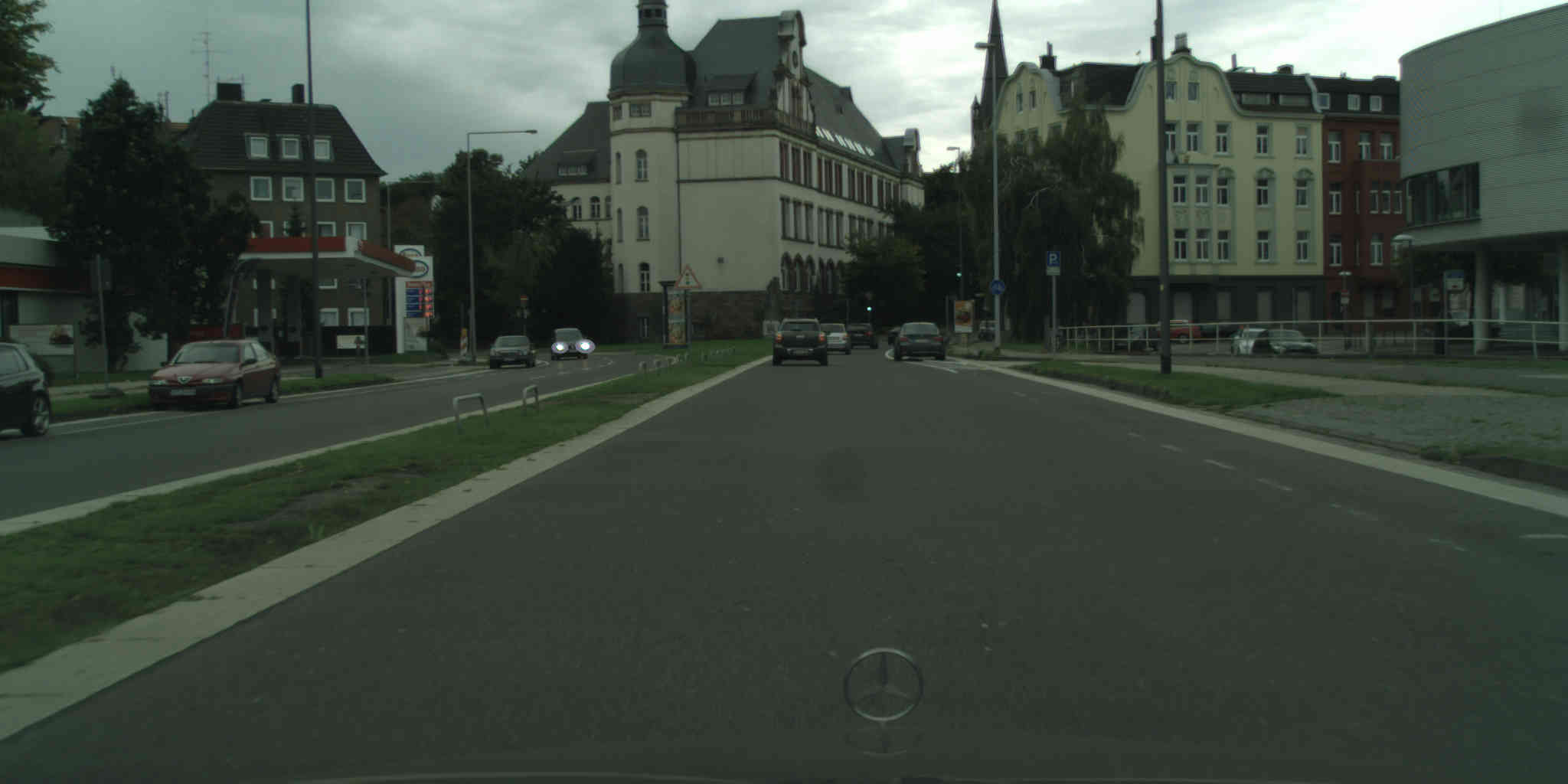} & \includegraphics[width=0.33\textwidth]{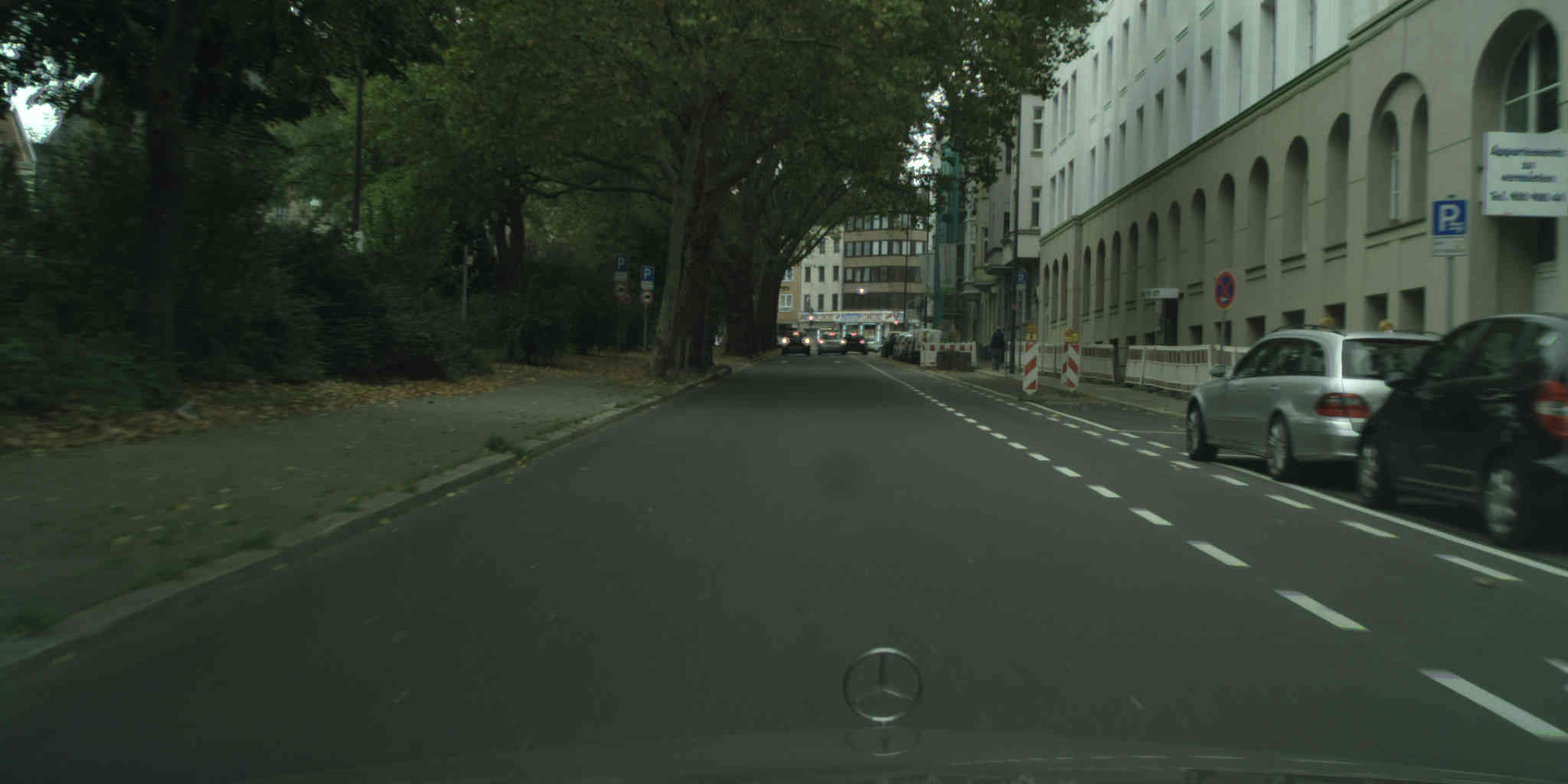}\\ 
     (a) & (b) & (c)\\
     \includegraphics[width=0.33\textwidth]{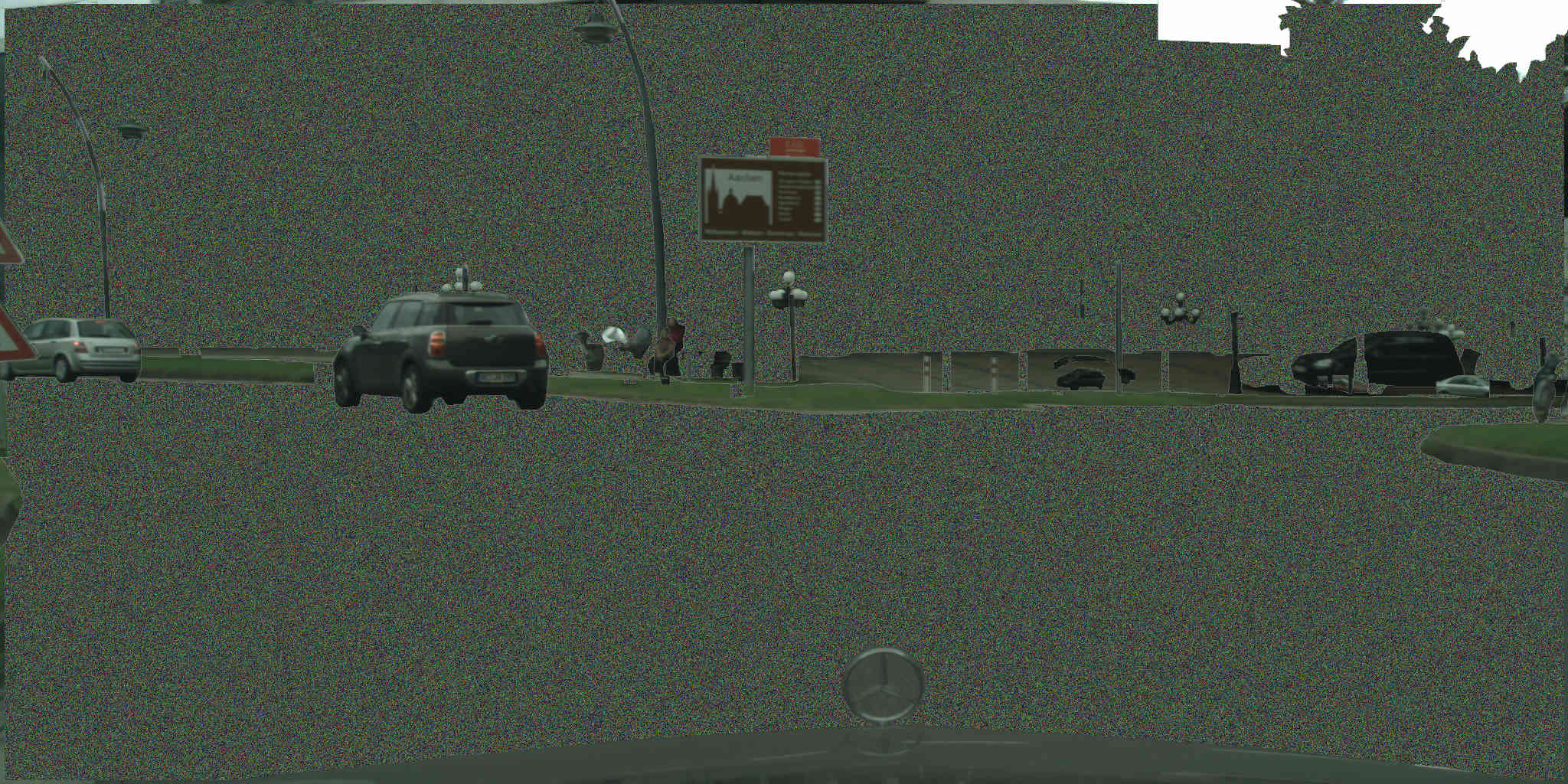} & \includegraphics[width=0.33\textwidth]{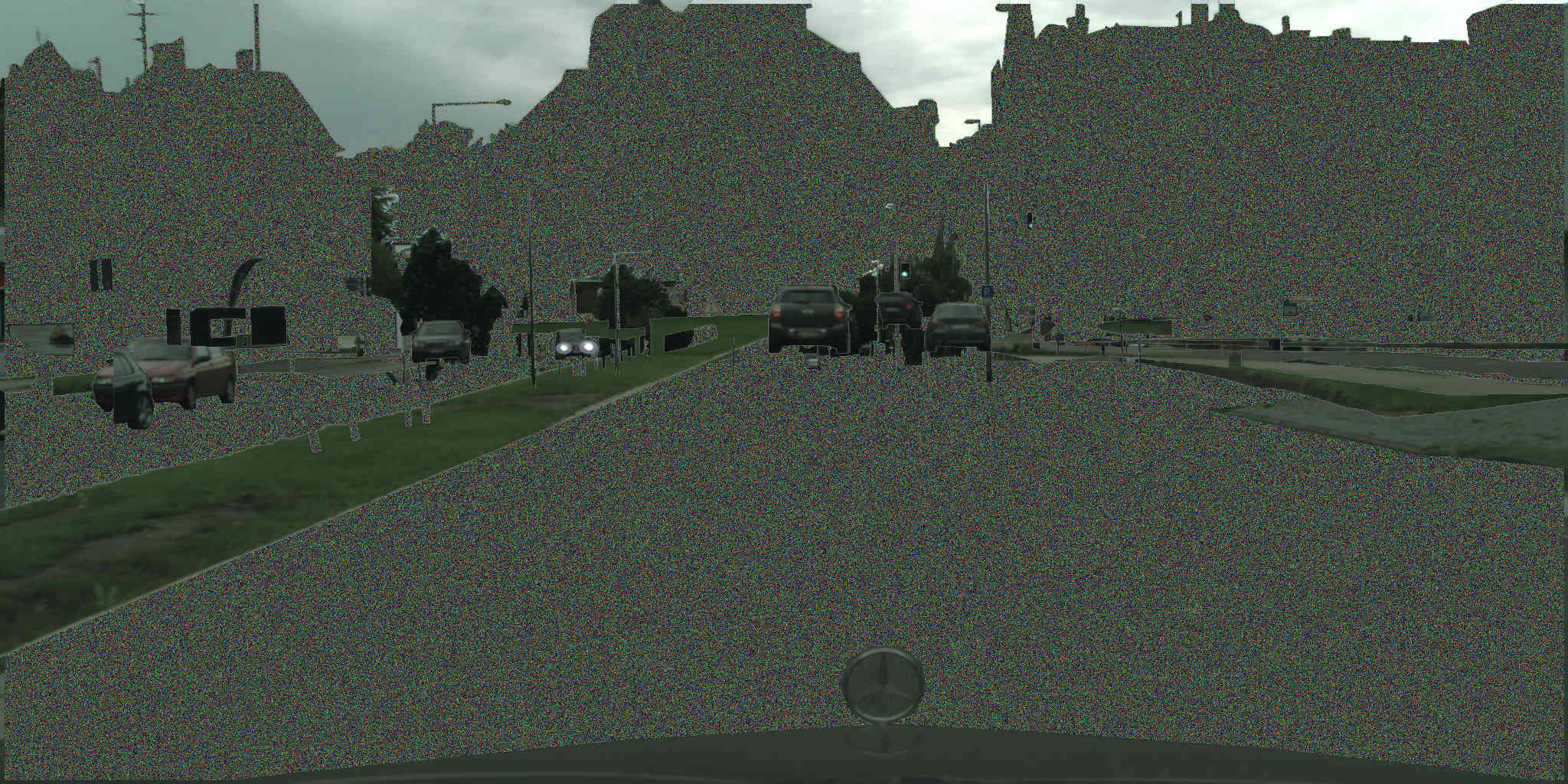} &  \includegraphics[width=0.33\textwidth]{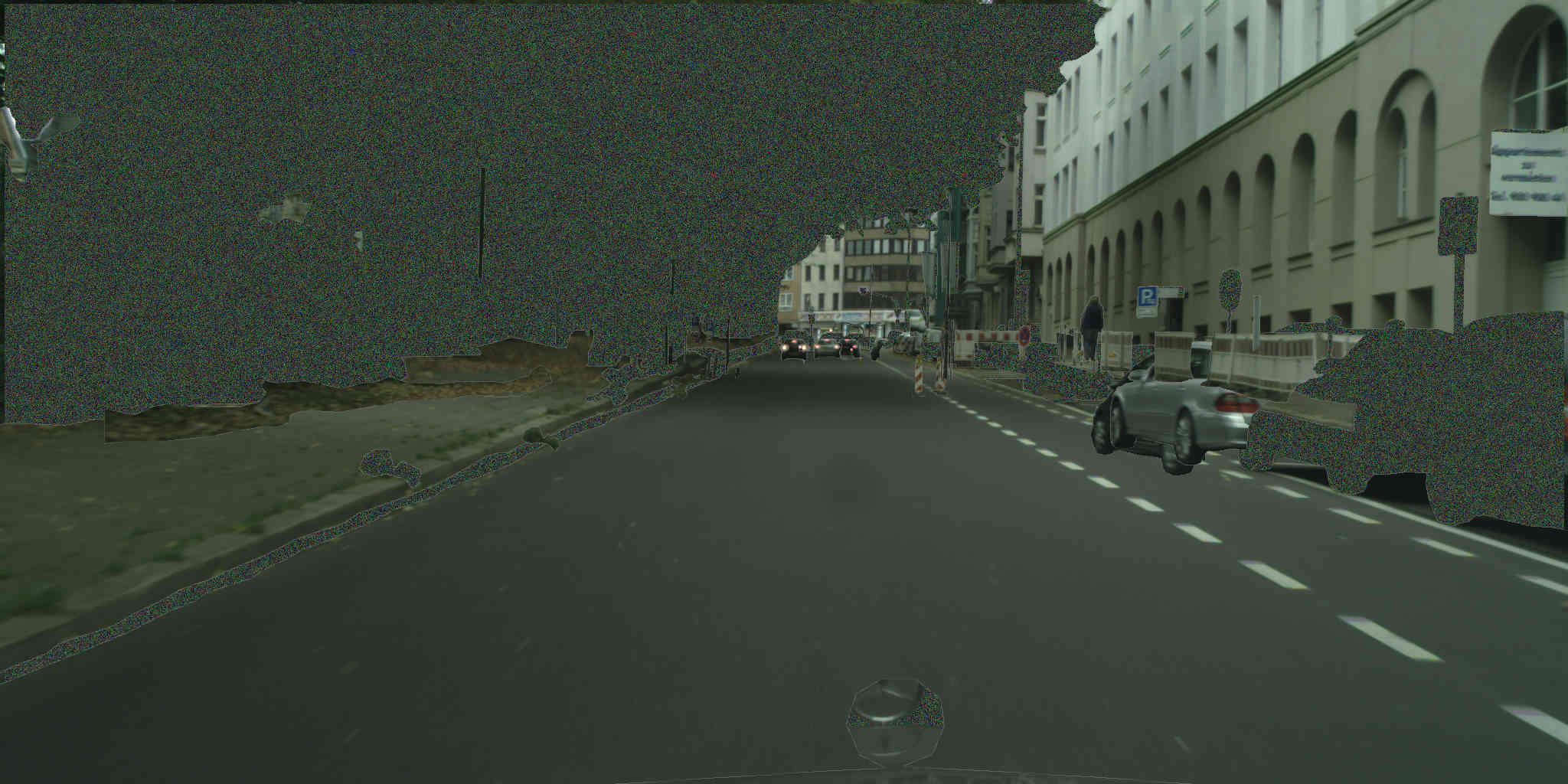}\\ 
     (d) & (e) & (f) \\
\end{tabular}
\end{center}
   \caption{\textbf{Examples of original cityscapes images (\textit{top row}) and PanDA generated images (\textit{bottom row})}}
\label{fig:examples}
\end{figure*}

Annotating images for panoptic segmentation can be costly, since every pixel in an image has to be semantically labeled and pixels that belong to \textit{things} classes must have an additional instance ID associated. One could argue that it is orders of magnitude harder for a human annotator to generate panoptic ground truth than to assign a single class ID to an image as needed in visual recognition. But with great challenges usually come great opportunities, PanDA takes advantage of the rich information embedded in panoptic annotations and uses it to synthesize new images with ground truth.

A fundamental feature that distinguishes PanDA from other data augmentation methods is that PanDA does not aim to create images that look realistic to human eyes. Many 3D simulation based methods \cite{beery2019synthetic,richter2016playing,varol2017learning} and cut-paste methods \cite{shetty2019not,dvornik2019importance,fang2019instaboost} implicitly or explicitly suggest that the level of realism is key to achieving high performances. Similary, GAN based image synthesis methods \cite{frid2018gan,bowles2018gan,huang2017beyond} also usually optimize for realism, since the very goal of the generator network in GANs is to synthesize images realistic enough to fool the discriminator network into thinking that they are real \cite{goodfellow2014generative}. In contrast, PanDA takes more principled approaches to balance classes, breaks local pixel structures, and increases variations of objects, and the final images synthesized do not look fully realistic to human eyes.

As shown in Fig \ref{fig:schematics}, PanDA first decomposes a panoptically labelled training image by using the segmentation ground truth to divide it into foreground and background segments. It is worth noting that foreground segments not only include instances in \textit{things} classes, but also segments in \textit{stuff} classes which are considered background in InstaBoost \cite{fang2019instaboost} and \cite{dvornik2019importance}. Unfilled pixels are padded with white noise patterns instead of in-painting as seen in \cite{fang2019instaboost} and \cite{shetty2019not}. Noise patterns are used because they do not belong to any known categories, making ground truth assignment unambiguous. Foreground segments are then overlaid on top of the new background image one by one. The same manipulation is applied to the ground truth segmentation image to generate the new ground truth segmentation.

For foreground segments, we can perform a series of pixel space operations such as dropout, resize and shift to control different aspects of each individual object instance. \textbf{Dropout} is used to remove segments from an image. It serves to revome well segmented and classfied objects from the image.  The \textbf{resize} operation changes the size of a segment while preserving the original aspect ratio, it simulates object movement in depth and introduces more size variance in objects. It resembles zooming and multi-scale training \cite{xiong2019upsnet} on the individual object level. Random \textbf{shifting} moves the segment in x and y in pixel space. It prevents memorization of object locations by the model and breaks the local pixel relationship between the object and its background, thus creates new local contexts for the object. Resize and shift together simulate 3 dimensional translation of objects in space, and dropout controls the frequency of objects. Because the ground truth depth information to recover the depth ordering of segments is not available, and we do not want larger segments to occlude smaller ones, we sort the segments by their area and put largest segments on the bottom layer. It is worth noting that more operations can be added to the repertoire of the PanDA tool set, and different variants of the aforementioned operations may also be implemented. In this paper, we limit the scope to the operations discussed above to allow for in-depth experiments with PanDA.

\section{Experiments} 
\subsection{UPSNet and Seamless Scene Segmentation Models}

The UPSNet model and the Seamless Scene Segmentation (SSS) models are two independently developed top-performing panoptic segmentation models. To date, they are also the only two models with code published to reproduce the results.

The UPSNet model \cite{xiong2019upsnet} is one of the top performing single models on both Cityscapes and MS COCO panoptic challenges. It uses a shared CNN feature extractor backbone and multiple heads on top of it. The backbone is a Mask R-CNN feature extractor built on a deep residual network (ResNet) with a feature pyramid network (FPN). UPSNet has an instance head which follows the Mask R-CNN design and a semantic head that consists of a deformable convolution based sub-network. The outputs of the two heads are then fed into the panoptic head which generates the final panoptic prediction.

The SSS model \cite{porzi2019seamless}, which was developed around the same time early in 2019, roughly follows the same meta-architecture with a Mask R-CNN instance head and a Mini Deeplab (MiniDL) semantic head taking input from a common FPN feature extractor and feeding into a fusion head for panoptic prediction. At the time of its publication, it became the top performing model in Cityscapes panoptic challenge with a ResNet-50 backbone \cite{porzi2019seamless}.

 We used all the original hyper-parameters of UPSNet and SSS models for training and inference. We only scale (1) the number of training iterations to keep the number of epochs constant across a wide range of dataset sizes and (2) the learning rate based on the number of GPUs used in the training. All trainings start with  ImageNet pretrained ResNet-50/101 models as reported in \cite{xiong2019upsnet} and \cite{porzi2019seamless}. The UPSNet models are trained with 8x Nvidia 2080Ti GPUs with 11GB VRAM. The SSS experiments are conducted on Amazon Web Services with 8x Nvidia Tesla V100 GPUs with 32GB VRAM. For the in-depth experiments, we choose UPSNet over SSS because it fits in our server's GPUs.

\subsection{Datasets}

The Cityscapes \cite{cordts2016cityscapes} panoptic dataset has 2,975 training, 500 validation, and 1,525 test images of ego-centric driving scenes in urban environments in many cities in Europe. Examples are shown in Fig \ref{fig:examples} (a-c).  The dense annotation covers 97\% of the pixels which consists of 9 \textit{things} classes and 11 \textit{stuff} classes. We choose Cityscapes as the main testbed for PanDA for several reasons: 1) It is one of the most popular panoptic datasets available and it has a total of 19 diverse classes that covers a wide range of driving related scenarios. Many modern models\cite{xiong2019upsnet,porzi2019seamless,gao2019ssap} report performance on Cityscapes and have published code available.  2) Results on a relatively small set are likely to generalize to other specialized domains where annotated panoptic data is scarce. 3) The small size also makes both data synthesis and training cost manageable. As a result, it is suitable for exploratory studies like this.

To demonstrate the generalizability of PanDA, we also performed additional experiments on the MS COCO dataset, which has 118K training and 5k validation images including 53 \textit{stuff} classes and 80 \textit{things} classes of common objects. It is both orders of magnitude larger in size and more diverse regarding number of classes.

\subsection{Augmenting Cityscapes with PanDA}

\begin{figure}[h]
\begin{center}
\includegraphics[width=0.7\linewidth]{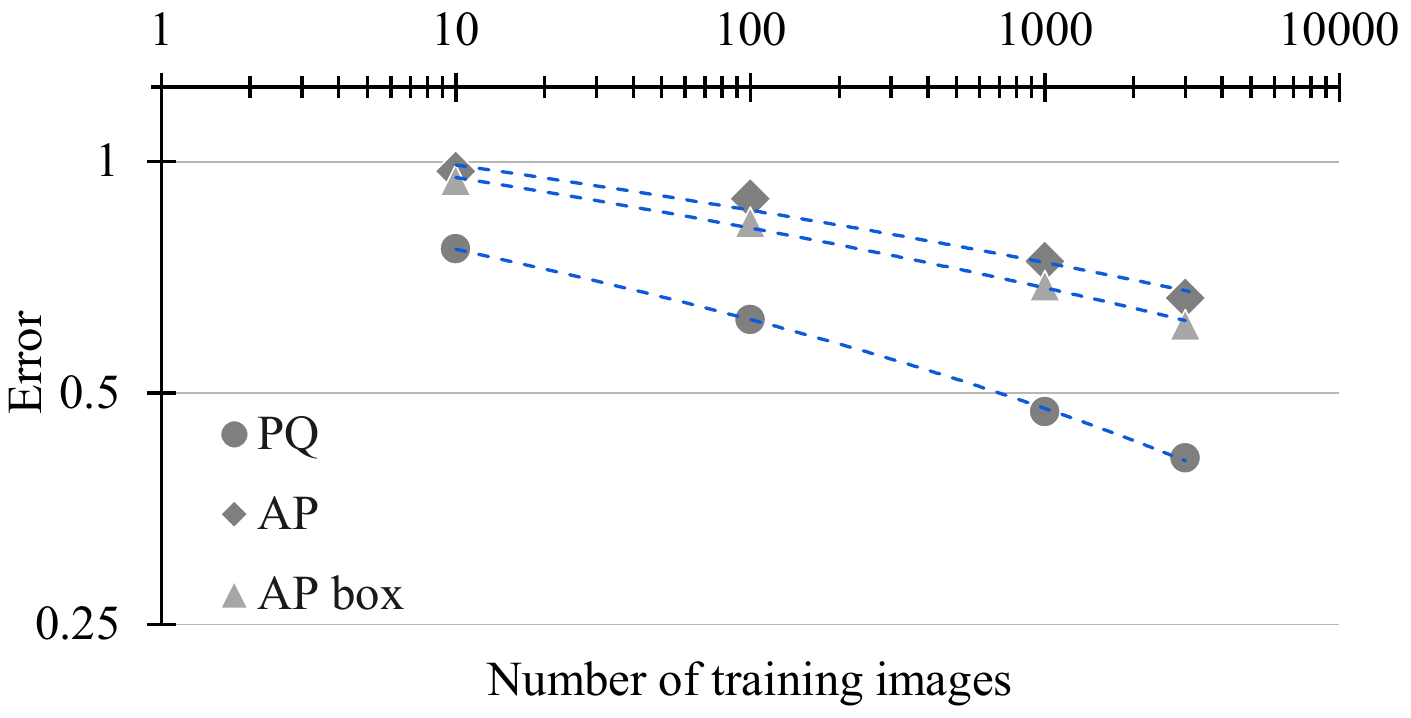}
\end{center}
   \caption{\textbf{Model performance vs training set size on Cityscapes.} We train UPSNet-50 models on various Cityscapes subsets ranging from 10 to 2,975 images. \textit{PQ}: panoptic quality. \textit{AP}: mean average precision of instance segmentation evaluated at 0.5:0.95. \textit{AP box}: mean average precision of detection evaluated at 0.5:0.95. Dashed curves are log-linear fits. Panoptic segmentation, instance segmentation, and instance detection performance are summarized by PQ, AP, and AP box, respectively}
\label{fig:performance_vs_number}
\end{figure}

 The fact that multi scale inference and pretraining with additional data improves performance of the lightweight UPSNet-50 model suggests that model capacity is not the major limiting factor for the final performance and that data augmentation may further improve performance. To investigate the relationship between model performance and size of training dataset, we trained the UPSNet-50 model from the same ImageNet pretrained backbone on various subsets of the Cityscapes training set that consist of 10, 100, 1,000, and 2,975 images. Fig \ref{fig:performance_vs_number} shows near perfect log-linear relationships between the number of training images and various performance metrics (PQ: $r^2=0.9995$, mean average precision for instance segmentation evaluated at 0.5:0.95 (AP): $r^2=0.9744$, mean average precision for detection evaluated at 0.5:0.95 (AP box): $r^2=0.9948$). It suggests that adding training images is likely to further improve model performance.
 
 To test whether adding training images indeed helps, we used PanDA to generate synthetic images from the Cityscapes training set. A key discrepancy between model training and and the final PQ score is that classes with large areas provide overwhelming training signals during training and thus are favored during training, whereas SQ - the first term of the PQ formula - weights small and large objects equally. To mitigate the issue, we aim to balance average per class pixel count per image by applying random dropout of segments where the dropout rate is linearly proportional to the size of the segments. We also apply resizing and shifting to introduce size and location variance of objects. 
 
 We performed a grid search of the parameters of PanDA to optimize PQ of UPSNet-50 model on the full Cityscapes dataset. To demonstrate the generalizability and robustness, we \textbf{froze PanDA's parameters} and \textbf{used it throughout the rest of the paper} with the exception of the Ablation Study section. Specifically, drop out is performed as follows: segments with areas between 10\% and 50\% of an image are dropped out with probability linearly proportional to their sizes. Segments occupying over 50\% of an image are guaranteed to drop out, and segments smaller than 10\% of the image frame are never dropped out.  We couple resize and shift with a zooming operation: enlarged segments are pushed to the periphery and shrunk ones are pulled towards the center. This simulates an approaching motion of the object. Each segment is resized in the range of 0.5x to 1.5x, with scales drawn from a uniform distribution.
 
 In original Cityscapes, \textit{road} and \textit{building} together occupy over 50\% pixels of an image on average which supports our observation that some large classes dominate learning signals. While PanDA removes a large proportion of non-background pixels on average, it significantly reduces the dominance of common large classes while largely preserving small and rare classes (Supplementary Figure 1 and 2). 
 
 Examples of original and synthetic image pairs are shown in Fig \ref{fig:examples} (and Supplementary Fig \ref{fig:additional_cityscapes_examples}): some of the commonly seen classes with large area such as \textit{road} and \textit{building} are more frequently dropped out and smaller instances such as \textit{person} and \textit{pole} are often kept. It has caught our attention that the synthetic images are no longer realistic and coherent scenes but rather nonsensical to human eyes. However we show in later sections that adding these synthetic images improves model performance.

\begin{table}[h]
\begin{center}
\caption{\textbf{Results on Cityscapes.} PanDA augmented Cityscapes datasets generated with a single frozen set of parameters improve all performance of UPSNet-50, UPSNet-101, SSS-50. \textit{PQ}: panoptic quality. \textit{SQ}: segmentation quality. \textit{RQ}: recognition quality. \textit{PQ$^{th}$}: PQ for \textit{things} classes. \textit{PQ$^{st}$}: PQ for \textit{stuff} classes. \textit{mIoU}: mean intersection over union. \textit{AP}: mean average precision of instance segmentation evaluated at 0.5:0.95. \textit{AP$^{box}$}: mean average precision of detection evaluated at 0.5:0.95}
\label{table:performance_main}
\begin{tabular}{l c|c c c|c c c c c}
\hline
 Model & PanDA & PQ & SQ & RQ & PQ$^{Th}$ & PQ$^{St}$ & mIoU & AP & AP$^{Box}$ \\
\hline
\hline
\multirow{2}{6em}{UPSNet-50} &  & 58.8 & 79.5 & 72.6 & 54.5 & 62.0 & 75.1 & 33.5 & 38.7 \\

 & \checkmark & \bf59.9 & \bf79.9 & \bf73.8 & \bf55.4 & \bf63.3 & \bf76.2 & \bf34.6 & \bf39.6 \\

\hline
\multirow{2}{6em}{UPSNet-101} & & 59.8 & 80.0 & 73.5 & 55.6 & 62.9 & 76.7 & 33.0 & 38.2 \\

& \checkmark & \bf60.5 & \bf80.2 & \bf74.1 & \bf56.3 & \bf63.6 & \bf77.1 & \bf35.0 & \bf40.6 \\
\hline
\multirow{2}{6em}{SSS-50} &  & 60.3 & -- & -- & 55.5 & 63.8 & 77.4 & 32.4 & 36.3 \\

& \checkmark & \bf60.9 & -- & -- & \bf56.4 & \bf64.2 & \bf78.0 & \bf33.8 & \bf37.0 \\
\hline
\end{tabular}
\end{center}
\end{table}

To demonstrate the usefulness of PanDA, we first trained original UPSNet and SSS models from ImageNet pre-trained ResNet backbones to establish baseline performance metrics (Table \ref{table:performance_main}). The reproduced results largely recapitulate those published in the original papers (For details, see Supplementary Table \ref{table:performance}). Then we generated synthetic training sets with PanDA and trained original UPSNet and SSS models on the augmented training sets. We first used PanDA to double the Cityscapes training set and trained the original UPSNet-50 model on the PanDA augmented Cityscapes dataset with a total of 3,000 images. The model (UPSNet-50 with PanDA in Table \ref{table:performance_main} and Supplementary Fig \ref{fig:main_results}) outperforms the baseline model (as well as the one reported in \cite{xiong2019upsnet}) in all metrics. In summary, UPS-50 trained on PanDA augmented images improves on the baseline by 1.1 PQ, 0.9 AP, 1.1 AP box, evaluated at 0.5:0.95) and 1.1 mean Intersection over Union (mIoU) (0.7 PQ, 1.0 AP, 1.3 AP box, and 1.0 mIoU improvements over \cite{xiong2019upsnet}) on the validation set. Additionally, overall and per class performance in both instance segmentation task (Supplementary Table \ref{table:instance_performance} and detection task (Supplementary Table \ref{table:detection_performance}) improve upon baseline.

One commonly used method to improve model performance at the cost of increased computation is to use a larger and deeper backbone. We trained a UPSNet-101 model which uses the deeper ResNet-101 and observed performance gain for all metrics except for AP and AP box. We froze PanDA parameters and trained UPSNet-101 model with PanDA augmented images. Table \ref{table:performance_main} shows that the PanDA enhanced UPSNet-101 model in turn outperforms all other UPSNet models. Two observations can be made from the additional experiments with UPSNet-101. First, it is remarkable that with a half-sized back bone, the PanDA enhanced UPSNet-50 model outperforms the baseline UPSNet-101 model in five out of eight metrics: PQ, RQ, PQ stuff, AP and AP box. Second, the fact that PanDA further improves UPSNet-101 suggests that using a deeper backbone and training on PanDA augmented images help the model in complementary ways. In other words, it demonstrates that PanDA generalizes across backbones.

In addition to UPSNet, we also conducted experiments on SSS, which is the only other model with code available to date. We trained the SSS model on PanDA augmented images with the same frozen parameters. The last row in Table \ref{table:performance_main} shows all-around improvements of the SSS model trained with PanDA. Together, the set of experiments with UPSNet-50/101 and SSS shows that PanDA is effective across models and backbones.

\subsection{PanDA Across Scales and Datasets}

\begin{table}[h]
\begin{center}
\caption{\textbf{PanDA generalization results.} We used the same set of parameters for all experiments in this section. As shown in the table, PanDA generalizes well not only across scales of Cityscapes subsets, but also to COCO subsets that are 10 times larger than the original Cityscapes dataset}
\label{table:generalization}
\begin{tabular}{l c c |c c c|c c c c c}
\hline
Dataset & \# images & PanDA & PQ & SQ & RQ & PQ$^{Th}$ & PQ$^{St}$ & mIoU & AP & AP$^{Box}$ \\
\hline
\hline
\multirow{8}{6em}{Cityscapes} & 10 & & 23.0 & \bf58.1 & 29.6 & 9.2 & 33.1 & \bf36.3 & 2.9 & 5.5\\
 & 10 & \checkmark & \bf23.5 & 54.9 & \bf30.3 & \bf12.5 & \bf30.3 & 34.3 & \bf4.6 & \bf7.6\\
& 100 & & 37.7 & \bf72.5 & 48.4 & 26.4 & 45.9 & 52.4 & 10.5 & 16.7\\
& 100 & \checkmark & \bf40.1 & 70.6 & \bf51.0 & \bf29.5 & \bf47.7 & \bf54.1 & \bf12.7 & \bf19.1\\
& 1,000 & & 52.7 & 77.9 & 65.8 & 45.5 & 57.9 & 68.7 & 25.8 & 31.1\\
& 1,000 & \checkmark & \bf55.0 & \bf78.4 & \bf68.5 & \bf49.1 & \bf59.2 & \bf70.1 & \bf27.3 & \bf32.5\\
& 3,000 & & 59.3 & 79.7 & 73.0 & 54.6 & 62.7 & 75.2 & 33.3 & 39.1\\
& 3,000 & \checkmark & \bf59.9 & \bf79.9 & \bf73.8 & \bf55.4 & \bf63.3 & \bf76.2 & \bf34.6 & \bf39.6\\
\hline
\multirow{2}{6em}{COCO} & 30,000 & & 36.5 & 76.2 & 45.7 & 41.6 & \bf28.8 & 45.3 & 26.5 & 29.0\\
& 30,000 & \checkmark & \bf37.4 & \bf76.9 & \bf46.6 & \bf43.2 & 28.7 & \bf45.9 & \bf28.0 & \bf31.2\\
\hline
\end{tabular}
\end{center}
\end{table}

To test how well PanDA generalizes across scales, we applied it to smaller subsets of Cityscapes. Such generalization would be particularly useful if one develops a new panoptic task in a new domain where annotated data is expensive and scarce. In Table \ref{table:generalization} (and Supplementary Table \ref{fig:performance_vs_number_supp}), we show the consistent performance improvement on UPSNet models trained with PanDA augmented Cityscapes subsets across scales of 10, 100, 1,000, and 3,000 images.

We then ask whether the improvement with PanDA is specific to the Cityscapes dataset. Arguably, the ego-centric driving scenarios in urban environments in Cityscapes are a specialized domain which may raise concerns of generalizability across domains. In addition, Cityscapes is one of the smaller panoptic datasets available. Although previous experiments demonstrate that PanDA performs well when the dataset scales down, it remains unknown how well it performs when scaled up. To address the concerns, we applied PanDA to a 30,000 image subset of COCO dataset which is obtained by going through a list of the original 118K training set with step size 4. The COCO 30K subset will not only test whether PanDA generalizes to a different domain (examples in Supplementary Fig \ref{fig:additional_coco_examples}), but also breaks the 3,000-image upper limit of Cityscapes. The bottom two rows of Table \ref{table:generalization} show that PanDA indeed works on the 10x larger COCO dataset, which further supports the claim that PanDA generalizes well across scales and domains.

\subsection{Data Efficiency}

In this section, we explore the data efficiency of PanDA generated images. As shown in Fig \ref{fig:performance_vs_number} (and Supplementary Fig \ref{fig:performance_vs_number_supp}), PQ, AP, and AP box scales linearly with the logarithm of number of training images. The log-linear regression functions are,
\begin{align}
     PQ=6.3255ln(n) + 8.5413 , (r^2=0.9995) \\
     AP=5.4384ln(n) - 11.494 , (r^2=0.9744) \\
     AP^{box}=5.8271ln(n) - 8.7893 , (r^2=0.9948)
\end{align}

where $n$ is the number of original images used to train the model, PQ, AP, and AP box are respective model performance. PQ, AP, AP box values are plugged into the regression functions to in turn estimate \textbf{effective training set size $N$} for experiments with either original images only or original plus PanDA images. We define \textbf{data efficiency (DE)} as,
\begin{align}
    DE =  \frac{N^{aug} - N^{orig.}}{N^{orig.}}\times 100\%
\end{align}
 
where N$^{orig.}$ is the estimated effective training set size of models trained on original images only, and N$^{aug}$ is that of models trained on original and PanDA images. The definition is specific to the case where the ratio of original and synthetic images is 1. A higher DE suggests higher per image data efficiency, and one would expect a DE of 100\% if a synthetic image is as informative as an original image.

\begin{table}[h]
\begin{center}
\caption{\textbf{Data efficiency.} Three sets of estimates of effective training set sizes are made per experiment with PQ, AP, and AP box, respectively. \# original images: number of original images used in training. $DE$: data efficiency in percent. $N$: number of effective training images estimated by model performance. Superscripts \textit{orig.} and \textit{aug} denote model trained with original images only or original and PanDA images. Subscripts denote which performance metric is used to estimate $N$ or $DE$}
\label{table:data_efficiency}
\begin{tabular}{c |c c c|c c c|c c c }
\hline
 \# orig. img & $N_{PQ}^{orig.}$  & $N_{PQ}^{aug}$ & $DE_{PQ}$ & $N_{AP}^{orig.}$  & $N_{AP}^{aug}$ & $DE_{AP}$ & $N_{AP box}^{orig.}$  & $N_{AP box}^{aug}$ & $DE_{AP box}$ \\
\hline\hline
10 & 9.8 & 10.6 & 8.2 & 14.1 & 19.3 & 36.7 & 11.6 & 16.7 & 43.4 \\
\hline
100 & 100.5 & 146.8 & 46.1 & 57.1 & 85.5 & 49.9 & 79.4 & 119.8 & 51.0 \\
\hline
1000 & 1076.1 & 1547.9 & 43.9 & 951.0 & 1253.1 & 31.8 & 939.6 & 1194.8 & 27.2 \\
\hline
3000 & 2822.6 & 3358.7 & 19.0 & 3918.2 & 4796.6 & 22.4 & 3462.4 & 4040.7 & 16.7 \\
\hline
\end{tabular}
\end{center}
\end{table}

Two conclusions can be drawn from results in Table \ref{table:data_efficiency}. First, across scales and metrics, PanDA images are not as efficient as original images because none of DEs are near 100\%. However, this is expected since we only reuse the object instances in the original images, and the synthetic images have only 40\% non-background pixels per image on average. Second, the fact that a synthetic image can be half as efficient as a real one suggests that there is significant amount of information embedded in the original images that is not learned by current models. One can expect superior models to capture this additional information without the help of data augmentation.

\subsection{Ablation Study}

\begin{table}[h]
\begin{center}
\caption{\textbf{Ablation study on Cityscapes dataset.} Training on the original dataset for more iterations does not improve model performance. Best performance is achieved by combining dropout, resize, and shift}
\label{table:ablation}
\begin{tabular}{c c c c|c c c|c c c c c}
\hline
 Extra Iterations & Dropout & Resize & Shift & PQ & SQ & RQ & PQ$^{Th}$ & PQ$^{St}$ & mIoU & AP & AP$^{Box}$ \\
\hline\hline
 & & & & 58.8 & 79.5 & 72.6 & 54.5 & 62.0 & 75.1 & 33.5 & 38.7\\
\hline
\checkmark & & & & 58.7 & \bf80.1 & 71.9 & 53.5 & 62.4 & 75.6 & 33.2 & 38.1 \\
\hline
\checkmark & \checkmark & & & 59.3 & 79.6 & 73.1 & 54.1 & 63.1 & 75.8 & 33.4 & 39.2 \\
\hline
\checkmark & \checkmark & \checkmark & & 59.0 & 79.8 & 72.6 & 55.3 & 61.7 & 75.8 & 33.4 & 39.3 \\
\hline
\checkmark & \checkmark &  & \checkmark & 59.3 & 79.9 & 72.8 & 53.9 & 63.2 & 76.0 & 34.1 & 39.4 \\
\hline
\checkmark & \checkmark & \checkmark & \checkmark & \bf59.9 & 79.9 & \bf73.8 & \bf55.4 & \bf63.3 & \bf76.2 & \bf34.6 & \bf39.6 \\
\hline
\end{tabular}
\end{center}
\end{table}

\begin{figure}[h]
\begin{center}
\begin{tabular}{ccc}
\includegraphics[width=0.32\linewidth]{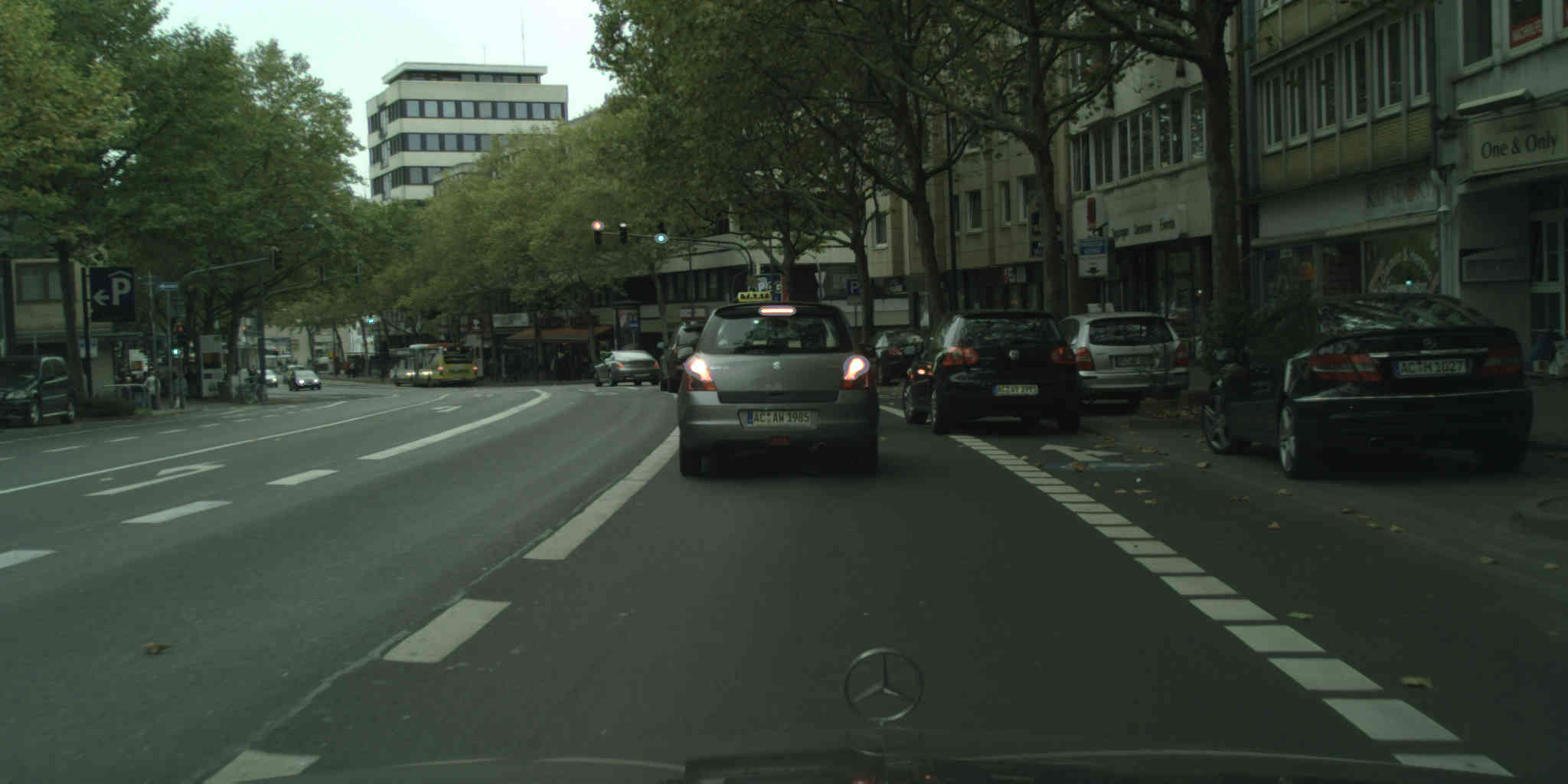} &
\includegraphics[width=0.32\linewidth]{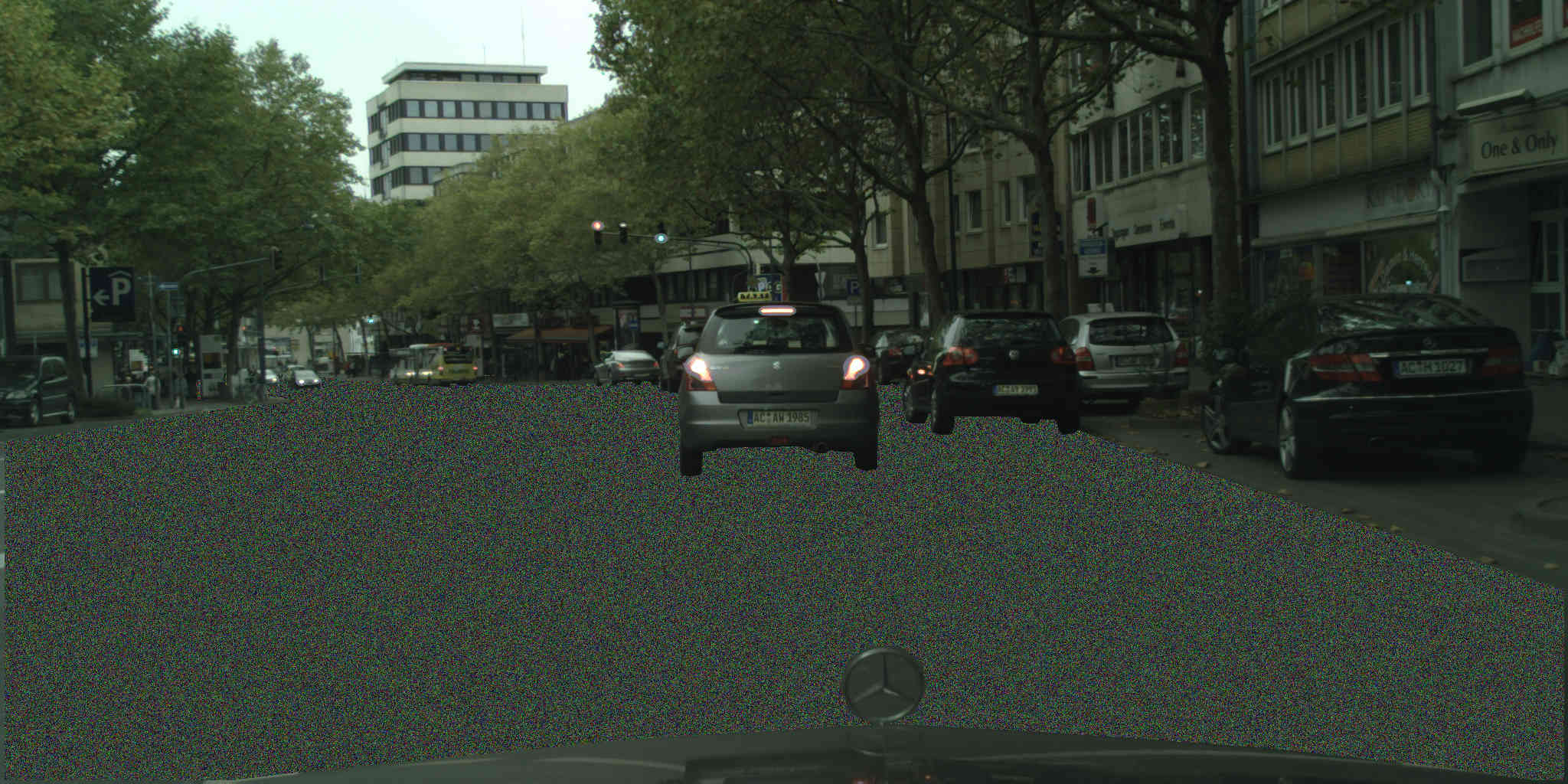} &
\includegraphics[width=0.32\linewidth]{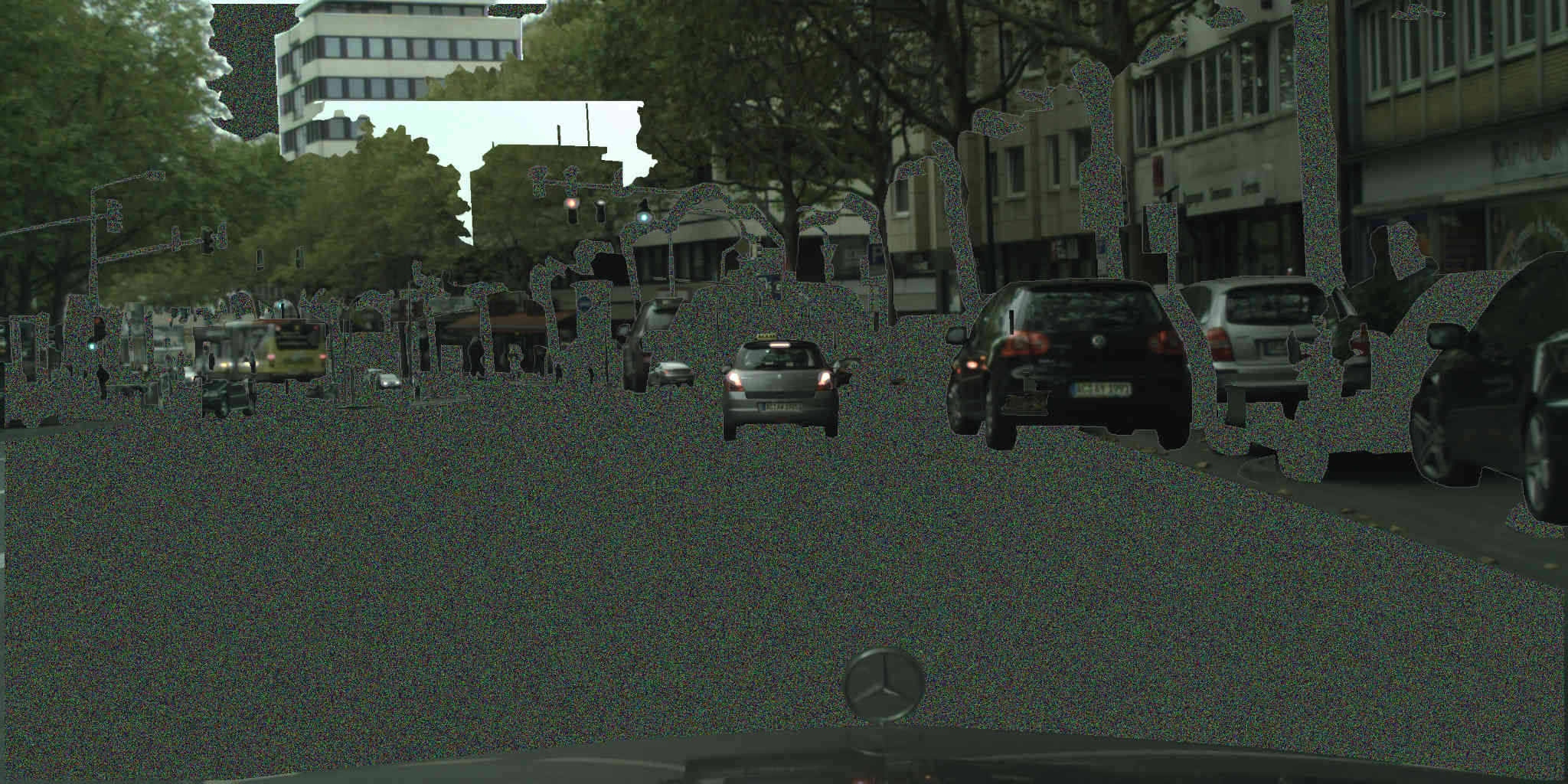} \\
(a) Original & (b) Dropout only & (c) Dropout+Resize+Shift\\
\end{tabular}
\end{center}
   \caption{\textbf{Examples of image with decreasing levels realism} (a) Original image. (b) Image with dropout only. The dominant segment - road is dropped out to balance per class pixel count. The composition of the picture is largely preserved and many objects still appear in realistic contexts. (c) Image with dropout, resize, and shift. In addition to dropout as seen in (b), segments are resized and moved to create new contexts. The additional operations make the new image unrealistic}
\label{fig:realism}
\end{figure}

We conducted an ablation study to evaluate the effectiveness of the individual operations of PanDA. We use the reproduced UPSNet-50 model trained on the original Cityscapes training set as the baseline (first row in Table \ref{table:ablation}). For each experiment group, certain subsets of PanDA image operations are ablated. We optimized PanDA's parameters under the ablation constraint and report the best performance obtained. Experiment groups are trained on their respective PanDA 1x augmented Cityscapes dataset. Doubling the number of training epochs without any augmented data (second row in Table \ref{table:ablation}) does not improve model performance which suggests that the original UPSNet training parameters are near optimal. PanDA with dropout operation only improves model performance, presumably by mitigating the data imbalance issue (Supplementary Fig \ref{fig:stats_1} and \ref{fig:stats_2}). Without resize or shifting, original object relationships are preserved and thus still realistic (Fig \ref{fig:realism}(b)). As more operations are included in PanDA, the level of realism of synthesized images decreases (Fig \ref{fig:realism}(c)). However, the fact that best performance is achieved with the least realistic image set suggest that, contrary to popular belief, the level of realism of synthetic images is not necessarily correlated with data augmentation performance.

\section{Discussion}

Compared with GAN based and 3D simulation based data augmentation methods, PanDA has several advantages. First of all, PanDA does not require training. GAN based methods need to be trained to generate realistic images in the desired domain first before the generative network can be used for data augmentation. Second, no additional data is needed for PanDA. As mentioned before, training GANs not only takes time but also usually requires a large unlabelled dataset from the same domain. 3D simulation based methods almost always require hand crafted 3D models of classes of interest. Finally, it is computationally very cheap to use PanDA since it operates exclusively in pixel space. Many image operations such as cropping and resizing could be further parallelized and optimized which allows for real time image synthesis at training time in an active learning fashion. 

As mentioned in the Related Work section, InstaBoost \cite{fang2019instaboost} relies on two operations - shift of instance segments and in-paint of background - to boost the performance of instance segmentation and detection. There are several key differences between PanDA and InstaBoost. First, in the panoptic segmentation task, a large part of the ``background'' in the instance segmentation task now belongs to the \textit{stuff} superclass. The ``background'' in the panoptic segmentation images usually takes up less than 15\% of the image, making it infeasible to use the same in-paint process. Secondly, the motivation behind in-painting the background is not explicitly explained and presumably to maintain the image's level of realism. In contrast, PanDA boosts panoptic segmentation performance despite losing image realism. Finally, in addition to the shift operation in InstaBoost, PanDA has resize and drop out operations which are shown to be essential for its effectiveness by the ablation study (Table \ref{table:ablation}). The method proposed by Dvornik et al. \cite{dvornik2019importance} adds new instances to original images and avoids holes in the background from removing instances as seen in InstaBoost. However, the pasting of new instances has to be guided by a CNN-based context model to determine the location and class of instances to add, and that model has to be trained which requires data and time. Similar to InstaBoost, the method only applies to \textit{things} classes which takes less than 50\% of pixels in an image on average. Shetty et al. \cite{shetty2019not} propose removal of objects to break context information and help deep models to generalize in classification and semantic segmentation. This approach relies heavily on a trainable CNN based in-paint network to fill the holes in the background. Data and time are needed to train the in-paint CNN network in this method. Additionally, experiments in the original paper show that removing large objects such as mountains makes it hard for the in-paint network to fill the hole left behind and indeed hurts model performance. In contrast, PanDA removes the large objects like road and buildings with high probabilities and we show that is essential for performance gain.

There are certain limitations within the current implementation of PanDA that leave doors open for more exploration. First, the functions used in background padding, dropout, resize, and shift can be further optimized or complemented with more optimized functions. For example, resizing can be drawn from a normal instead of uniform distribution. Operations such as rotation and warping can be added. Further optimizations can be performed in an automated way \cite{cubuk2018autoaugment}. Second, since the 3,000 training images in Cityscapes can be divided into 78,000 segments, if we allow the creation of hybrid images where segments in the new image can come from different original images, we could further expand the image variation combinatorially. Third, it remains unclear how many useful synthetic images can be generated with PanDA per original image. In principle, one could create an infinite number of synthetic images from the original dataset. Due to the high computation cost of training with very large datasets, this paper limits the scope to doubling the number of training images. It will be an interesting direction to investigate the limit of the number of useful synthetic images per original image.

It is worth noting that for challenges on ego-centric driving datasets like Cityscapes, it is also possible to improve model performance by pretraining on a larger dataset from a related domain. For example, better performing models can be obtained by pretraining on MS COCO and fine tuning with Cityscapes\cite{xiong2019upsnet}. However, in a new and specialized domain, there might not exist a dataset available to pretrain on. In addition, for a large dataset like MS COCO, pretraining on smaller datasets is unlikely to help.

Finally, although PanDA operations are well justified from the standpoint of statistics, we were surprised that the best performing PanDA augmented datasets do not look natural or realistic to human eyes at all. Many PanDA generated images look qualitatively similar to the ones shown in Fig \ref{fig:examples} where positioning and occlusion of objects are not realistic. Many objects in the synthetic images appear to be ``floating" on top of the noise background out of the original context (Fig \ref{fig:examples} (d), Fig \ref{fig:realism} (c)); sometimes they cluster and overlap each other in the wrong depth order (Fig \ref{fig:examples} (f)). It is known that contextual information helps human visual detection\cite{neider2006scene}. We suspect that by taking objects out of their original context, PanDA presents harder challenges to the model and therefore forces the model to pay more attention to the pixels within the object foreground. Secondly, despite the fact that PanDA drastically reduced the total non-background pixels per synthetic image, the augmented datasets are more balanced. It suggests that a balanced but small dataset might be more helpful than a large but unbalanced dataset.

\section{Conclusions}

We present a simple and efficient method for data augmentation of annotated panoptic images to improve panoptic segmentation performance. PanDA is computationally cheap, and requires no training or additional data. After training with PanDA augmented datasets, top performing panoptic segmentation models further improve performance on two popular datasets, Cityscapes and MS COCO. To the best of our knowledge, PanDA is the first pixel-space data augmentation method with demonstrated performance gain for leading models on panoptic segmentation tasks. Further improvement is possible with fine-tuned parameters. The effectiveness of unrealistic images suggests that we should reconsider maximizing realism in image synthesis for data augmentation. Finally, PanDA opens new opportunities for exploring efficient pixel space data augmentation methods for detection and segmentation datasets, and we believe the community would benefit from these explorations in data augmentation. Code will be available on GitHub upon publication.

\clearpage
%
%
\bibliographystyle{splncs04}
\bibliography{main}

\clearpage

\setcounter{figure}{0}
\setcounter{table}{0}

    \centering
    \vfill
    {\Large
        \textbf{PanDA: Panoptic Data Augmentation - Appendix}\\
    }


\begin{figure}[h]
\begin{center}
\includegraphics[width=0.8\linewidth]{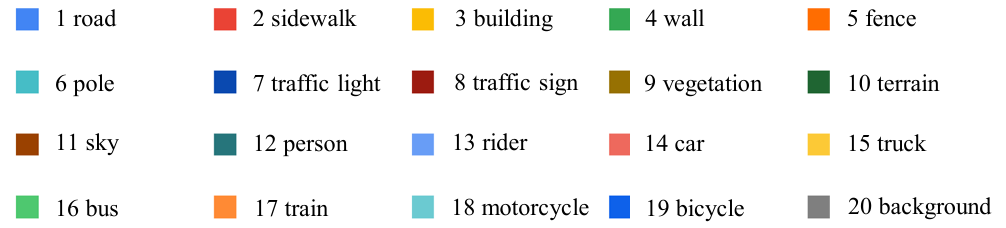} \\
\includegraphics[width=0.8\linewidth]{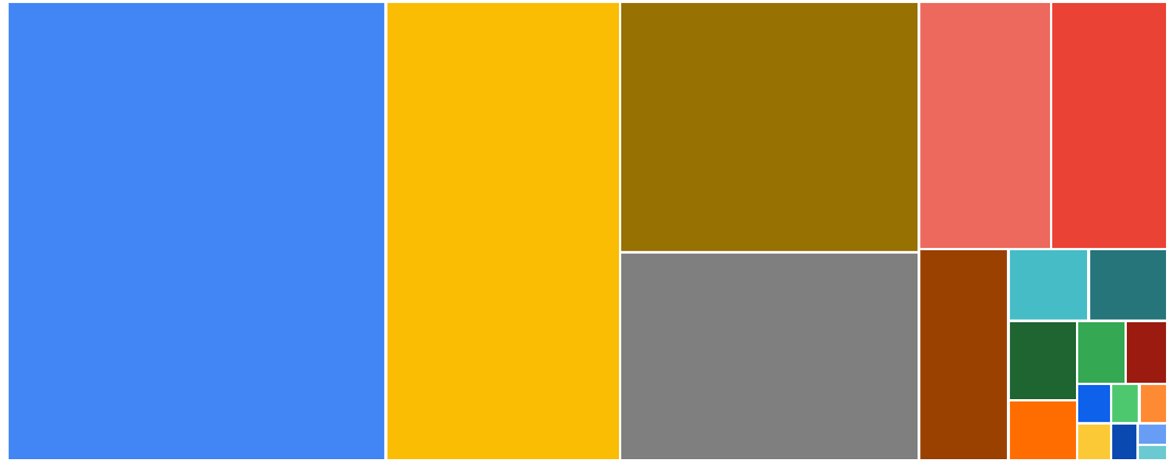} \\
Original Cityscapes \\ 
\includegraphics[width=0.8\linewidth]{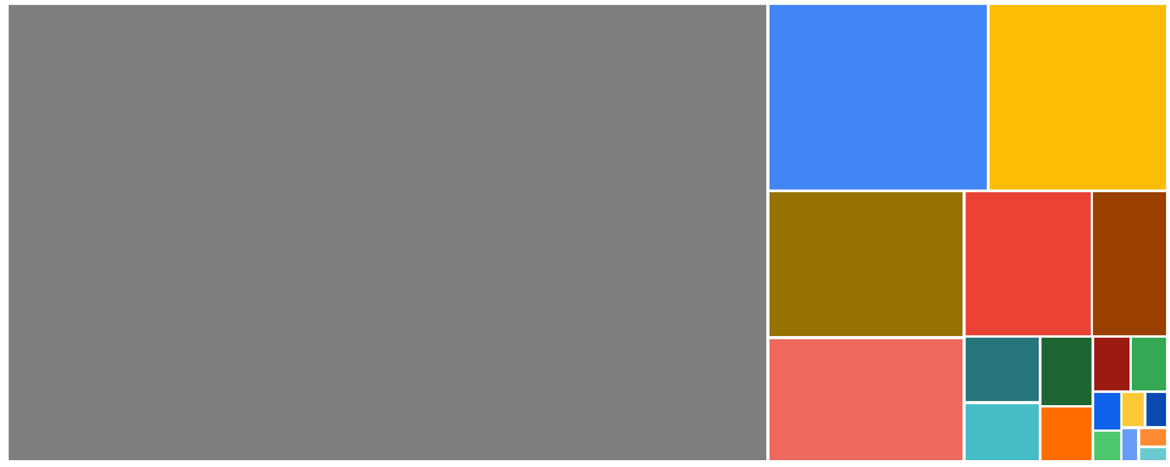} \\
PanDA Cityscapes \\
\end{center}
\caption{\textbf{Treemap views of average pixel count by class per image of Cityscapes training set.}  Class 20 is the background which is considered out of area of interest. Classes 1 and 3 together occupy the majority of the non-background pixels of an image in the original Cityscapes, and make up less than half the non-background pixels. }
\label{fig:stats_1}
\end{figure}

\begin{figure}[h]
\begin{center}
\includegraphics[width=1\linewidth]{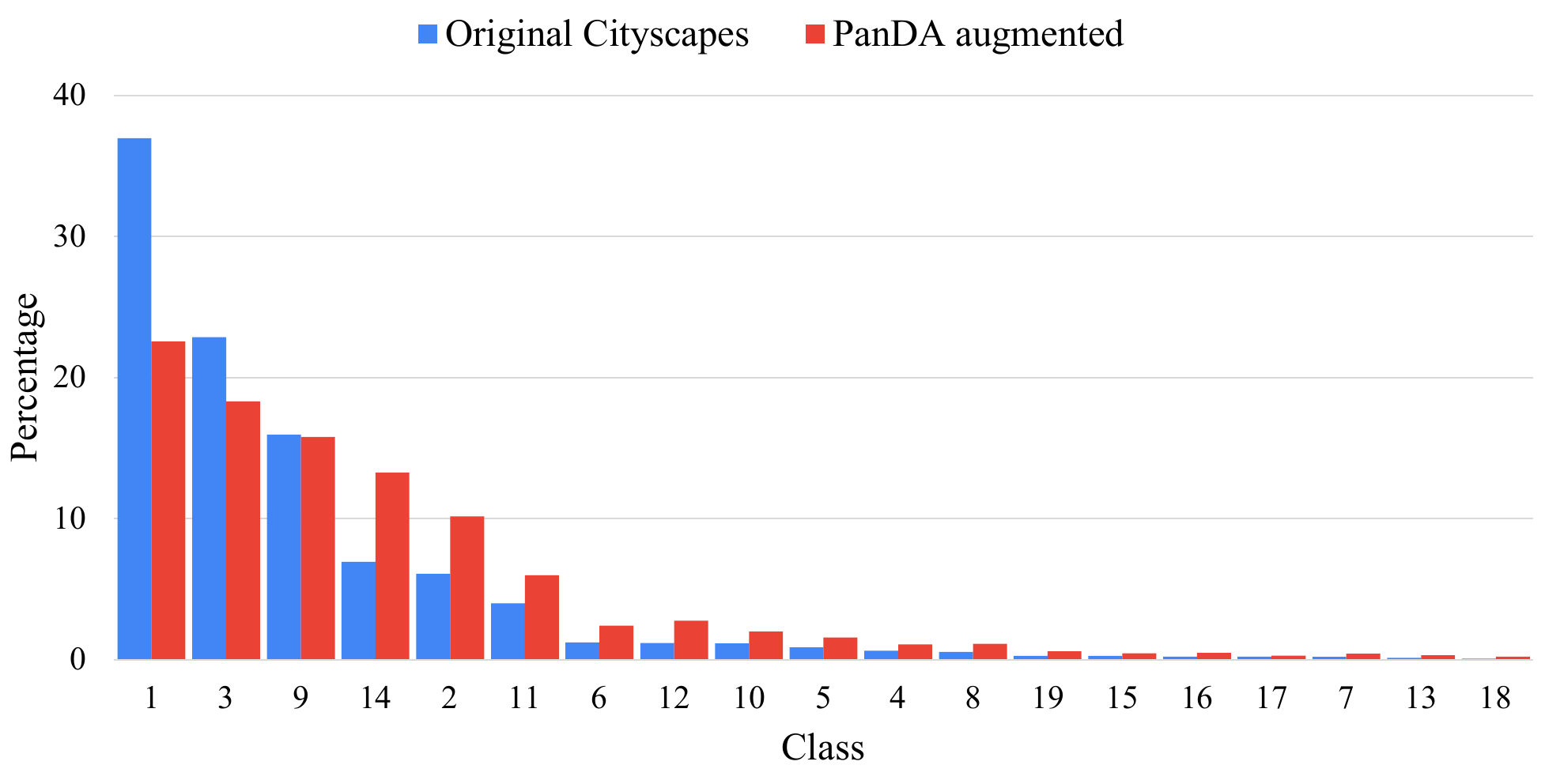}
\end{center}
\caption{\textbf{Per class pixel percentage per image.} Bar graph of average per class pixel percentage of non-background classes per image. Each bar is computed by dividing the average number of pixels of a given class by the sum of the average number of non-background pixels. Pixel percentages of common classes are reduced and those of rare classes are increased, making the synthetic images more class-balanced.}
\label{fig:stats_2}
\end{figure}

\begin{figure}[h]
\begin{center}
\includegraphics[width=0.9\linewidth]{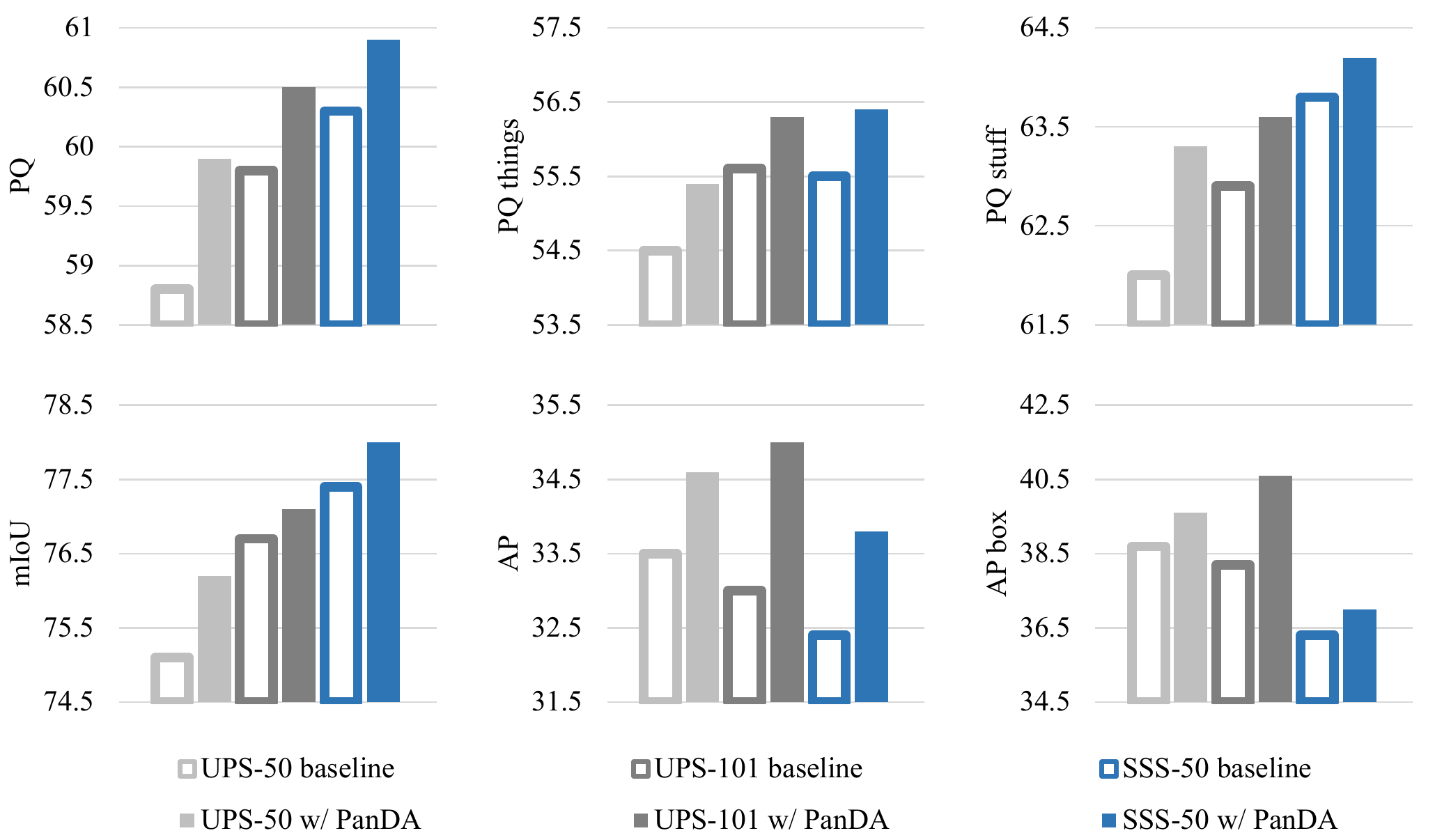} 

\end{center}
   \caption{\textbf{Results on Cityscapes} In each case data augmentation with PanDA improves performance. \textit{PQ}: panoptic quality. \textit{PQ things}: PQ for \textit{things} classes. \textit{PQ stuff}: PQ for \textit{stuff} classes. \textit{mIoU}: mean intersection over union. \textit{AP}: segmentation average precision evaluated at 0.5:0.95. \textit{AP$^{box}$}: bounding box average precision evaluated at 0.5:0.95}
\label{fig:main_results}
\end{figure}

\begin{table}[h]
\begin{center}
\caption{\textbf{Extended Results on Cityscapes.} Despite the fact that our baseline performance (best of 5 runs) is lower than that reported in the original UPSNet paper \cite{xiong2019upsnet}, our models trained on PanDA augmented datasets outperform the original UPSNet-50 model without COCO pretraining in all metrics. PanDA 1x consists of 3,000 original Cityscapes images plus 3,000 PanDA synthetic images, PanDA 2x consists of 3,000 original Cityscapes images plus 6,000 PanDA images. PQ: panoptic quality. SQ: segmentation quality. RQ: recognition quality. \textit{PQ$^{th}$}: PQ for \textit{things} classes. \textit{PQ$^{st}$}: PQ for \textit{stuff} classes. \textit{mIoU}: mean intersection over union. \textit{AP}: segmentation average precision evaluated at 0.5:0.95. \textit{AP$^{box}$}: bounding box average precision evaluated at 0.5:0.95}
\label{table:performance}
\begin{tabular}{l|c c c|c c c c c}
\hline
 Model & PQ & SQ & RQ & PQ$^{Th}$ & PQ$^{St}$ & mIoU & AP & AP$^{Box}$ \\
\hline\hline
UPSNet-50 \textit{Xiong et al.}\cite{xiong2019upsnet} & 59.3 & 79.7 & 73.0 & 54.6 & 62.7 & 75.2 & 33.3 & 39.1\\
\hline
UPSNet-50 baseline & 58.8 & 79.5 & 72.6 & 54.5 & 62.0 & 75.1 & 33.5 & 38.7 \\
\hline
UPSNet-50 w/ PanDA 1x & 59.9 & 79.9 & \bf73.8 & 55.4 & \bf63.3 & \bf76.2 & 34.6 & 39.6 \\
\hline
UPSNet-50 w/ PanDA 2x & \bf60.0 & \bf80.3 & 73.5 & \bf55.8 & 63.1 & \bf76.2 & \bf35.6 & \bf40.5 \\
\hline
\hline
UPSNet-101 baseline & 59.8 & 80.0 & 73.5 & 55.6 & 62.9 & 76.7 & 33.0 & 38.2 \\
\hline
UPSNet-101 w/ PanDA 1x & \bf60.5 & \bf80.2 & \bf74.1 & \bf56.3 & \bf63.6 & \bf77.1 & \bf35.0 & \bf40.6 \\
\hline
\hline
SSS-50 \textit{Porzi et al.} \cite{porzi2019seamless} & 60.3 & -- & -- & 56.1 & 63.3 & 77.5 & 33.5 & -- \\
\hline
SSS-50 baseline & 60.3 & -- & -- & 55.5 & 63.8 & 77.4 & 32.4 & 36.3 \\
\hline
SSS-50 w/ PanDA 1x & \bf60.9 & -- & -- & \bf56.4 & \bf64.2 & \bf78.0 & \bf33.8 & \bf37.0 \\
\hline
\end{tabular}
\end{center}
\end{table}

\begin{table}[h]
\begin{center}
\caption{\textbf{Per class instance segmentation results on Cityscapes.} Segmentation AP are reported. We observe not only large relative improvement on rare classes such as train and bicycle (18.4\% and 13.1\%, respectively), but also small gains on common classes such as car and person (3.5\% and 3.5\%, respectively)}
\label{table:instance_performance}
\begin{tabular}{l|c c c c c c c c}
\hline
 Models & person & rider & car & truck & bus & train & motorcycle & bicycle \\
\hline\hline
UPSNet-50 \textit{Xiong et al.}\cite{xiong2019upsnet} & 31.2 & 25.1 & 50.9 & 33.6 & 55.0 & 33.2 & 19.2 & 18.3\\
UPSNet-50 our basline & 31.1 & 24.8 & 51.0 & 33.2 & 53.5 & 35.8 & 19.4 & 18.9 \\
UPSNet-50 w/ PanDA 1x & 31.5 & \bf25.7 & 52.2 & 34.3 & 54.4 & 37.9 & \bf20.8 & 20.3 \\
UPSNet-50 w/ PanDA 2x & \bf32.3 & 25.6 & \bf52.7 & \bf36.9 & \bf56.6 & \bf39.3 & 20.6 & \bf20.7 \\
\hline
\end{tabular}
\end{center}
\end{table}
\begin{table}[h]
\begin{center}
\caption{\textbf{Per class detection results on Cityscapes.} AP box are reported}
\label{table:detection_performance}
\begin{tabular}{l|c c c c c c c c}
\hline
 Models & person & rider & car & truck & bus & train & motorcycle & bicycle \\
\hline\hline
UPSNet-50 \textit{Xiong et al.}\cite{xiong2019upsnet} & 38.6 & 42.1 & 56.6 & 33.4 & 56.3 & 27.2 & \bf28.4 & 30.1\\
UPSNet-50 our basline & 38.5 & 41.6 & 56.3 & 33.6 & 55.4 & 26.0 & 27.1 & 31.2 \\
UPSNet-50 w/ PanDA 1x & 39.0 & 43.4 & 57.8 & 33.6 & 55.8 & 27.2 & \bf28.4 & 31.2 \\
UPSNet-50 w/ PanDA 2x & \bf40.1 & \bf43.4 & \bf58.6 & \bf36.3 & \bf57.5 & \bf29.3 & 27.4 & \bf31.7 \\
\hline
\end{tabular}
\end{center}
\end{table}

\begin{figure}[h]
\begin{center}
\includegraphics[width=0.9\linewidth]{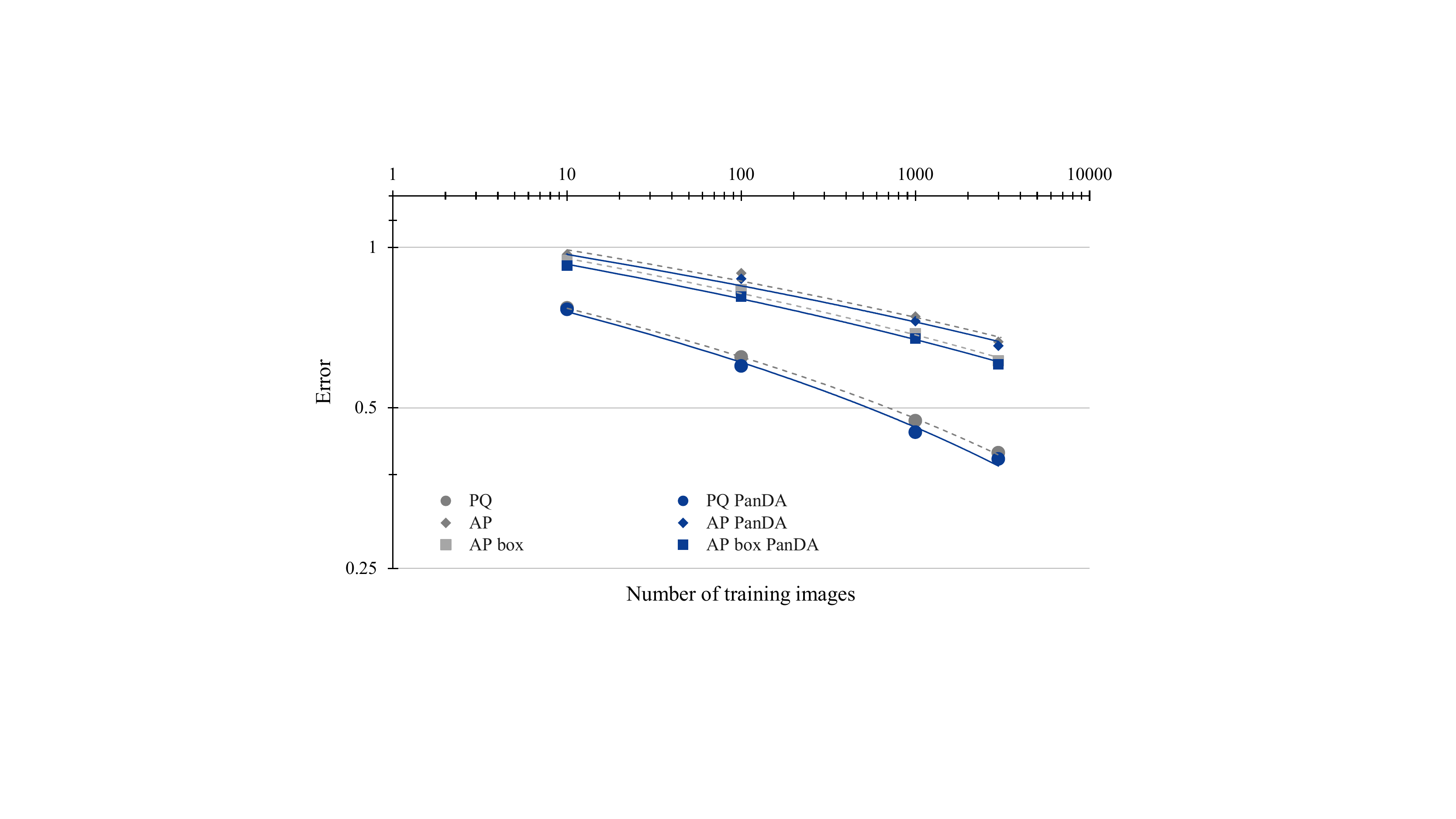}
\end{center}
   \caption{\textbf{Model performance vs training set size on Cityscapes.} We train UPSNet-50 models on various Cityscapes subsets ranging from 10 to 2,975 images. \textit{PQ}: panoptic quality. \textit{AP}: instance segmentation average precision evaluated at 0.5:0.95. \textit{AP box}: bounding box average precision evaluated at 0.5:0.95. Dashed curves are log-linear fits of baseline experiments, solid curves are fits of PanDA experiments. Panoptic segmentation, instance segmentation, and instance detection performance are summarized by PQ, AP, and AP box, respectively. PanDA enhanced models consistently outperform original models across scales in all metrics. Data efficiency (DE) corresponds to the amount of right shift of PanDA curves to account for the improved performance with PanDA }
\label{fig:performance_vs_number_supp}
\end{figure}

\begin{figure}[h]
\centering
\begin{center}
\centering
\addtolength{\tabcolsep}{-5pt} 
\begin{tabular}{c c c c }
    Original & Original annotation & PanDA & PanDA annotation \\
    
    {\includegraphics[width=0.25\textwidth]{figures/aachen_000000_000019_leftImg8bit.jpg}} & {\includegraphics[width=0.25\textwidth]{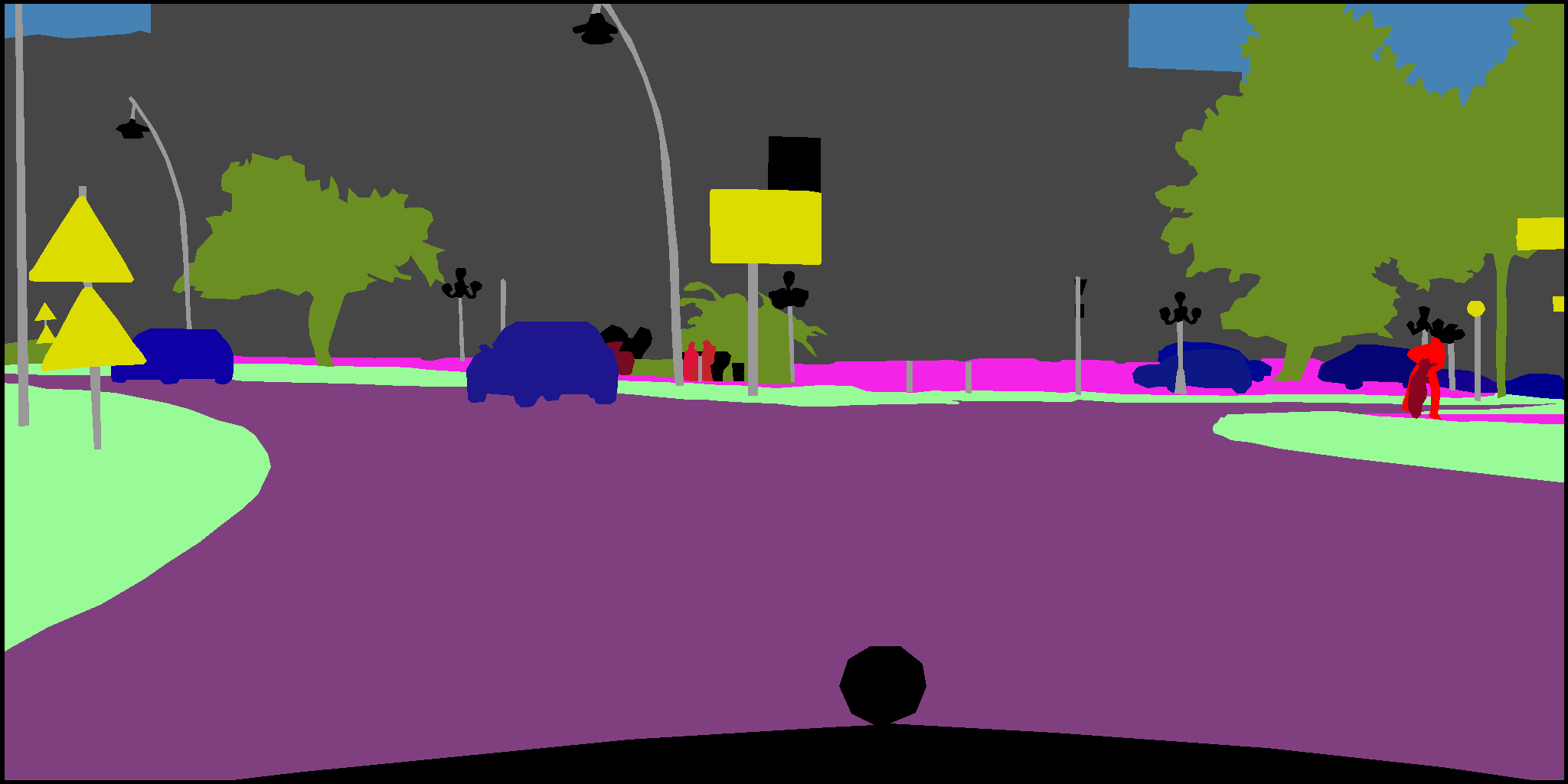}} & {\includegraphics[width=0.25\textwidth]{figures/aachen_500000_000019_leftImg8bit.jpg}} & {\includegraphics[width=0.25\textwidth]{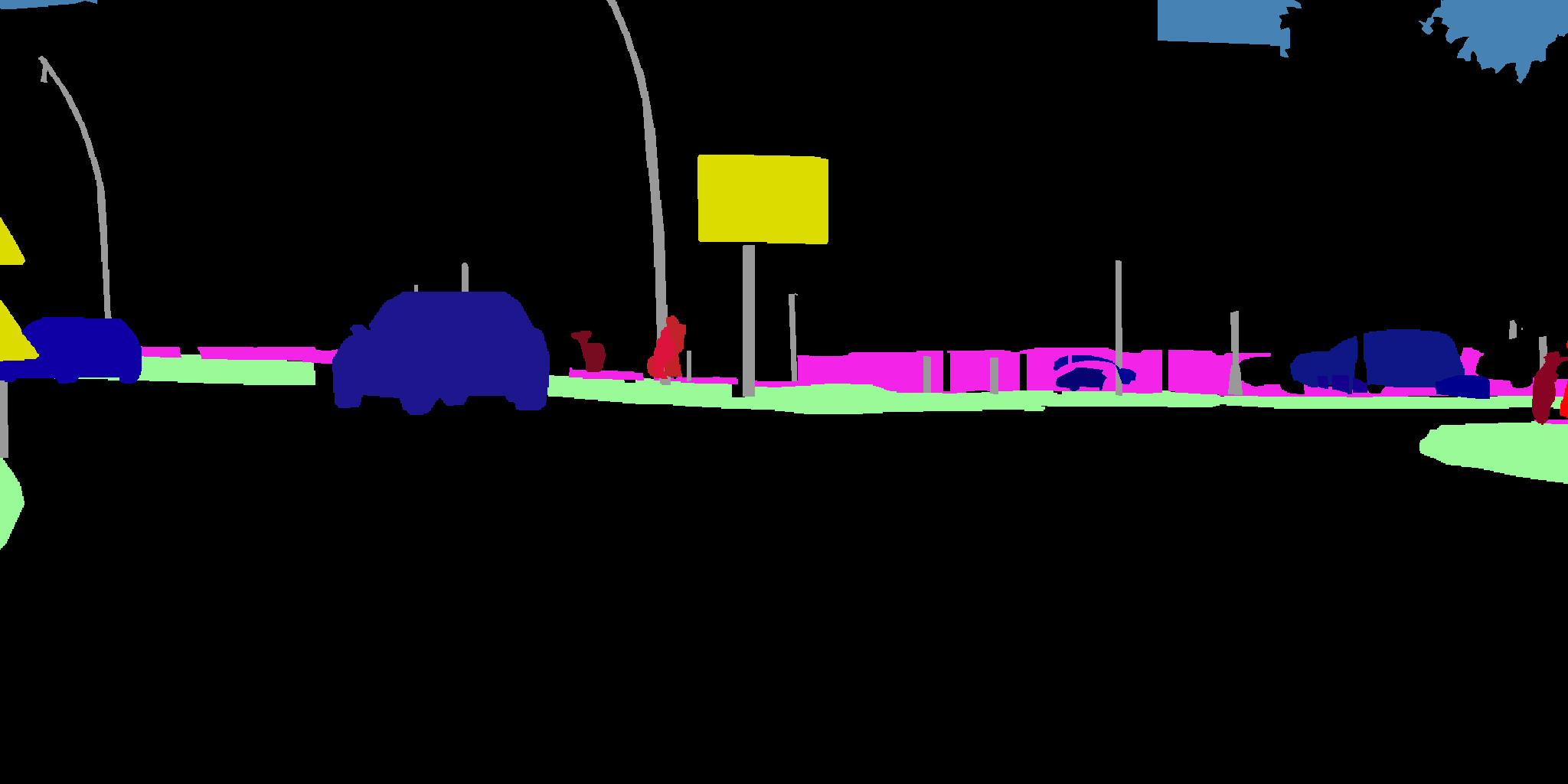}}\\
    
    {\includegraphics[width=0.25\textwidth]{figures/aachen_000001_000019_leftImg8bit.jpg}} & {\includegraphics[width=0.25\textwidth]{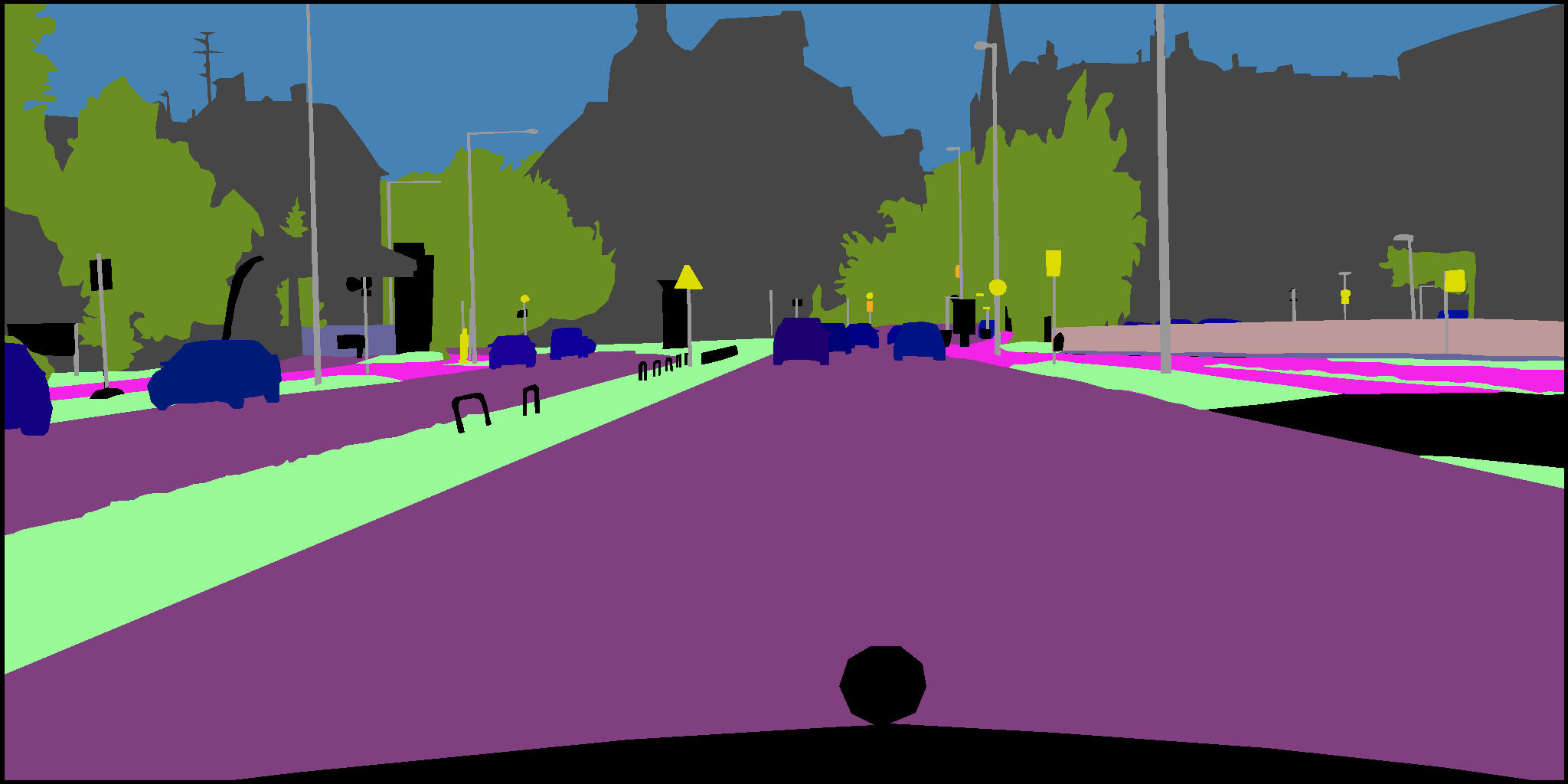}} & {\includegraphics[width=0.25\textwidth]{figures/aachen_500001_000019_leftImg8bit.jpg}} & {\includegraphics[width=0.25\textwidth]{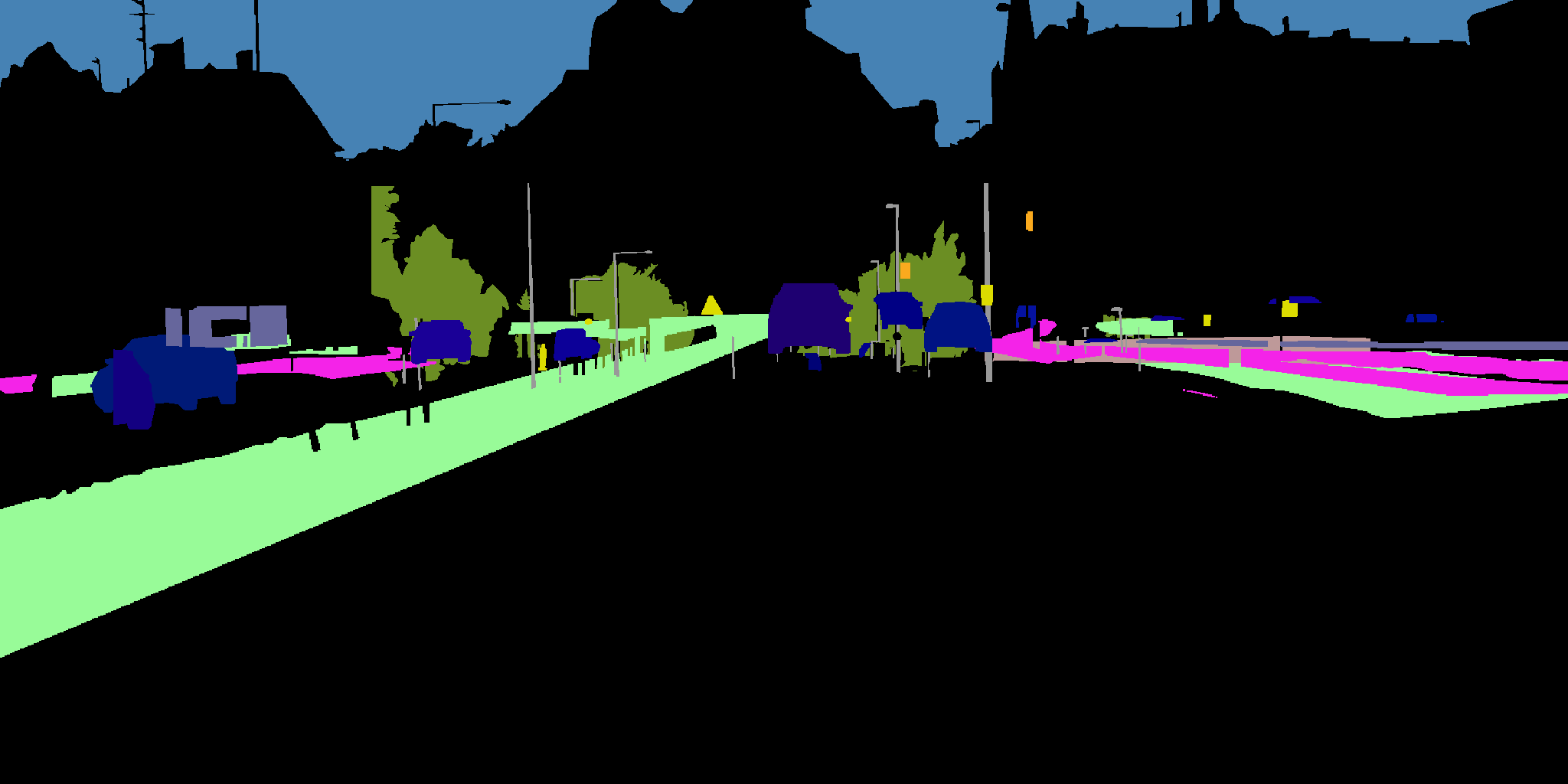}}\\
    
    {\includegraphics[width=0.25\textwidth]{figures/aachen_000002_000019_leftImg8bit.jpg}} & {\includegraphics[width=0.25\textwidth]{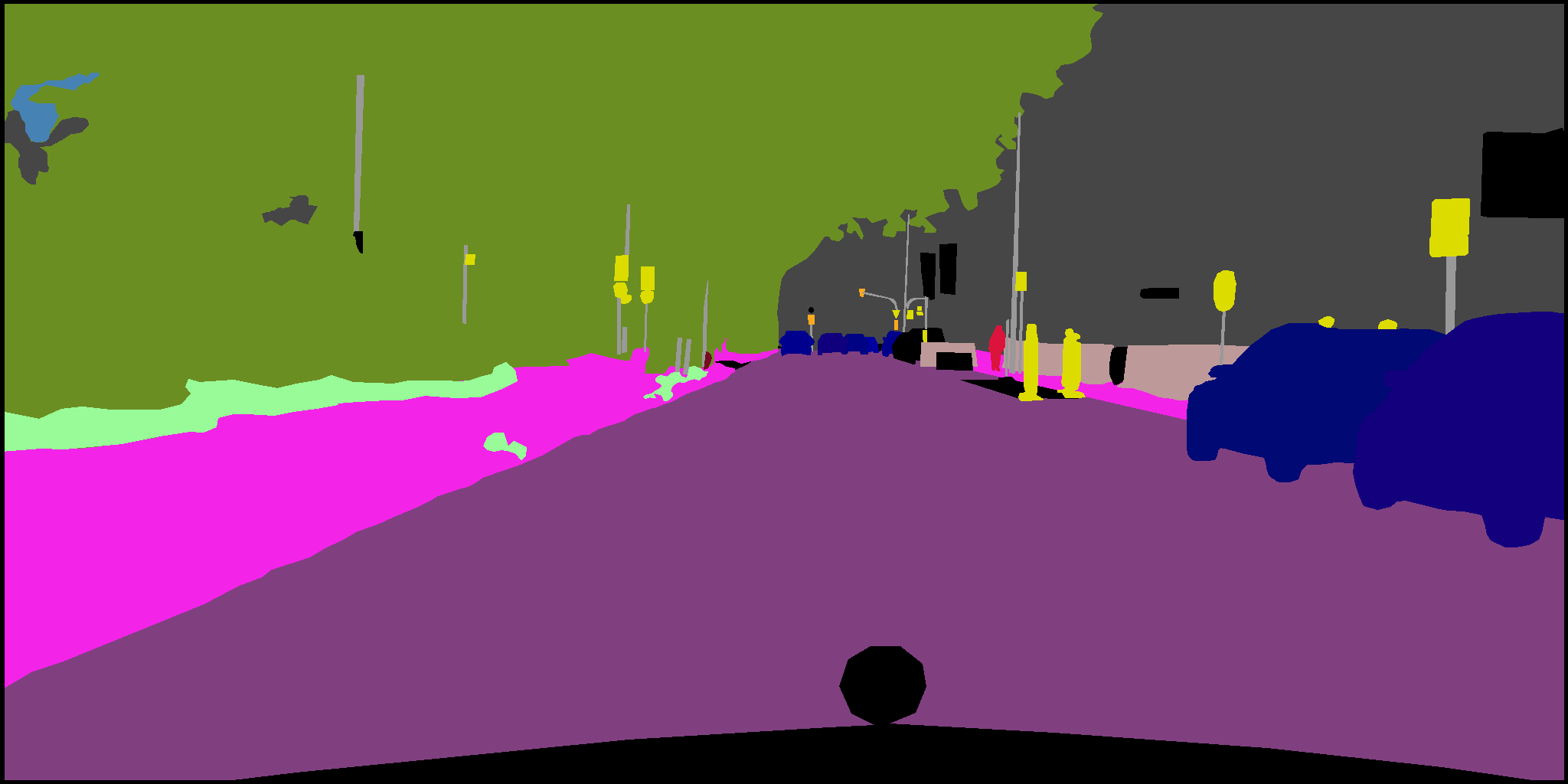}} & {\includegraphics[width=0.25\textwidth]{figures/aachen_500002_000019_leftImg8bit.jpg}} & {\includegraphics[width=0.25\textwidth]{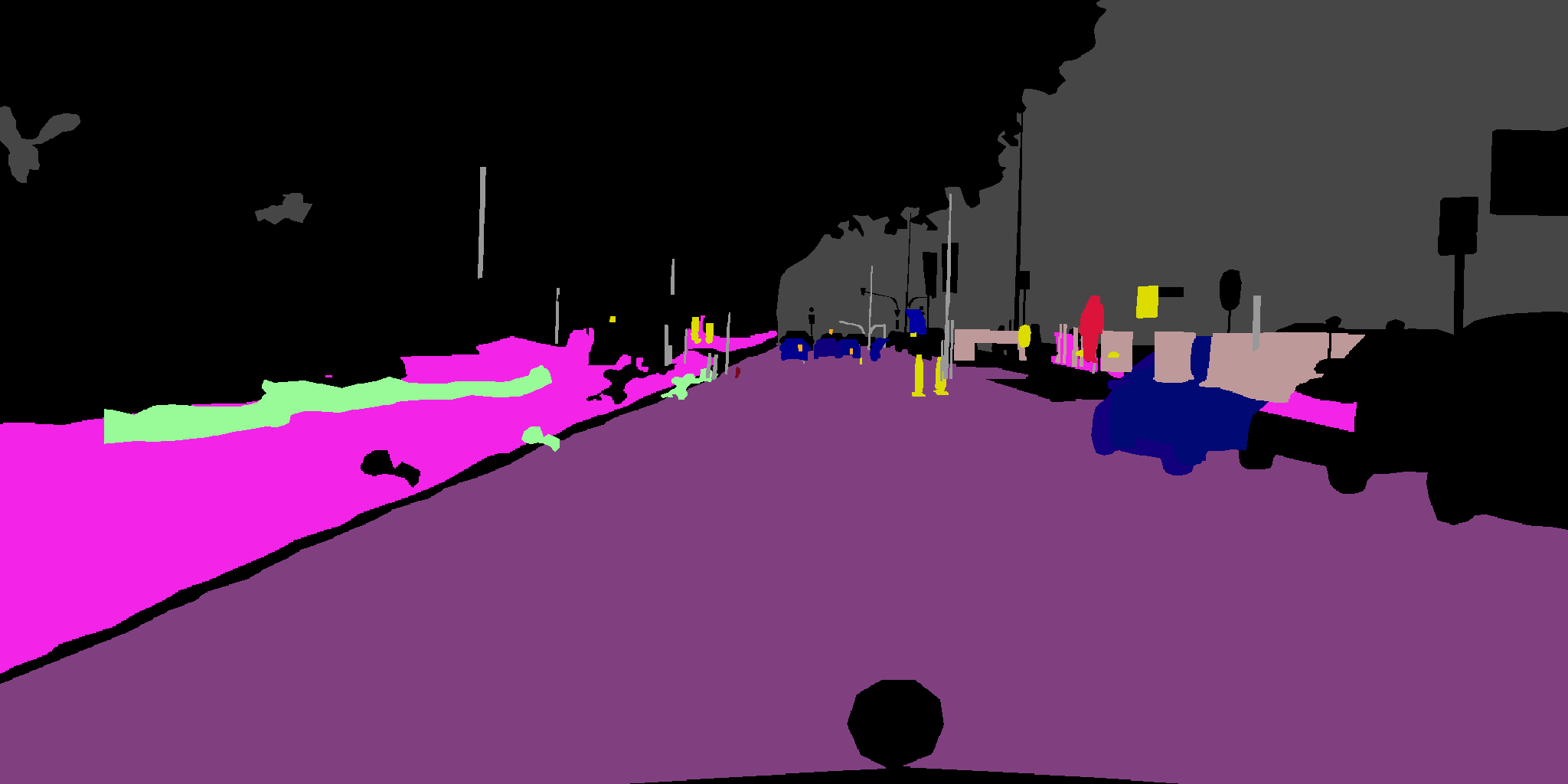}}\\
    
    {\includegraphics[width=0.25\textwidth]{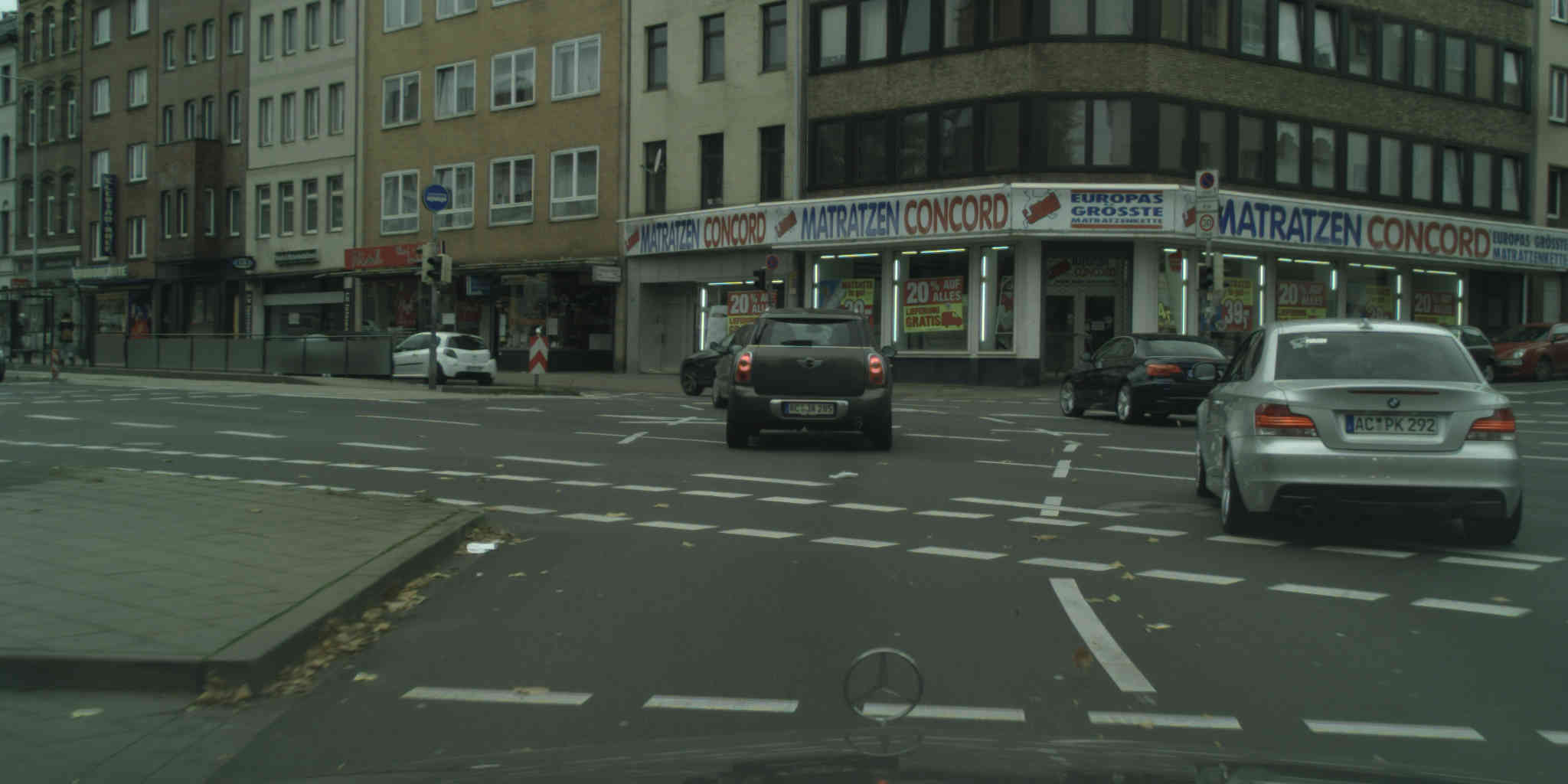}} & {\includegraphics[width=0.25\textwidth]{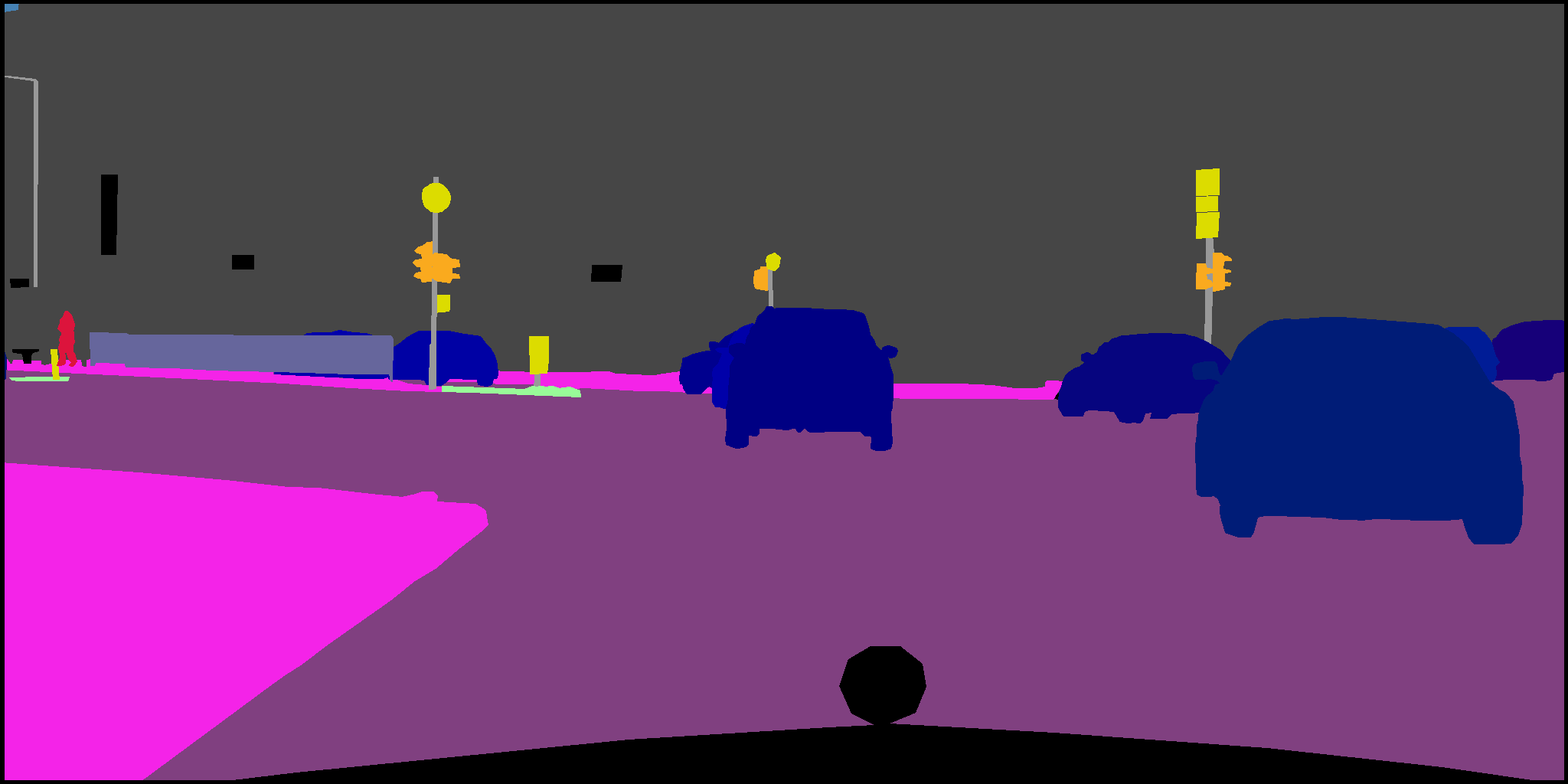}} & {\includegraphics[width=0.25\textwidth]{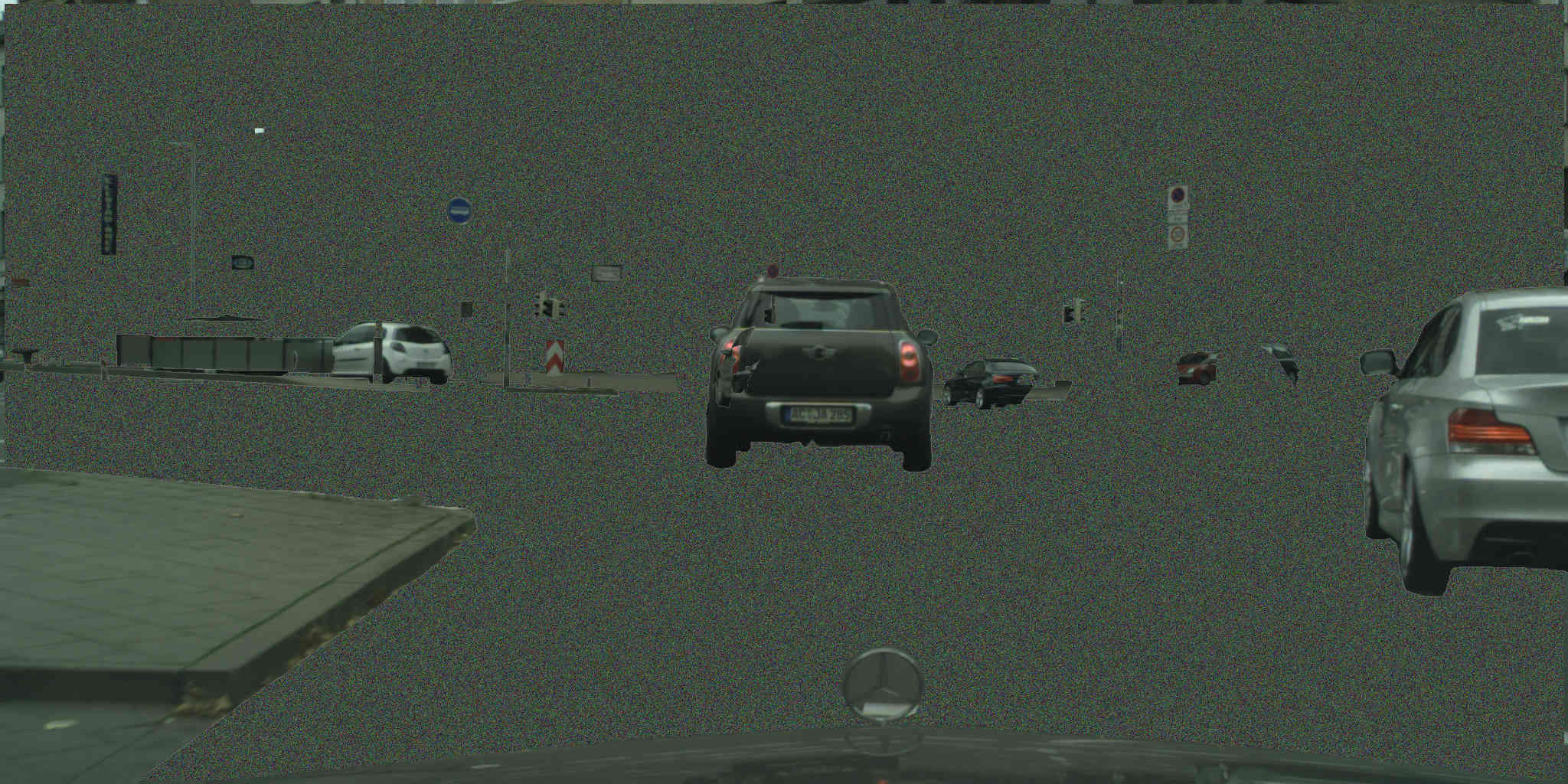}} & {\includegraphics[width=0.25\textwidth]{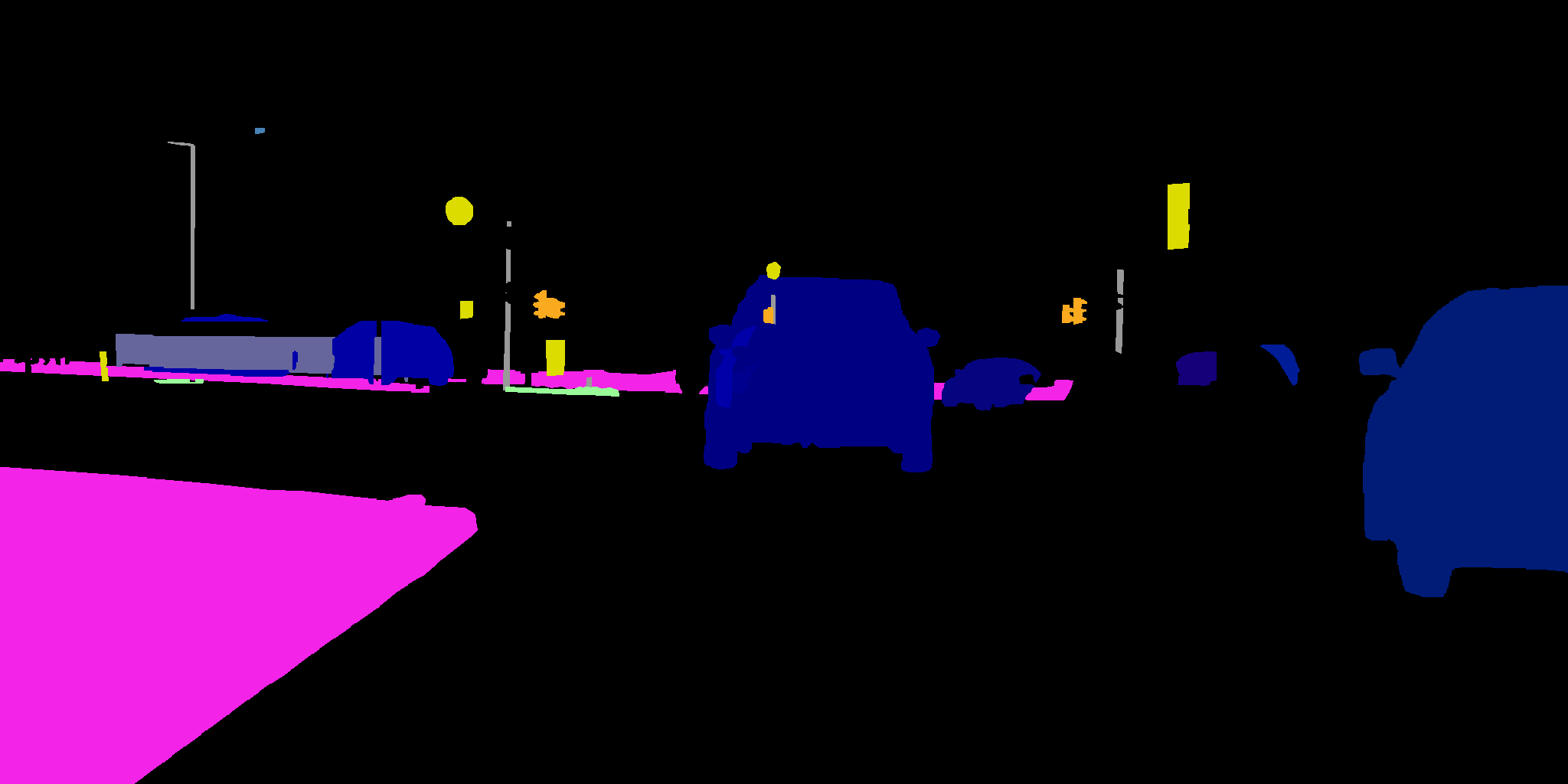}}\\
    
    {\includegraphics[width=0.25\textwidth]{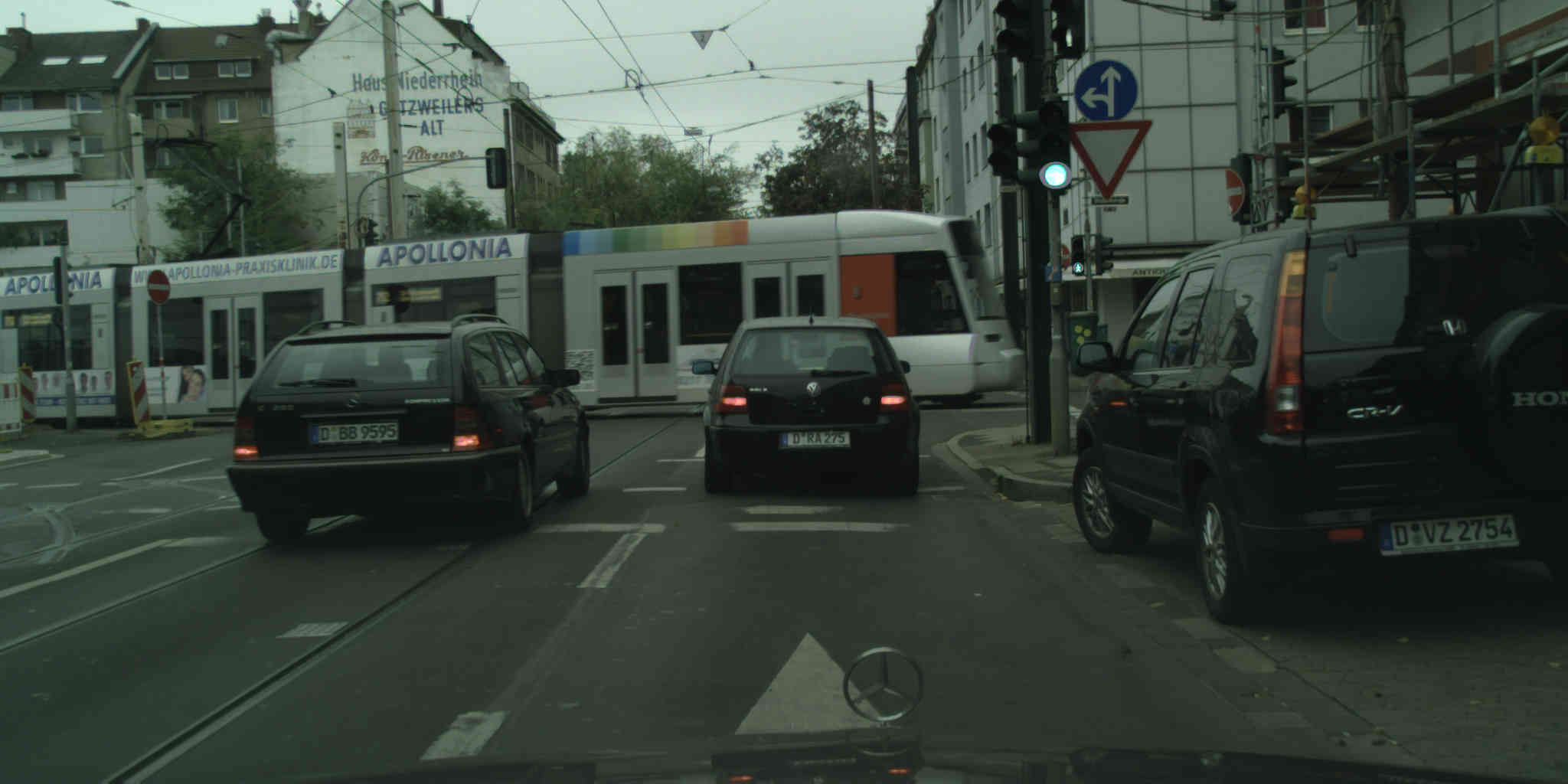}} & {\includegraphics[width=0.25\textwidth]{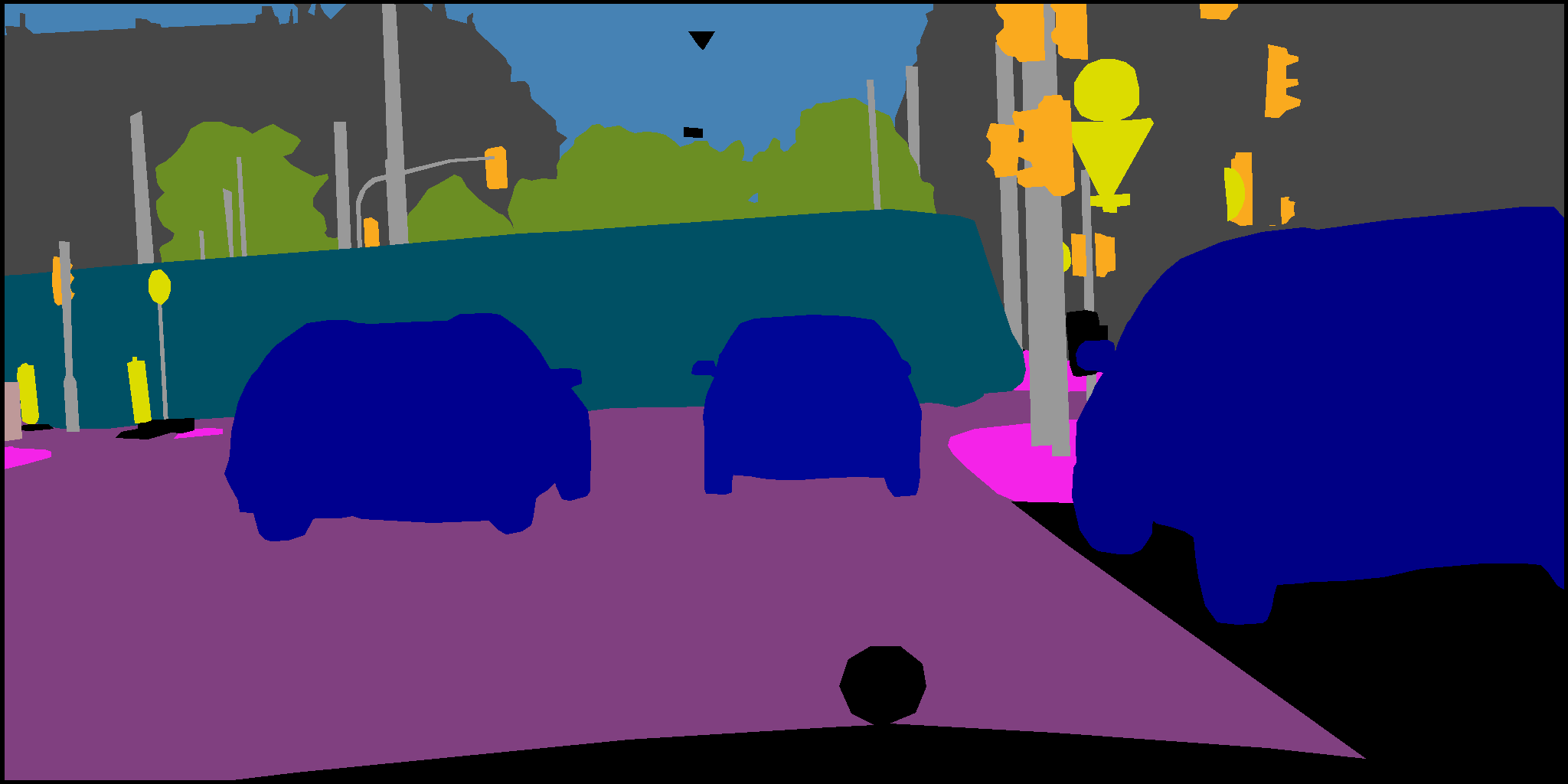}} & {\includegraphics[width=0.25\textwidth]{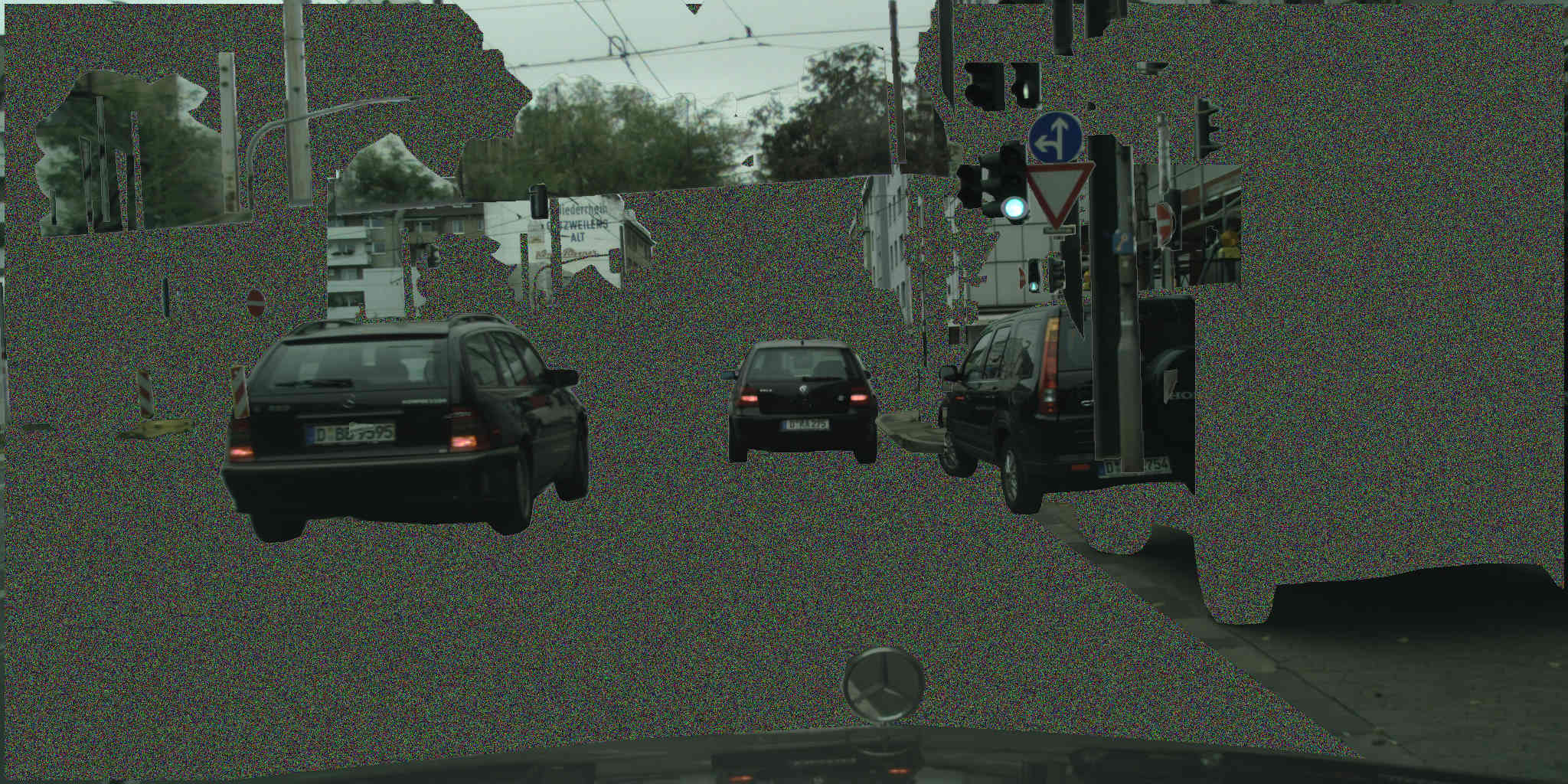}} & {\includegraphics[width=0.25\textwidth]{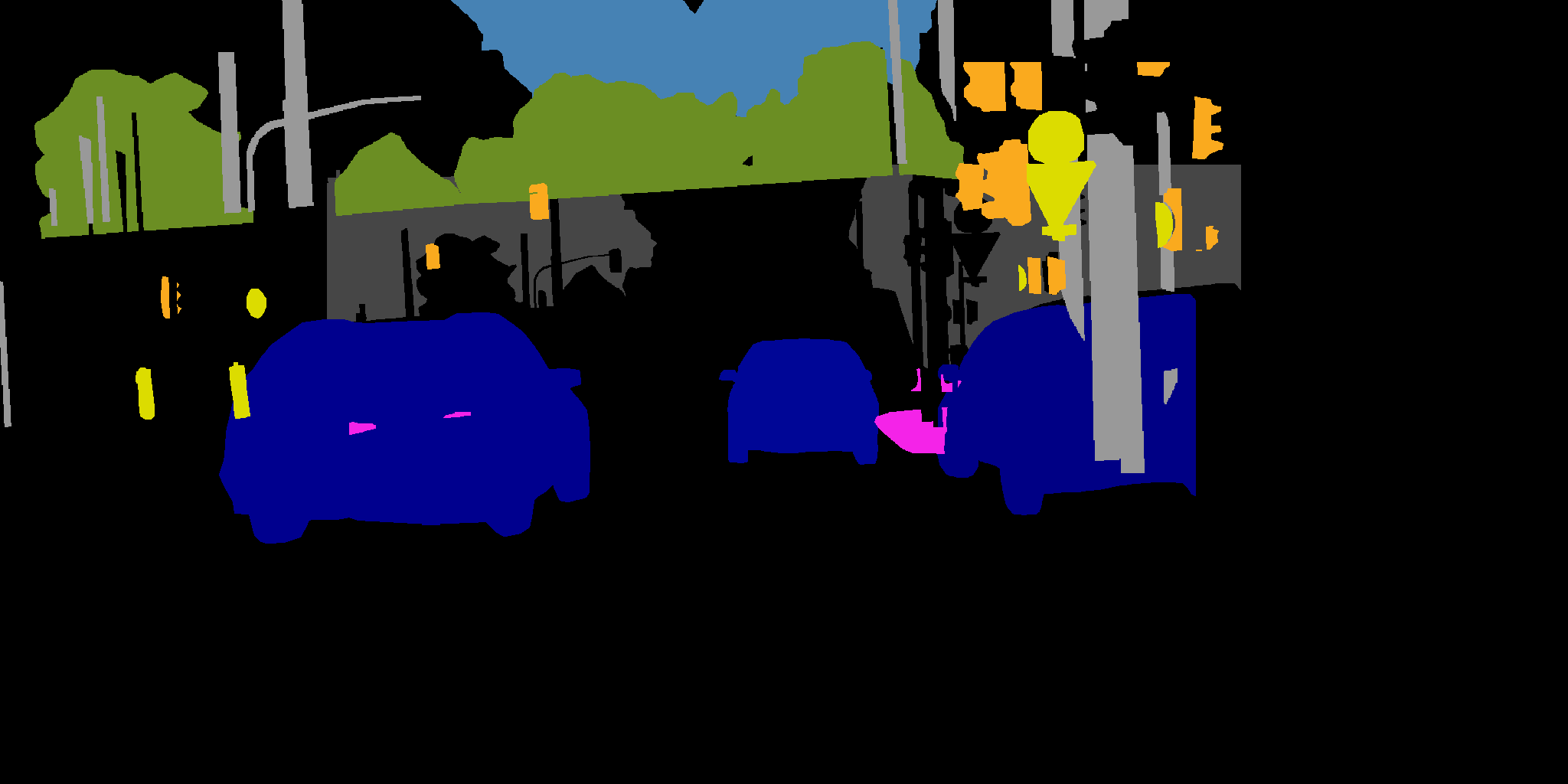}}\\
    
    {\includegraphics[width=0.25\textwidth]{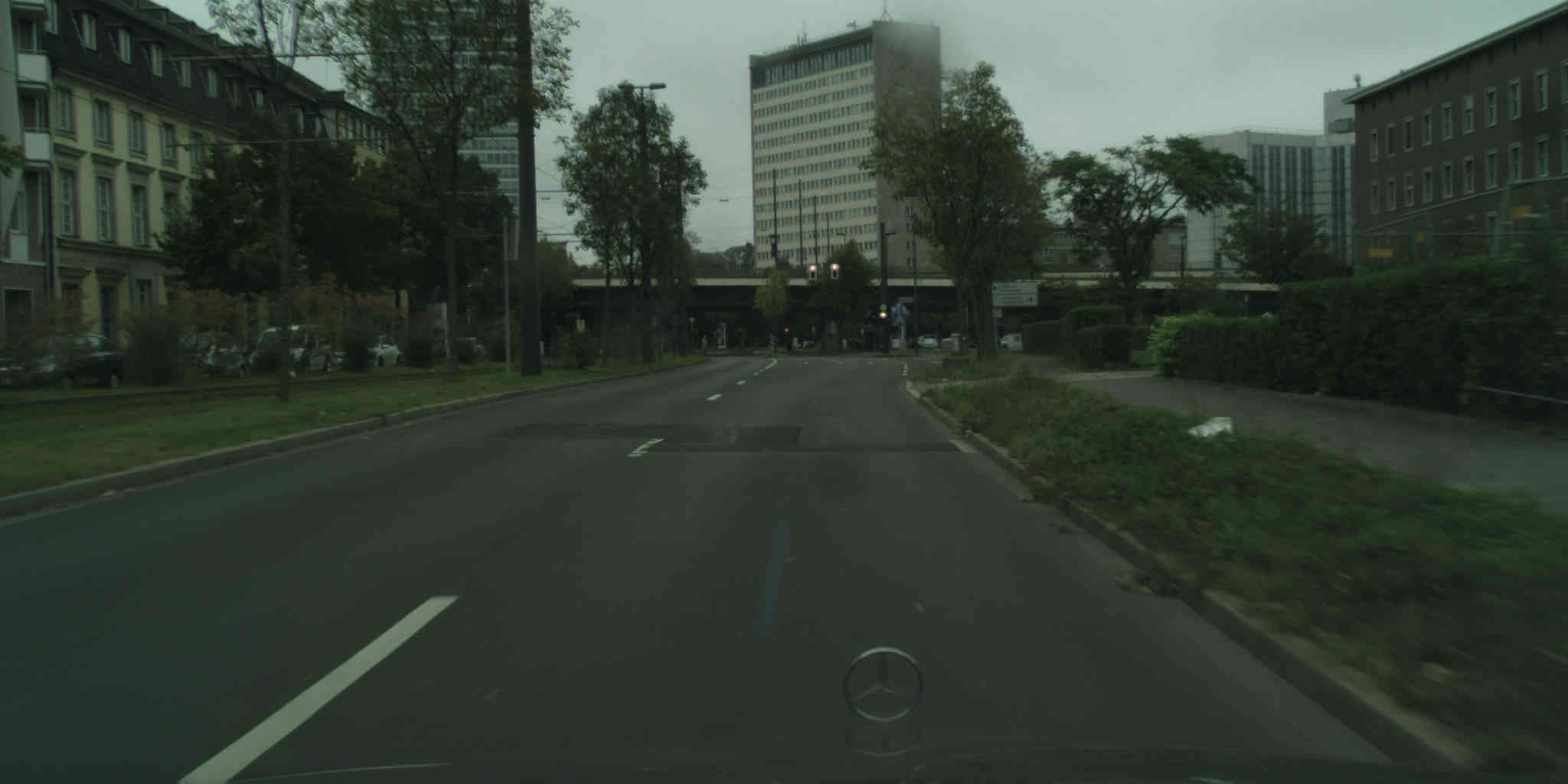}} & {\includegraphics[width=0.25\textwidth]{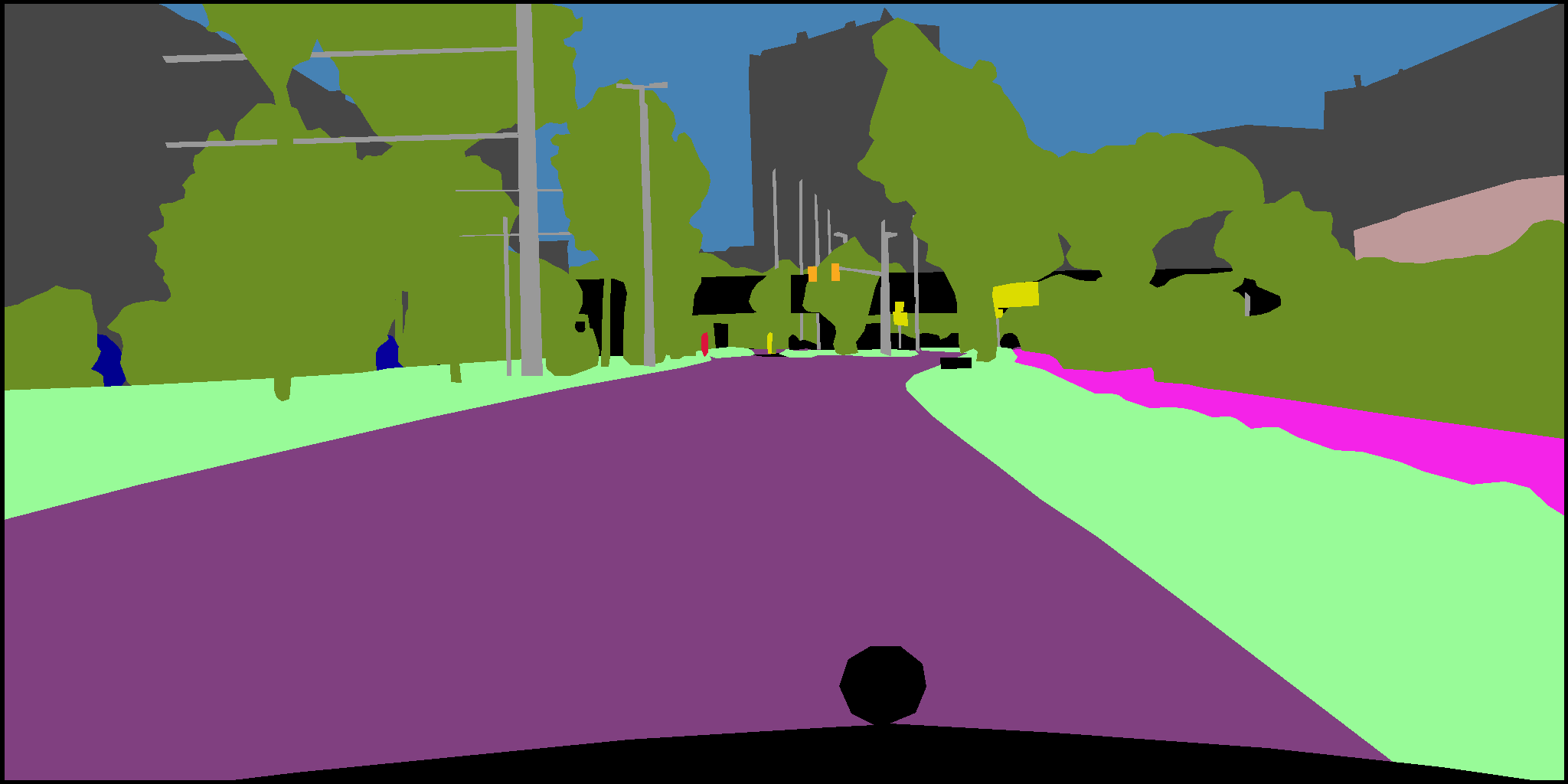}} & {\includegraphics[width=0.25\textwidth]{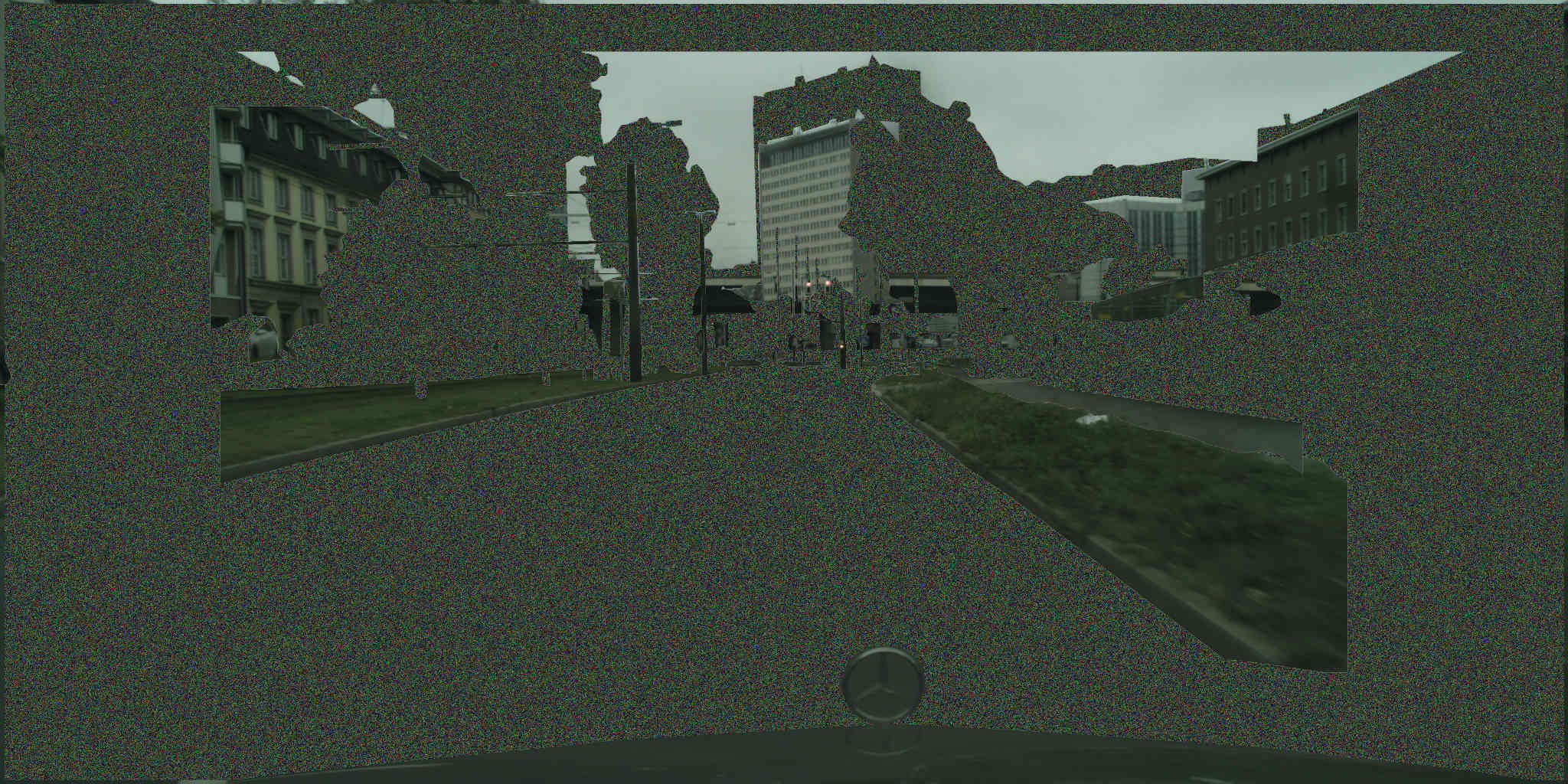}} & {\includegraphics[width=0.25\textwidth]{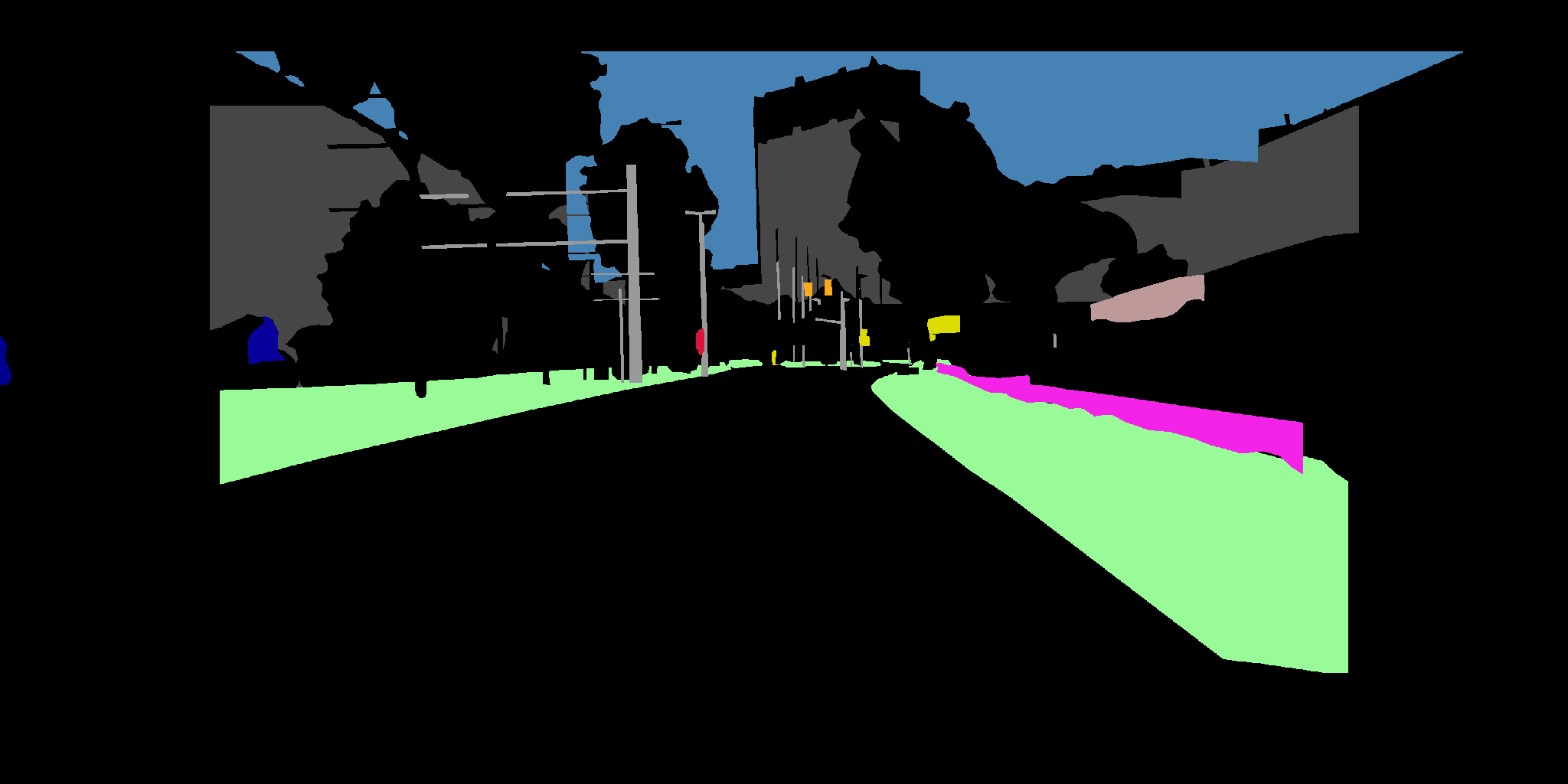}}\\
    
    {\includegraphics[width=0.25\textwidth]{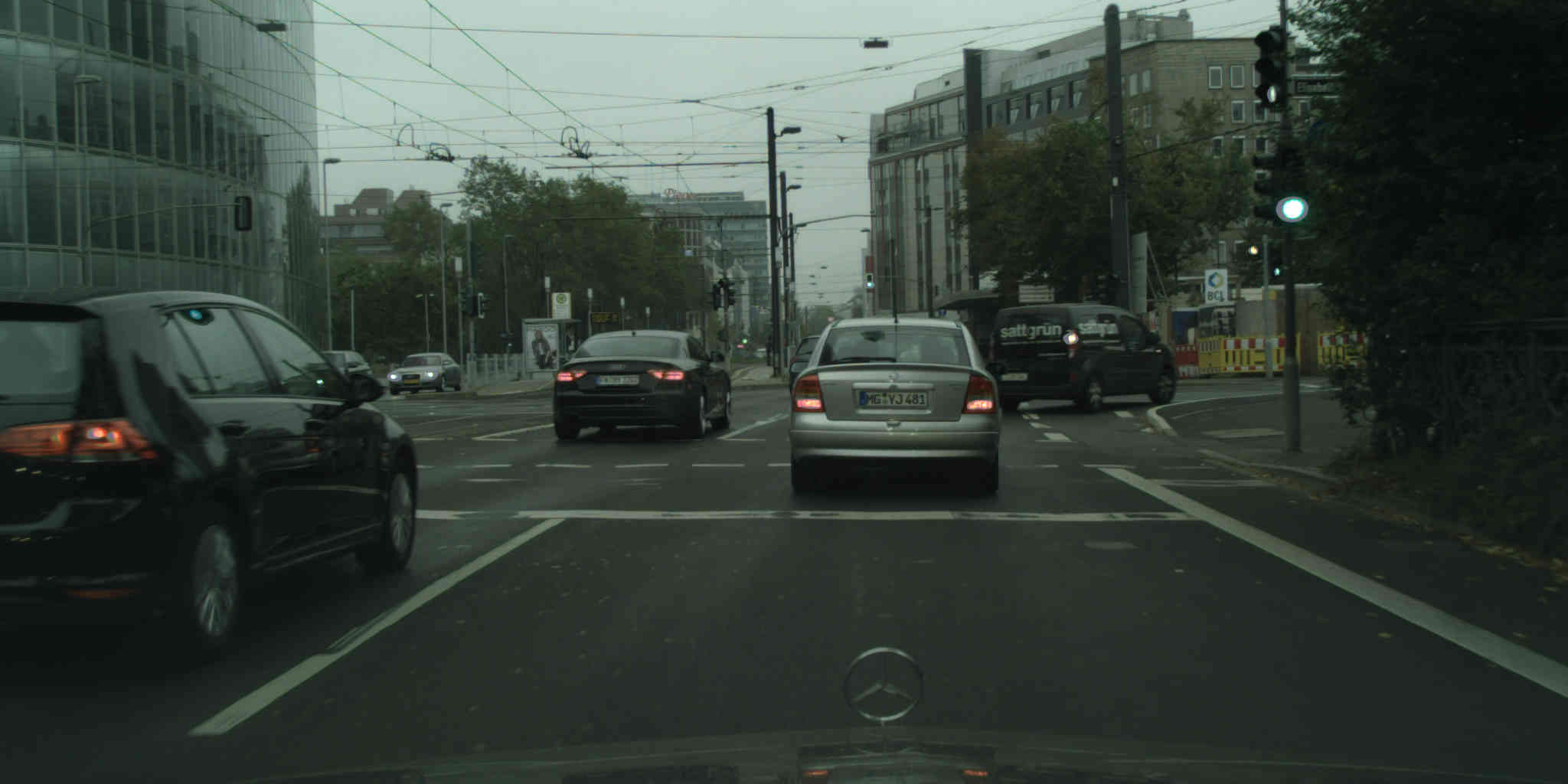}} & {\includegraphics[width=0.25\textwidth]{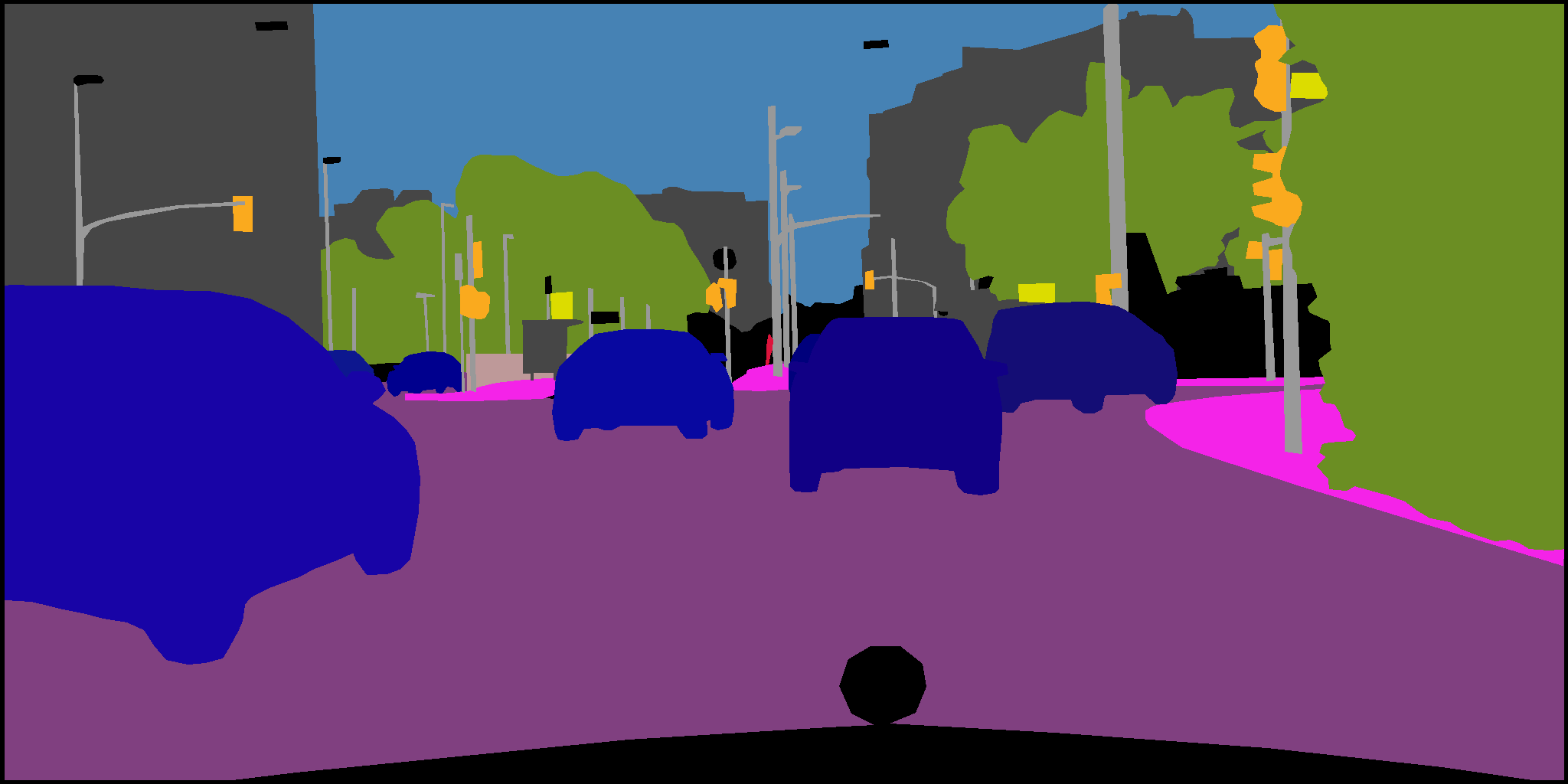}} & {\includegraphics[width=0.25\textwidth]{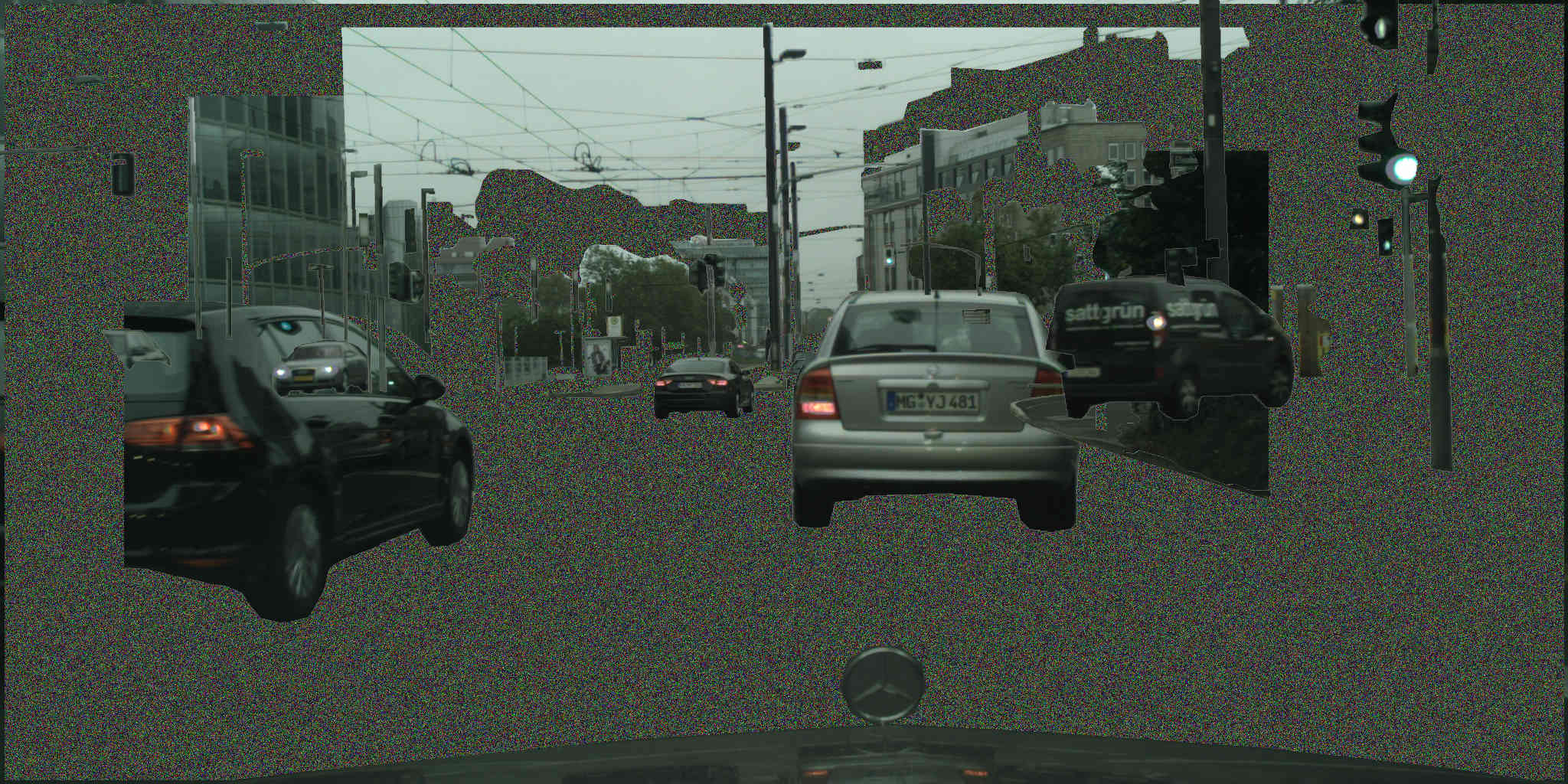}} & {\includegraphics[width=0.25\textwidth]{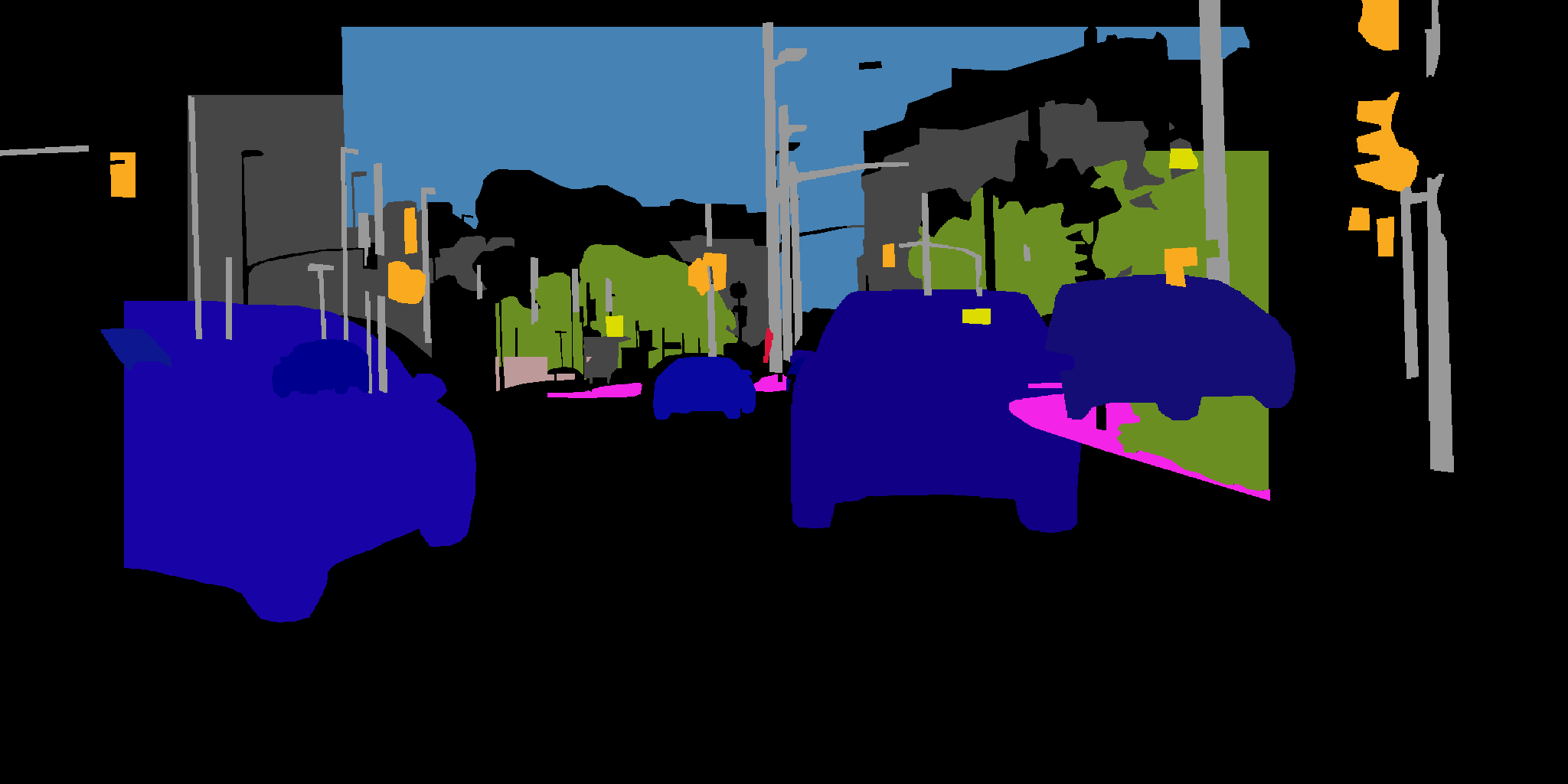}}\\
    
    {\includegraphics[width=0.25\textwidth]{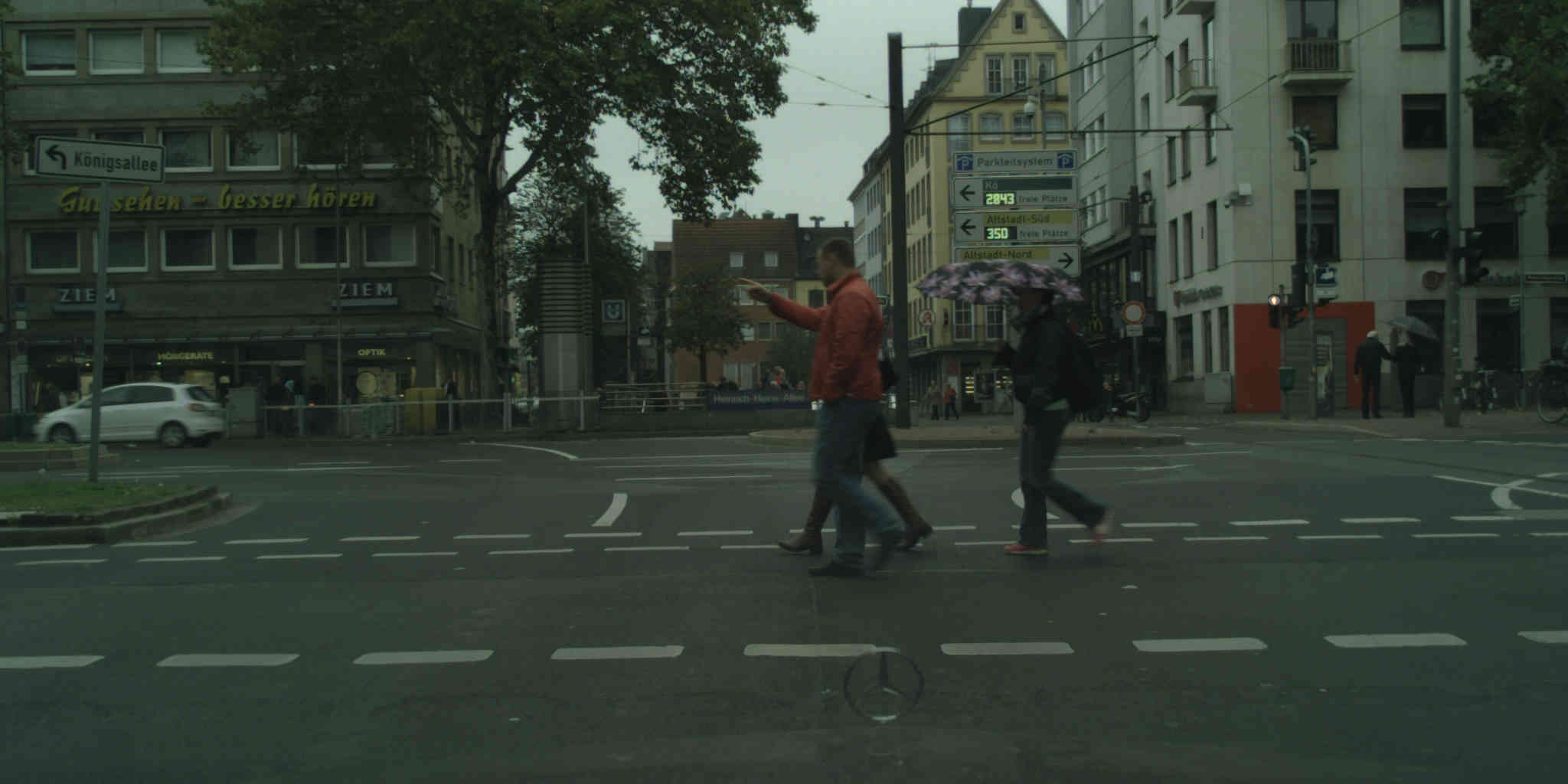}} & {\includegraphics[width=0.25\textwidth]{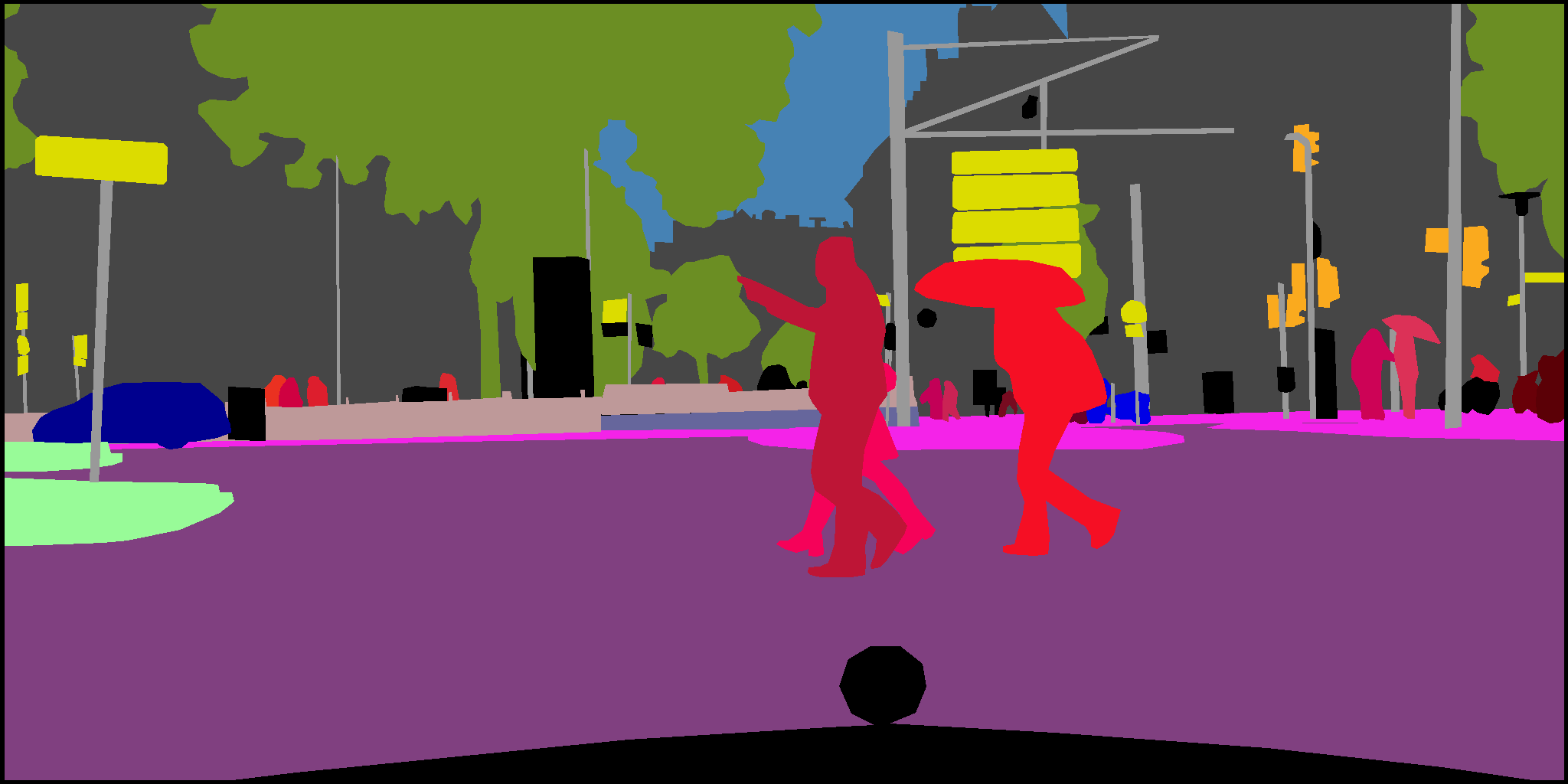}} & {\includegraphics[width=0.25\textwidth]{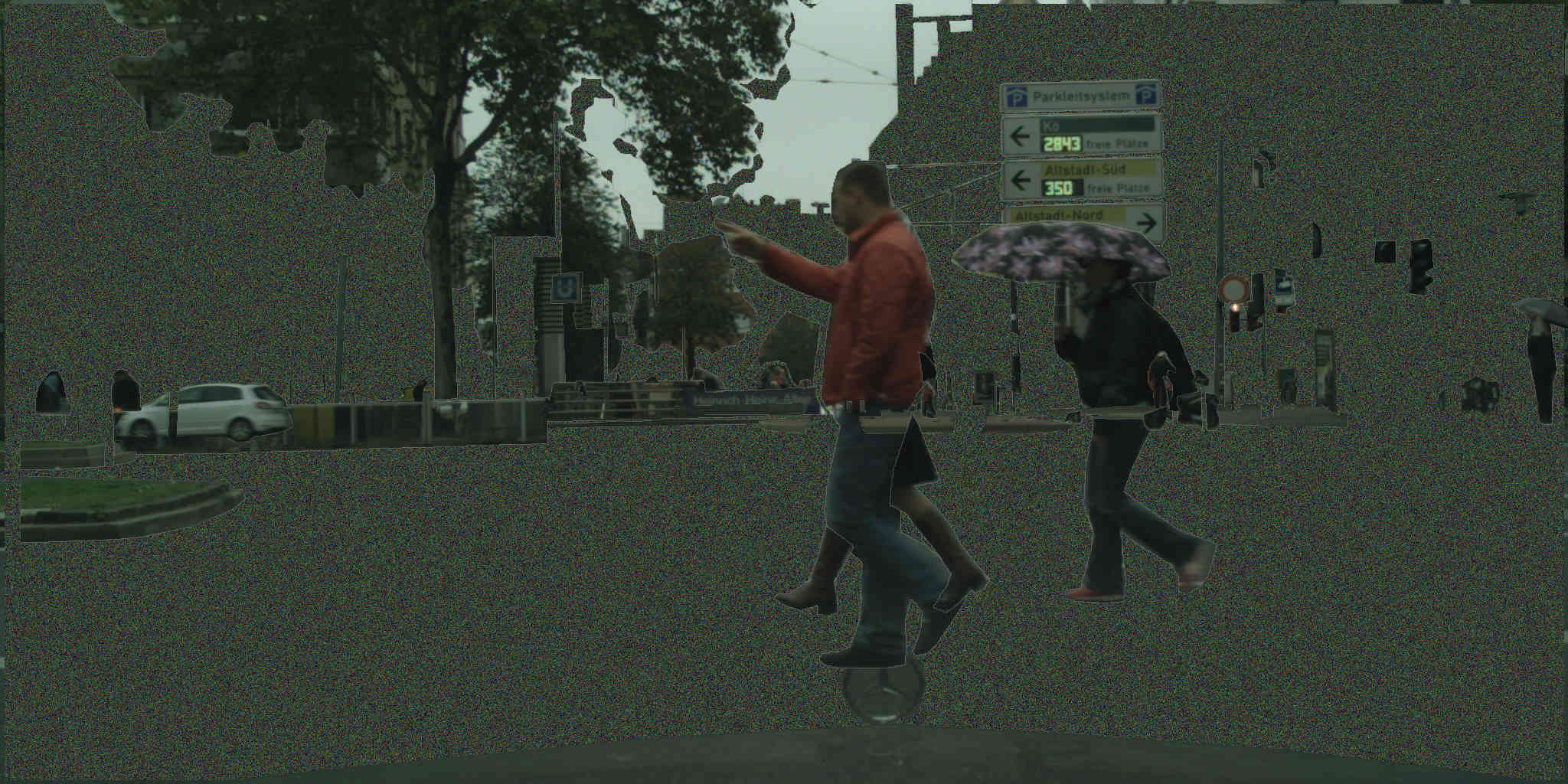}} & {\includegraphics[width=0.25\textwidth]{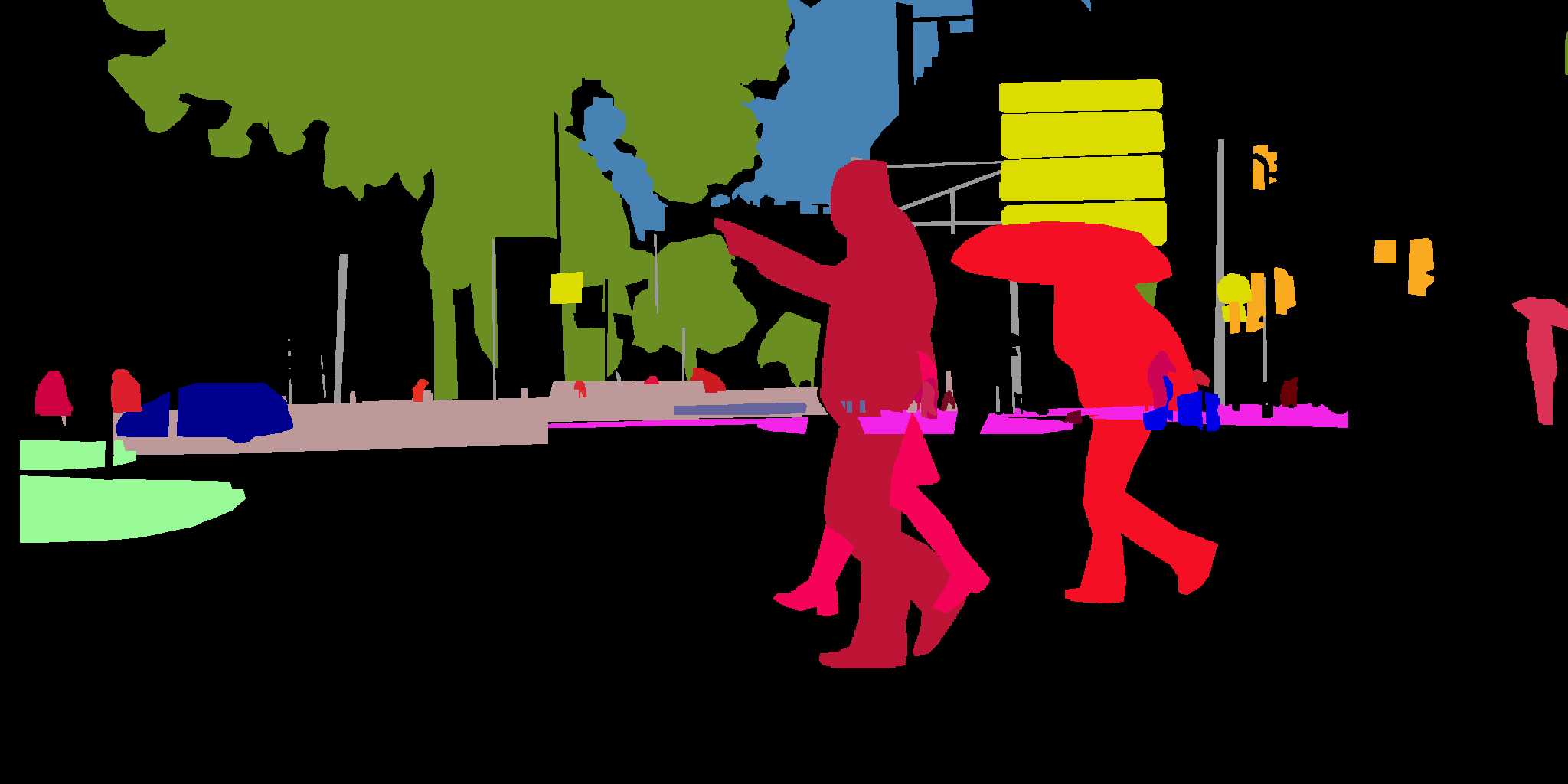}}\\

\end{tabular}
\addtolength{\tabcolsep}{5pt}
\end{center}
   \caption{\textbf{Examples of original and PanDA Cityscapes images with ground truth annotation.} \textit{Left two columns}: original Cityscapes images and ground truth annotation images. \textit{Right two columns}: PanDA generated images and ground truth annotation images.}
\label{fig:additional_cityscapes_examples}
\end{figure}

\begin{figure}[h]
\centering
\begin{center}
\centering
\addtolength{\tabcolsep}{-5pt} 
\begin{tabular}{c c c c }
    Original & Original annotation & PanDA & PanDA annotation \\ {\includegraphics[width=0.25\textwidth]{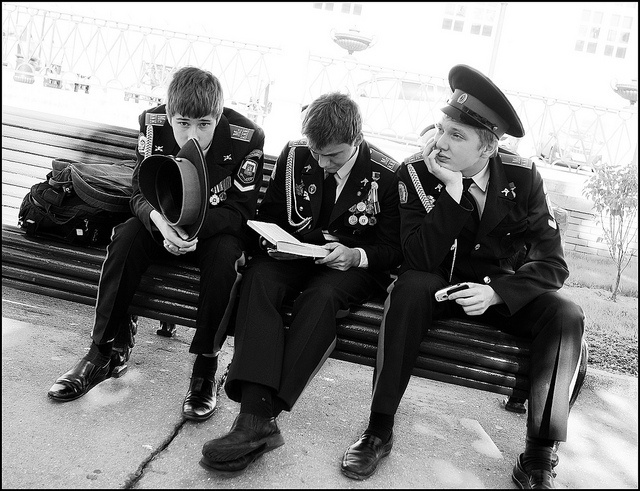}} & {\includegraphics[width=0.25\textwidth]{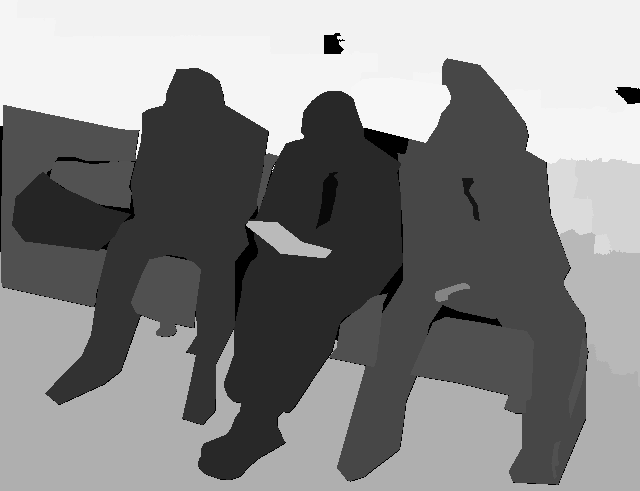}} & {\includegraphics[width=0.25\textwidth]{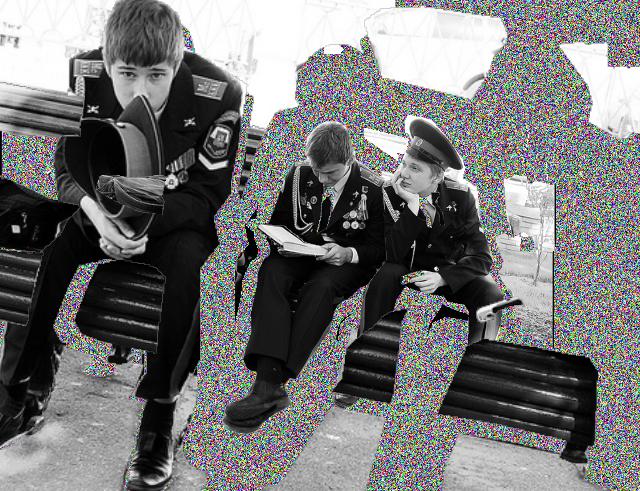}} & {\includegraphics[width=0.25\textwidth]{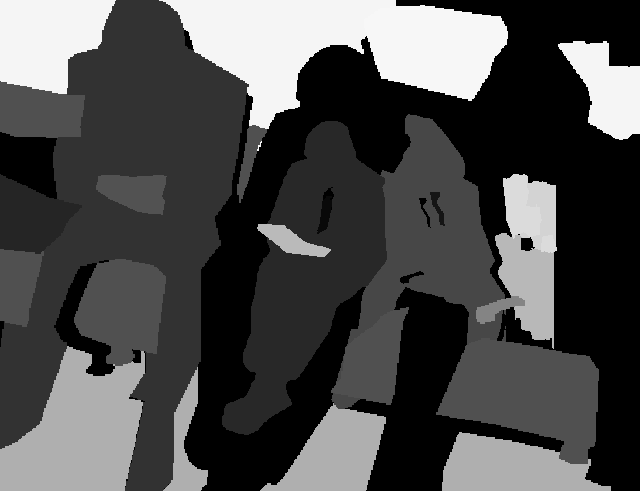}}\\ 
    
    {\includegraphics[width=0.25\textwidth]{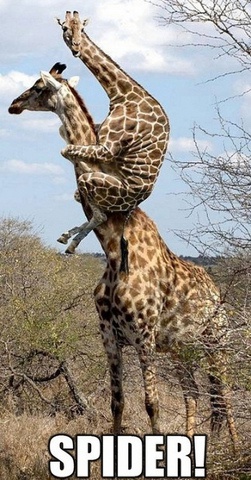}} & {\includegraphics[width=0.25\textwidth]{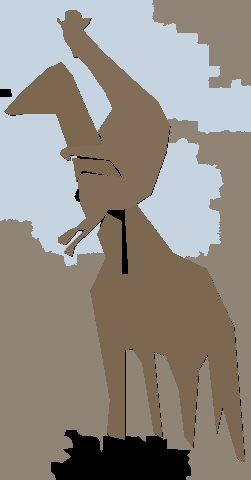}} & {\includegraphics[width=0.25\textwidth]{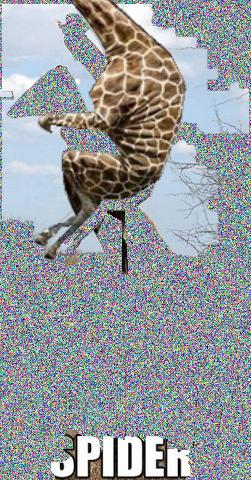}} & {\includegraphics[width=0.25\textwidth]{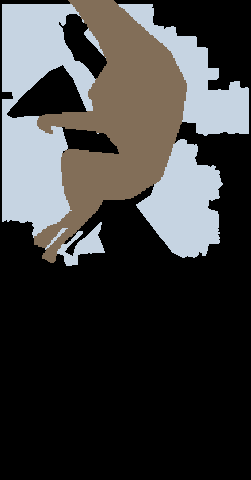}}\\ 
    
    {\includegraphics[width=0.25\textwidth]{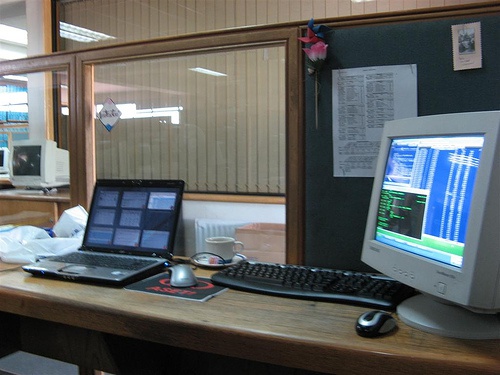}} & {\includegraphics[width=0.25\textwidth]{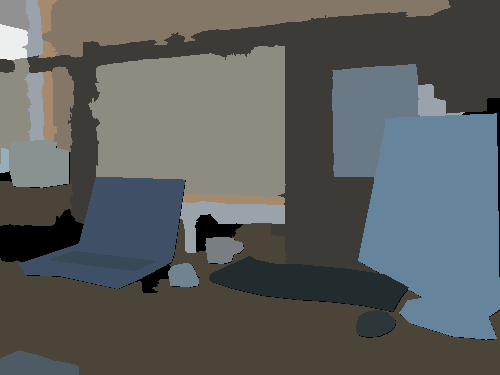}} & {\includegraphics[width=0.25\textwidth]{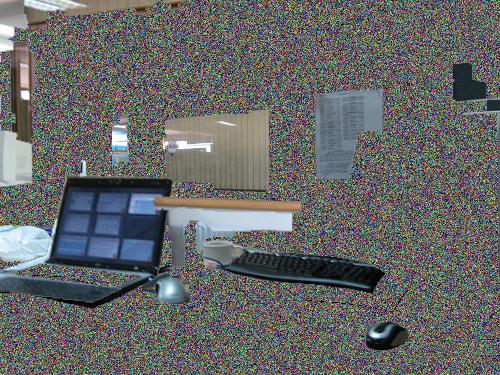}} & {\includegraphics[width=0.25\textwidth]{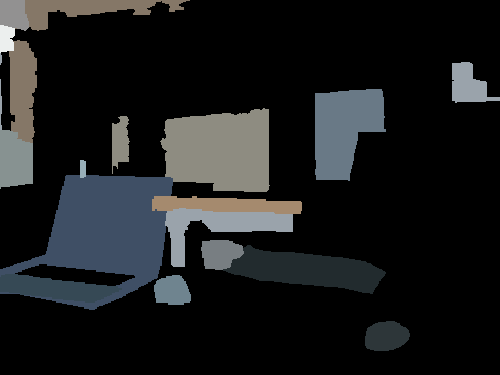}}\\ 
    
    {\includegraphics[width=0.25\textwidth]{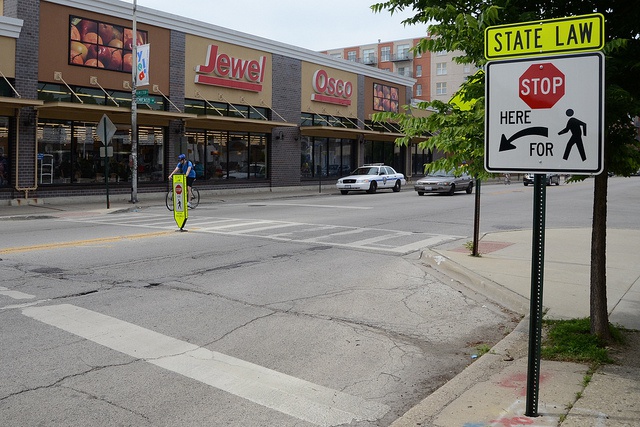}} & {\includegraphics[width=0.25\textwidth]{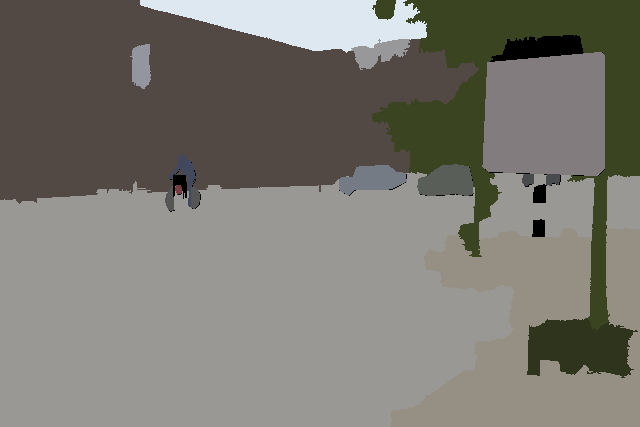}} & {\includegraphics[width=0.25\textwidth]{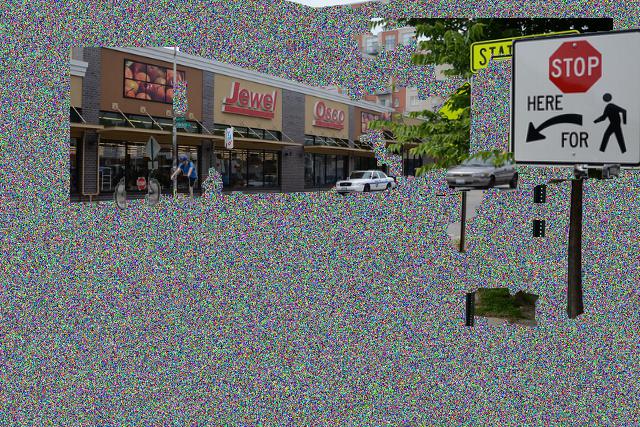}} & {\includegraphics[width=0.25\textwidth]{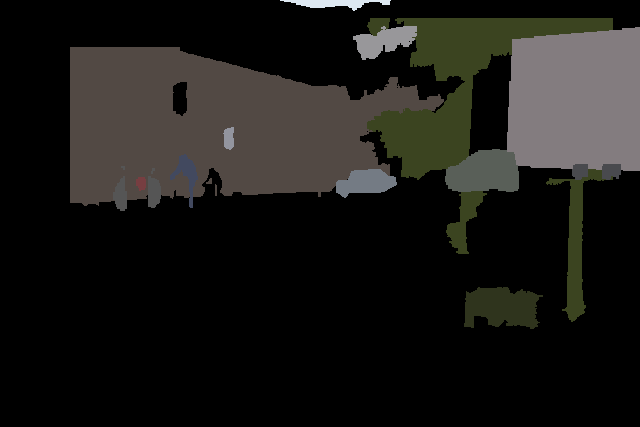}}\\ 
    
    {\includegraphics[width=0.25\textwidth]{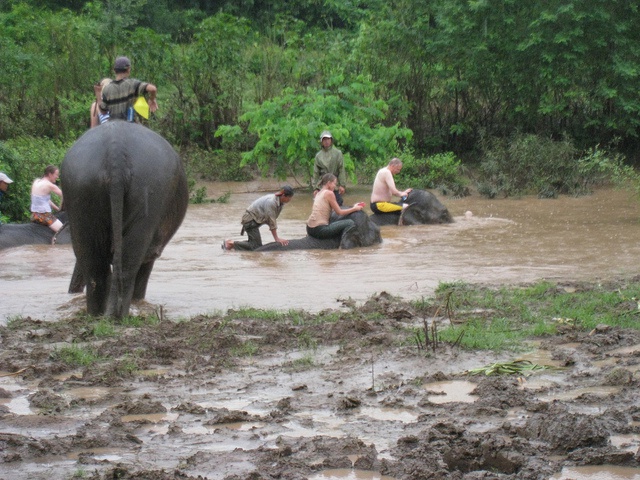}} & {\includegraphics[width=0.25\textwidth]{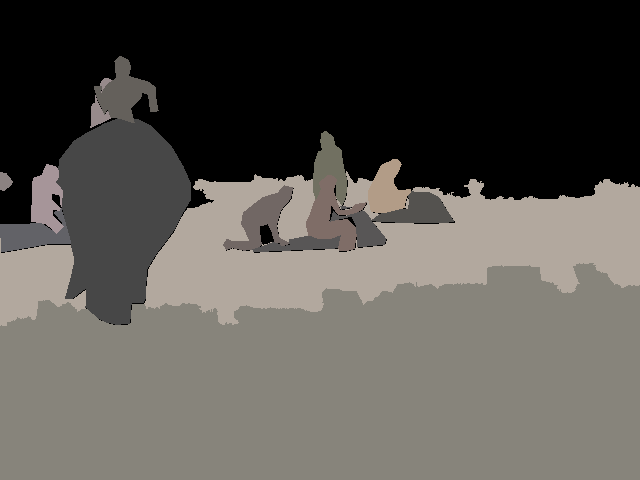}} & {\includegraphics[width=0.25\textwidth]{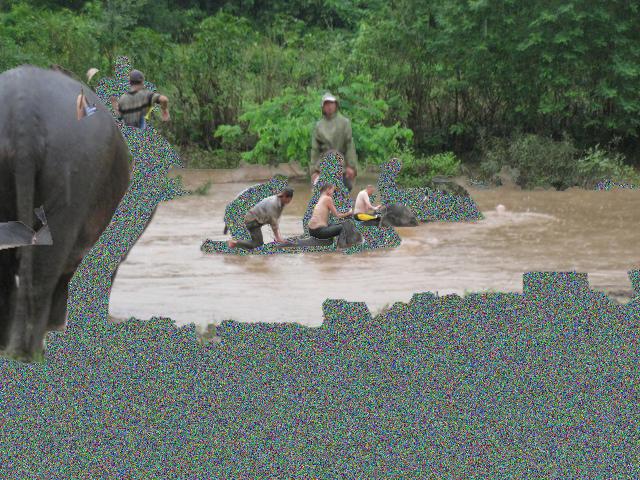}} & {\includegraphics[width=0.25\textwidth]{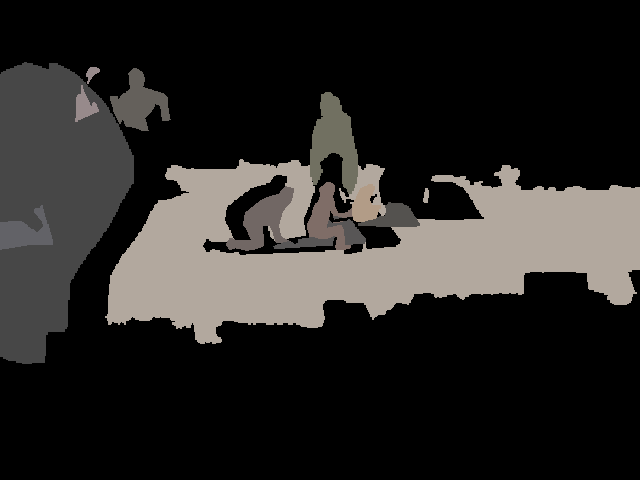}}\\

\end{tabular}
\addtolength{\tabcolsep}{5pt}
\end{center}
   \caption{\textbf{Examples of original and PanDA COCO images with ground truth annotation.} \textit{Left two columns}: original COCO images and ground truth annotation images. \textit{Right two columns}: PanDA generated images and ground truth annotation images.}
\label{fig:additional_coco_examples}
\end{figure}
\end{document}